\documentclass[11pt]{article}

\usepackage{microtype}
\usepackage{graphicx}
\usepackage{subfigure}
\usepackage{booktabs} % for professional tables
\usepackage{hyperref}
\usepackage{fullpage}%,cite,subcaption,titlesec,MnSymbol}
\usepackage{wrapfig}
\usepackage{hyperref,graphicx,amsmath,amssymb,bm,breakurl,epsfig,epsf,color,mathbbol,fmtcount,semtrans,multirow,comment,boldline,movie15,pgfplots}
\usepackage{tcolorbox}
\tcbuselibrary{skins}
\usepackage{tikz}
\usetikzlibrary{pgfplots.groupplots}
\usepackage[utf8]{inputenc} % allow utf-8 input
\usepackage[T1]{fontenc}    % use 8-bit T1 fonts
\usepackage{url}            % simple URL typesetting
\usepackage{booktabs}       % professional-quality tables
\usepackage{amsfonts}       % blackboard math symbols
\usepackage{nicefrac}       % compact symbols for 1/2, etc.
\usepackage{microtype}      % microtypography
\usepackage{mathtools}
\definecolor{darkred}{RGB}{150,0,0}
\definecolor{darkgreen}{RGB}{0,150,0}
\definecolor{darkblue}{RGB}{0,0,200}
\hypersetup{colorlinks=true, linkcolor=darkred, citecolor=darkgreen, urlcolor=darkblue}

\newtheorem{theorem}{Theorem}%[section]

\newtheorem{assumption}{Assumption}

\newenvironment{assbis}[1]
  {\renewcommand{\theassumption}{\ref{#1}$'$}%
   \addtocounter{assumption}{-1}%
   \begin{assumption}}
  {\end{assumption}}

\newtheorem{lemma}{Lemma}
\newtheorem{corollary}{Corollary}
\newtheorem{proposition}{Proposition}
\newtheorem{definition}{Definition}

\numberwithin{equation}{section}

\newcommand{\clr}[1]{\textcolor{black}{#1}}
\newcommand{\cln}[1]{\textcolor{red}{}}

\newcommand{\czmli}{\sqrt{{c_{0}}}}
\newcommand{\czsqr}{c_{0}}
\newcommand{\tnv}{\twonorm{\vct{v}}}
\newcommand{\lazb}{\bar{\lambda}_{0}}
\def \endprf{\hfill {\vrule height6pt width6pt depth0pt}\medskip}

\newenvironment{proof}{\noindent {\bf Proof} }{\endprf\par}

\newcommand{\tsn}[1]{{\left\vert\kern-0.25ex\left\vert\kern-0.25ex\left\vert #1 
    \right\vert\kern-0.25ex\right\vert\kern-0.25ex\right\vert}}

\newcommand{\noi}{\noindent}
\newcommand{\eps}{\varepsilon}
\newcommand{\cz}{c_0}
\newcommand{\cc}{\bar{c}}
\newcommand{\kz}{\nu}
\newcommand{\fnn}{f_{\text{nn}}}
\newcommand{\ff}{f_{\text{nn}}}
\newcommand{\wi}{k_\st}
\newcommand{\fF}[1]{f_{\text{nn},#1}}
\newcommand{\vc}{\text{vec}}
\newcommand{\flin}{f_{\text{lin}}}
\newcommand{\fln}[1]{f_{\text{lin},#1}}
\newcommand{\rest}{\text{rest}}

\newcommand{\Mb}{\bar{\M}}

\newcommand{\bp}{\bar{p}}
\newcommand{\zig}{\nu}
\newcommand{\bh}{\bar{h}}
\newcommand{\hb}{\bar{\vct{h}}}

\newcommand{\fs}{f^{\Dc}}
\newcommand{\Ncov}{\mathcal{N}_\eps(\Bal)}
\newcommand{\ft}{f^{\Tc}}
\newcommand{\ftv}{f^{\Tc\cup\Vc}}
\newcommand{\deff}{h_{\text{eff}}}
\newcommand{\defz}{\bar{h}_{\text{eff}}}
\newcommand{\deft}{\tilde{h}_{\text{eff}}}
\newcommand{\defg}{\bar{h}^{\nabla}_{\text{eff}}}

\newcommand{\bC}{\bar{C}}
\newcommand{\st}{\star}
\newcommand{\nt}{n_\mathcal{T}}
\newcommand{\h}{h}
\newcommand{\nv}{n_\mathcal{V}}
\newcommand{\distas}{\overset{\text{i.i.d.}}{\sim}}
\newcommand{\pleq}{\overset{{P}}{\leq}}

\newcommand{\rP}{\stackrel{{P}}{\longrightarrow}}

\newcommand{\beq}{\begin{equation}}
\newcommand{\ba}{\begin{align}}
\newcommand{\ea}{\end{align}}

\newcommand{\eeq}{\end{equation}}

\newcommand{\nn}{\nonumber}
\newcommand{\la}{\lambda}
\newcommand{\laz}{\la_0}
\newcommand{\blaz}{\bar{\la}_0}

\newcommand{\A}{{\mtx{A}}}

\newcommand{\Ub}{{\mtx{U}}}

\newcommand{\V}{{\mtx{V}}}

\newcommand{\B}{{{\mtx{B}}}}

\newcommand{\Gb}{{\mtx{G}}}

\newcommand{\Lc}{{\cal{L}}}
\newcommand{\Lcz}{{\cal{L}}^{0-1}}
\newcommand{\Lczh}{{\hat{\cal{L}}}^{0-1}}
\newcommand{\Lch}{{\widehat{\cal{L}}}}

\newcommand{\Jc}{{\cal{J}}}
\newcommand{\Dc}{{\cal{D}}}
\newcommand{\Dci}{{\cal{D}}_{\text{init}}}

\newcommand{\Pb}{{\mtx{P}}}

\newcommand{\La}{{\boldsymbol{{\Lambda}}}}

\newcommand{\bSi}{{\boldsymbol{{\Sigma}}}}

\newcommand{\bB}{{\bar{B}}}

\newcommand{\Iden}{{\mtx{I}}}
\newcommand{\M}{{\mtx{M}}}

\newcommand{\order}[1]{{\cal{O}}(#1)}
\newcommand{\ordet}[1]{{\widetilde{\cal{O}}}(#1)}

\newcommand{\z}{{\vct{z}}}

\newcommand{\tn}[1]{\|{#1}\|_{\ell_2}}
\newcommand{\tone}[1]{\|{#1}\|_{\ell_1}}
\newcommand{\lix}[1]{\|{#1}\|_{\Xc}}
\newcommand{\lif}[1]{\text{dist}_{\FB}({#1})}
\newcommand{\lit}[1]{\text{max}(#1)}%_{\Ttc}
\newcommand{\lia}[1]{\text{avg}(#1)}%

\newcommand{\tin}[1]{\|{#1}\|_{\ell_\infty}}
\newcommand{\trow}[1]{\|{#1}\|_{2,\infty}}

\newcommand{\Lcg}{\tilde{\Lc}}

\newcommand{\tf}[1]{\|{#1}\|_{F}}
\newcommand{\te}[1]{\|{#1}\|_{\psi_1}}

\newcommand{\Cc}{\mathcal{C}}
\newcommand{\Ac}{\mathcal{A}}
\newcommand{\Acr}{\mathcal{A_\text{ridge}}}
\newcommand{\Bal}{{\boldsymbol{\Delta}}}
\newcommand{\del}{\delta}

\newcommand{\Rc}{\mathcal{R}}

\newcommand{\bt}{{\boldsymbol{\theta}}}

\newcommand{\bts}{{\boldsymbol{\theta}_\st}}
\newcommand{\btb}{\bar{\boldsymbol{\theta}}}
\newcommand{\btid}{\boldsymbol{\theta}^\text{ideal}}
\newcommand{\btt}{\tilde{\boldsymbol{\theta}}}
\newcommand{\bal}{{\boldsymbol{\alpha}}}
\newcommand{\bgam}{{\boldsymbol{\gamma}}}
\newcommand{\bab}{{\boldsymbol{\bar{\alpha}}}}
\newcommand{\sbl}[1]{\sigma_{\boldsymbol{\alpha^{(#1)}}}}
\newcommand{\bl}[1]{{\boldsymbol{\alpha}}^{(#1)}}

\newcommand{\bah}{{\widehat{\bal}}}
\newcommand{\bas}{{\boldsymbol{\alpha}_\st}}
\newcommand{\bth}{{\boldsymbol{\hat{\theta}}}}
\newcommand{\bPhi}{{\boldsymbol{\Phi}}}

\newcommand{\DD}{{D}}

\newcommand{\pa}{{\partial}}
\newcommand{\Nn}{\mathcal{N}}

\newcommand{\vb}{\vct{v}}
\newcommand{\Jb}{\mtx{J}}

\newcommand{\FB}{\mathbb{F}}

\newcommand{\Xb}{\mtx{\bar{X}}}

\newcommand{\w}{\vct{w}}

\newcommand{\li}{\left<}
\newcommand{\ri}{\right>}

\newcommand{\ab}{\vct{a}}

\newcommand{\hh}{{\vct{h}}}
\newcommand{\g}{{\vct{g}}}

\newcommand{\Tc}{\mathcal{T}}
\newcommand{\TVc}{\Tc\cup\Vc}
\newcommand{\Ttc}{\mathcal{T}_{\text{test}}}
\newcommand{\Fc}{\mathcal{F}}
\newcommand{\Fcl}{\mathcal{F}^{\text{lin}}}
\newcommand{\Xc}{\mathcal{X}}
\newcommand{\Yc}{\mathcal{Y}}

\newcommand{\gb}{\bar{\g}}

\newcommand{\xh}{\hat{\x}}

\newcommand{\xt}{\widetilde{\x}}
\newcommand{\yt}{\widetilde{y}}

\newcommand{\opnorm}[1]{\left\|#1\right\|}
\newcommand{\fronorm}[1]{\left\|#1\right\|_{F}}
\newcommand{\onenorm}[1]{\left\|#1\right\|_{\ell_1}}
\newcommand{\twonorm}[1]{\left\|#1\right\|_{\ell_2}}

\newcommand{\infnorm}[1]{\left\|#1\right\|_{\ell_\infty}}

\newcommand{\abs}[1]{\left|#1\right|}
\newcommand{\lab}{\bar{\la}}

\newcommand{\mult}{B^D\bar{M}N}
\newcommand{\liptwo}{20R^3B\nt^2\laz^{-2}(B\nt+1)}
\newcommand{\lipp}{5R^2\sqrt{B^3\nt^3\h}\laz^{-2}\tn{\yT}}
\newcommand{\lips}{6R^3B^2\sqrt{\nt^3\h}\laz^{-2}\tn{\yT}}

\newcommand{\lip}{\frac{5B^2\tn{\yT}}{\laz^2}}
\newcommand{\lipt}{6R^3B^2\Gamma\sqrt{\nt^3\h}\laz^{-2} \tn{\yT}}
\newcommand{\lipf}{{20R^4B^2\laz^{-2}\Gamma\nt^2\tn{\yT}}}
\newcommand{\lipsum}{30R^4B^2\laz^{-2}\Gamma(\nt^2+\nv^2) \tn{\yT}}
\newcommand{\lipnn}{120B^4\blaz^{-2}\Gamma(\nt^2+\nv^2) \tn{\yT}}

\newcommand{\x}{\vct{x}}
\newcommand{\xx}[1]{\vct{x}^{(#1)}}

\newcommand{\y}{\vct{y}}
\newcommand{\yT}{\vct{y}}

\newcommand{\W}{\mtx{W}}
\newcommand{\Ww}[1]{\mtx{W}^{(#1)}}
\newcommand{\Wt}{\tilde{\mtx{W}}}

\newcommand{\Vc}{{\cal{V}}}
\newcommand{\bgl}{{~\big |~}}

\definecolor{emmanuel}{RGB}{255,127,0}

\newcommand{\Kb}{{\mtx{K}}}
\newcommand{\Kbb}{{\mtx{\bar{K}}}}
\newcommand{\Kbh}{{\widehat{\mtx{K}}}}

\newcommand{\pb}{{\vct{p}}}

\newcommand{\R}{\mathbb{R}}
\newcommand{\Pro}{\mathbb{P}}

\newcommand{\E}{\operatorname{\mathbb{E}}}

\newcommand{\e}{\mathrm{e}}
\newcommand{\eb}{\vct{e}}

\newcommand{\vct}[1]{\bm{#1}}
\newcommand{\mtx}[1]{\bm{#1}}

\newcommand{\Pc}{{\cal{P}}}
\newcommand{\X}{{\mtx{X}}}

\newcommand{\vs}{\vspace{-2pt}}

\begin{document}

\onecolumn
\title{Generalization Guarantees for Neural Architecture Search with\\Train-Validation Split}
\author{Samet Oymak\thanks{{Department of Electrical and Computer Engineering, University of California, Riverside, CA.~~~~~~Email: \url{soymak@ucr.edu}}}\quad\quad\quad Mingchen Li\thanks{{Department of Computer Science and Engineering, University of California, Riverside, CA.}~~~~~~~~~Email: \url{mli176@ucr.edu}}\quad\quad\quad Mahdi Soltanolkotabi\thanks{Ming Hsieh Dept.~of Electrical Engineering, University of Southern California, Los Angeles, CA.~Email: \url{soltanol@usc.edu}}}
\maketitle
\vspace{-4pt}
\begin{abstract} Neural Architecture Search (NAS) is a popular method for automatically designing optimized architectures for high-performance deep learning. In this approach, it is common to use bilevel optimization where one optimizes the model weights over the training data (lower-level problem) and various hyperparameters such as the configuration of the architecture over the validation data (upper-level problem). This paper explores the statistical aspects of such problems with train-validation splits. In practice, the lower-level problem is often overparameterized and can easily achieve zero loss. Thus, a-priori it seems impossible to distinguish the right hyperparameters based on training loss alone which motivates a better understanding of the role of train-validation split. To this aim this work establishes the following results:\\
\noi$\bullet$ We show that refined properties of the validation loss such as risk and hyper-gradients are indicative of those of the true test loss. This reveals that the upper-level problem helps select the most generalizable model and prevent overfitting with a near-minimal validation sample size. Importantly, this is established for continuous search spaces which are relevant for popular differentiable search schemes. Extensions to transfer learning are developed in terms of the mismatch between training \& validation distributions.\\ 
\noi$\bullet$ We establish generalization bounds for NAS problems with an emphasis on an activation search problem. When optimized with gradient-descent, we show that the train-validation procedure returns the best (model, architecture) pair even if all architectures can perfectly fit the training data to achieve zero error.\\
\noi$\bullet$ Finally, we highlight rigorous connections between NAS, multiple kernel learning, and low-rank matrix learning. The latter leads to novel algorithmic insights where the solution of the upper problem can be accurately learned via efficient spectral methods to achieve near-minimal risk.
\end{abstract}

{\let\thefootnote\relax\footnotetext{This work will appear in the International Conference on Machine Learning (ICML) 2021.}}\vspace{-4pt}
%\vs\vs\vs
\section{Introduction}\label{sec:intro}

Hyperparameter optimization (HPO) is a critical component of modern machine learning pipelines. It is particularly important for deep learning applications where there are many possibilities for choosing a variety of hyperparameters to achieve the best test accuracy. A crucial hyperparameter for deep learning is the architecture of the network. The architecture encodes the flow of information from the input to output, which is governed by the network's graph and the set of nonlinear operations that transform hidden feature representations. In this case HPO is often referred to as Neural Architecture Search (NAS). NAS is critical to finding the most suitable architecture in an automated manner without extensive user trial and error.

HPO/NAS problems are often formulated as bilevel optimization problems and critically rely on a train-validation split of the data, where the parameters of the learning model (e.g.~weights of the neural network) are optimized over the training data (lower-level problem), and the hyperparameters are optimized over a validation data (upper-level problem). With an ever growing number of configurations/architecture choices in modern learning problems, there has been a surge of interest in differentiable HPO methods that focus on continuous hyperparameter relaxations. For instance, differentiable architecture search schemes often learn continuously parameterized architectures which are discretized only at the end of the training \cite{liu2018darts}. Similar techniques have also been applied to learning data-augmentation policies \cite{cubuk2020randaugment} and meta-learning \cite{franceschi2018bilevel,finn2017model}. These differentiable algorithms are often much faster and seamlessly scale to millions of hyperparameters \cite{lorraine2020optimizing}. However, the generalization capability of HPO/NAS with such large search spaces and the benefits of the train-validation split on this generalization remain largely mysterious.

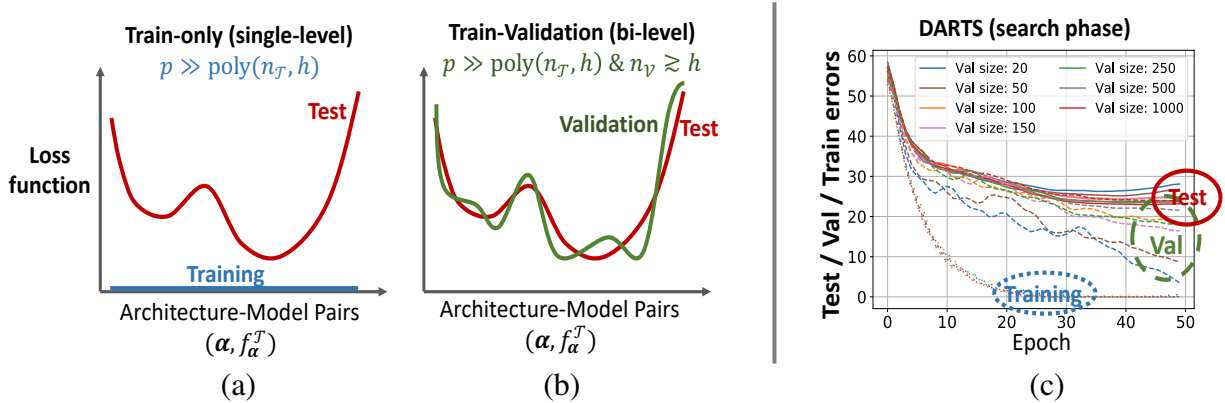
\begin{figure}\centering
\begin{tikzpicture}
	\node at (0,0) [scale=1]{\includegraphics[width=1\textwidth,height=0.3\textwidth]{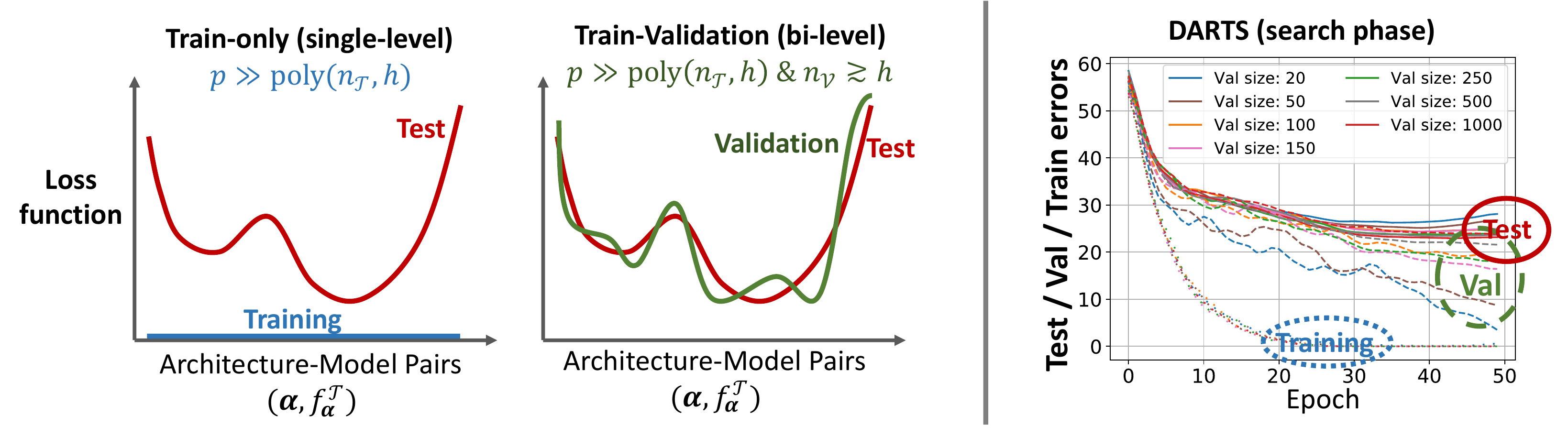}};
	\node at (-5,-2.65) [scale=0.7] {\LARGE{(a)}};
	\node at (-0.7,-2.65) [scale=0.7] {\LARGE{(b)}};
	\node at (5.75,-2.65) [scale=0.7] {\LARGE{(c)}};
\end{tikzpicture}
	\vspace{-20pt}\caption{\small{This figure depicts the typical scenario arising in modern hyperparameter optimization problems --such as those arising in NAS-- involving $p$ parameters, $\h$ hyperparameters and $\nt$ training data. In these modern settings, as depicted in Figure (a), the network is expressive enough so that it can perfectly fit the training data for all choices of (continuously-parameterized) architectures simultaneously. However, the network exhibits very different test performance for different hyperparameter choices. Perhaps surprisingly, as depicted in Figure (b), when we solve the problem via a train-validation split, as long as the amount of validation data $\nv$ is comparable to the number of hyperparameters $\h$ (i.e.~$\nv\gtrsim \h$), the validation loss uniformly concentrates around the test loss and enables the discovery of the optimal model even if the training loss is uniformly zero over all architectures. This paper aims to rigorously establish this phenomena (e.g., see our Theorem \ref{one layer nas supp}). Figure (c) shows NAS experiments on DARTS during the search phase. We evaluate train/test/validation losses of the continuously-parameterized supernet. These experiments are consistent with the depictions in Figures~(a) and (b) and our theory: Training error is consistently zero after 30 epochs. Validation error is never zero even with extremely few (20 or 50) samples. Validation error almost perfectly tracks test error as soon as validation size is mildly large (e.g.~$\geq 250$) which is also comparable with the number of architectural parameters ($h=224$).}}\label{overview fig}
\end{figure}

Addressing the above challenge is particularly important in modern overparameterized learning regimes where the training loss is often not indicative of the model's performance as large networks with many parameters can easily overfit to training data and achieve zero loss. To be concrete, let $\nt$ and $\nv$ denote the training and validation sample sizes and $p$ and $\h$ the number parameters and hyperparmeters of the model. In deep learning, NAS and HPO problems typically operate in a regime where
\vs\begin{align}
\boxed{p:=\#~\text{params.}\geq \nt\geq \nv\geq \h:=\#~\text{hyperparams.}}\label{roi eq}
\vs\vs\end{align}
Figure \ref{overview fig} depicts such a regime (e.g.~when $p\gg \text{poly}(\nt)$)) where the neural network model is in fact expressive enough to perfectly fit the dataset \emph{for all possible combinations of hyperparameters}. Nevertheless, training with a train-validation split tends to select the right hyperparameters where the corresponding network achieves stellar test accuracy. This leads us to the main challenge of this paper\footnote{While we do provide guarantees for generic HPO problems (cf.~Sec.~\ref{sec val gen}), the emphasis of this work is NAS and the search for the optimal architecture rather than broader class of hyperparameters.}:
\vspace{-0.1cm}
\begingroup
\addtolength\leftmargini{-0.11in}
\begin{quote}
\emph{How does train-validation split for NAS/HPO over large continuous search spaces discover near-optimal hyperparameters that generalize well despite the overparameterized nature of the problem?}
\end{quote}
\endgroup
\vs\vs
To this aim, in this paper, we explore the statistical aspects of NAS with train-validation split and provide theoretical guarantees to explain its generalization capability in the practical data/parameter regime of \eqref{roi eq}. Specifically, our contributions and the basic outline of the paper are as follows:

$\bullet$ \textbf{Generalization with Train-Validation Split (Section \ref{sec val gen}):} We provide general-purpose uniform convergence arguments to show that refined properties of the validation loss (such as risk and hyper-gradients) are indicative of the test-time properties. This is shown when the lower-level of the bilevel train-validation problem is optimized by an algorithm which is (approximately) Lipschitz with respect to the hyperparameters. Our result applies as soon as the validation sample size scales proportionally with the \emph{effective dimension of the hyperparameter space and only logarithmically in this Lipschitz constant}. We then utilize this result to obtain an end-to-end generalization bound for bilevel optimization with train-validation split under generic conditions. We also show that the aforementioned Lipschitzness condition holds in a variety of settings, such as: When the lower-level problem is strongly convex (the ridge regularization strength is allowed to be one of the hyperparameters) as well as a broad class of kernel and neural network learning problems (not necessarily strongly-convex) discussed next.

$\bullet$ \textbf{Generalization Guarantees for NAS (Sections \ref{sec:FMAP} \& \ref{sec deep}):} We also develop guarantees for NAS problems. Specifically, we first develop results for a \emph{neural activation search} problem that aims to determine the best activation function (among a continuously parameterized family) for overparameterized shallow neural network training. We study this problem in connection to a \emph{feature-map/kernel learning} problem involving the selection of the best feature-map among a continuously parameterized family of feature-maps. 
Furthermore, when the lower-level problem is optimized via gradient descent, we show that the \emph{bilevel problem is guaranteed to select the activation that has the best generalization capability}. This holds despite the fact that with any choice of the activation, the network can perfectly fit the training data. We then extend our results to deeper networks by similarly linking the problem of finding the optimal architecture to the search for the optimal kernel function. Using this connection, we show that train-validation split achieves the best excess risk bound among all architectures while requiring few validation samples and provide insights on the role of depth and width. 

$\bullet$ \textbf{Algorithmic Guarantees via Connection to Low-rank Learning (Section \ref{sec rank 1 factor}):} The results stated so far focus on generalization and are not fully algorithmic in the sense that they assume access to an approximately optimal solution of the upper-level (validation) problem (see \eqref{opt alpha}). As mentioned earlier, this is not the case for the lower-level problem: we specifically consider algorithms such as gradient descent. This naturally raises the question: Can one provably find such an approximate solution with a few validation samples and a computationally tractable algorithm? Towards addressing this question, we connect the shallow neural activation search problem to a novel low-rank matrix learning problem with an overparameterized dimension $p$. We then provide a two stage algorithm on a train-validation split of the data to find near-optimal hyperparameters via a spectral estimator that also achieves a near-optimal generalization risk. Perhaps unexpectedly, this holds as long as the matrix dimensions obey $\h\times p\lesssim (\nt+\nv)^2$ which allows for the regime \eqref{roi eq}. In essence, this demonstrates that it is possible to tractably solve the upper problem in the regime of \eqref{roi eq} even when the problem can easily be overfitting for all choices of hyperparameters. This is similar in spirit to practical NAS problems where the network can fit the data even with poor architectures.

\section{Preliminaries and Problem Formulation}\label{probset}
We begin by introducing some notation used throughout the paper. We use $\X^\dagger$ to denote the Moore–Penrose inverse of a matrix $X$. $\gtrsim,\lesssim$ denote inequalities that hold up to an absolute constant. We define the norm $\lix{\cdot}$  over an input space $\Xc$ as $\lix{f}:=\sup_{\x\in\Xc} |f(\x)|$. $\ordet{\cdot}$ implies equality up to constant/logarithmic factors. $c,C>0$ are used to denote absolute constants. Finally, we use $\Ncov$ to denote an $\eps$-Euclidean ball cover of a set $\Bal$.

Throughout, we use $(\x,y)\sim \Dc$ with $\x\in \Xc$ and $y\in\Yc$, to denote the data distribution of the feature/label pair. We also use $\Tc=\{(\x_i,y_i)\}_{i=1}^{\nt}$ to denote the training dataset and $\Vc=\{(\xt_i,\yt_i)\}_{i=1}^{\nv}$ the validation dataset and assume $\Tc$ and $\Vc$ are drawn i.i.d.~from $\Dc$. Given a loss function $\ell$ and a hypothesis $f:\Xc\rightarrow\Yc$, we define the population risk and the empirical validation risk as follows\vs
\begin{align}
\Lc(f)=\E_{\Dc}[\ell(y,f(\x))],\text{ }\Lch_\Vc(f)=\frac{1}{\nv}\sum_{i=1}^{\nv} \ell(\yt_i,f(\xt_i)).\label{val loss}
\end{align}
For binary classification with $y\in\{-1,1\}$ also define the test classification error as $\Lcz(f)=\Pro(yf(\x)\leq 0)$. We focus on a \emph{bilevel empirical risk minimization} (ERM) problem over train/validation datasets involving a hyperparameter $\bal\in\R^\h$ and a hypothesis $f$. Here, the model $f$ (which depends on the hyperparameter $\bal$) is typically trained over the training data $\Tc$ with the hyperparameters fixed (lower problem). Then, the best hyperparameter is selected based on the validation data (upper-level problem). 

While the training of the lower problem is typically via optimizing an (possibly regularized) empirical risk of the form $\Lch_\Tc(f)=\frac{1}{\nt}\sum_{i=1}^{\nt} \ell(y_i,f(\x_i))$, we do not explicitly require a global optima of this empirical loss and assume that we have access to an algorithm $\Ac$ that returns a model based on the training data $\Tc$ with hyperparameters fixed at $\bal$ 
\[\vs
\ft_\bal=\Ac(\bal,\Tc).\vs
\]
We provide some example scenarios with the corresponding algorithm below.

\noindent\emph{Scenario 1: Strongly Convex Problems.} The lower-level problem ERM is strongly convex with respect to the parameters of the model and $\Ac$ returns its unique solution. A specific example is learning the optimal kernel given a predefined set of kernels per \S\ref{sec:FMAPsec}. 

\noindent\emph{Scenario 2: Gradient Descent \& NAS.} In NAS, $f$ is typically a neural network and $\bal$ encodes the network architecture. Given this architecture, starting from randomly initialized weights, $\Ac$ trains the weights of $f$ on dataset $\Tc$ by running fixed number of gradient descent iterations. See \S\ref{sec:NAS} and \S\ref{sec deep} for more details. 

As mentioned earlier, modern NAS problems typically obey \eqref{roi eq} where the lower-level problem involves fitting an overparameterized network with many parameters whereas the number of architectural parameters $\h$ is typically less than 1000 and obeys $\h=\text{dim}(\bal)\leq \nv$. Intuitively, this is the regime in which all lower-level problems have solutions perfectly fitting the data. However, the under-parameterized upper problem can potentially guide the algorithm towards the right model. Our goal is to provide theoretical insights for this regime. To select the optimal model, given hyperparameter space $\Bal$ and tolerance $\del>0$, the following Train-Validation Optimization (TVO) returns a $\del$-approximate solution to the validation risk $\Lch_\Vc$ (upper problem)\vs\vs\vs\vs
\begin{align}
&\bah\in \{\bal\in\Bal\bgl \Lch_\Vc(\ft_\bal)\leq \min_{\bal\in\Bal}\Lch_{\Vc}(\ft_\bal)+\del\} \label{opt alpha}.\tag{TVO}
\end{align}
\vs\vs

\vs\vs
\section{Generalization with Train-Validation Split}\label{sec val gen}
In this section we state our generic generalization bounds for bilevel optimization problems with train-validation split. Next, in Sections \ref{sec:FMAP} and \ref{sec deep}, we utilize these generic bounds to establish guarantees for neural architecture/activation search --which will necessitate additional technical innovations. We start by introducing the problem setting in Section \ref{probset}. We then introduce our first result in Section \ref{lowvalgen} which controls the generalization gap between the test and validation risk as well as the corresponding gradients. Then, in Section \ref{end2end}, we relate training and validation risks which, when combined with our first result, yields an end-to-end generalization bound for the train-validation split.
\vs\vs
\subsection{Low validation risk implies good generalization}\label{lowvalgen}
{Our first result connects the test (generalization) error to that of the validation error.} A key aspect of our result is that we establish uniform convergence guarantees that hold over continuous hyperparameter spaces which is particularly insightful for differentiable HPO/NAS algorithms such as DARTS \cite{liu2018darts}. Besides validation loss, we will also establish the uniform convergence of the hyper-gradient $\nabla_\bal \Lch_\Vc(\ft_\bal)$ of the upper problem under similar assumptions. Such concentration of hyper-gradient is insightful for gradient-based bilevel optimization algorithms to solve \eqref{opt alpha}. Specifically, we will answer how many validation samples are required so that upper-level problems (hyper-)gradient concentrates around its expectation. Our results rely on the following definition and assumptions. 
\vs
\begin{definition} [Effective dimension] \label{cover assump} For a set $\Bal\in\R^\h$ of hyperparameters we define its effective dimension $\deff$ as the smallest value of $\deff>0$ such that $\abs{\Ncov}\le ({\bC}/{\eps})^{\deff}$ for all $\eps>0$ and a constant $\bC>0$.
\end{definition}\vs\vs
The effective dimension captures the degrees of freedom of a set $\Bal$. In particular, if $\Bal\in\R^\h$ has Euclidean radius $R$, then $\deff=\h$ with $\bC=3R$ so that it reduces to the number of hyperparameters. However, $\deff$ is more nuanced and can also help incorporate problem structure/prior knowledge (e.g.~sparse neural architectures have less degrees of freedom).\footnote{In the empirical process theory literature this is sometimes also referred to as the uniform entropy number e.g.~see \cite[Definition 2.5]{mendelson2003few}} \vs
\begin{assumption} \label{algo lip} $\Ac(\cdot)$ is an $L$-Lipschitz function of $\bal$ in $\lix{\cdot}$ norm, that is, for all pairs $\bal_1,\bal_2\in \Bal$, we have $\lix{\ft_{\bal_1}-\ft_{\bal_2}}\leq L\tn{\bal_1-\bal_2}$.
\end{assumption}\vs\vs

\begin{assumption}\label{loss assump}For all hypotheses $\ft_\bal$, the loss $\ell(y,\cdot)$ is $\Gamma$-Lipschitz over the feasible set $\{\ft_\bal(\x)\bgl \x\in\Xc\}$. Additionally, $\ell(y,\ft_\bal(\x))-\E[\ell(y,\ft_\bal(\x))]$ has bounded subexponential ($\|\cdot\|_{\psi_1}$) norm with respect to the randomness in $(\x,y)\sim\Dc$. 
\end{assumption}\vs\vs
Assumption \ref{algo lip} (and a less stringent version stated in Assumption \ref{algo lipFB}) is key to our NAS generalization analysis and we show it holds in a variety of scenarios. Assumption \ref{loss assump} requires the loss or gradient on a sample $(\x,y)$ to have a sub-exponential tail. While the above two assumptions allow us to show that the validation error is indicative of the test error, the two additional assumptions (which parallel those above) allow us to show that the hypergradient is concentrated around gradient of the true loss with respect to the hyperparameters. As mentioned earlier such concentration of the hyper-gradient is insightful for gradient-based bilevel optimization algorithms.
\begin{assbis}{algo lip}\label{algo lipp}
For some $R\geq1$ and all $\bal_1,\bal_2\in\Bal$ and $\x\in\Xc$, hyper-gradient obeys $\tn{\nabla_\bal\ft_{\bal_1}(\x)}\leq R$ and $\tn{\nabla_\bal\ft_{\bal_1}(\x)-\nabla_\bal\ft_{\bal_2}(\x)}\leq RL\tn{\bal_1-\bal_2}$.
\end{assbis}

\begin{assbis}{loss assump}\label{loss assumpp}
$\ell'(y,\cdot)$ is $\Gamma$-Lipschitz and the hyper-gradient noise $\nabla\ell(y,\ft_\bal(\x))-\E[\nabla\ell(y,\ft_\bal(\x))]$ over the random example $(\x,y)\sim\Dc$ has bounded subexponential norm as well.
\end{assbis}
Our first result establishes a generalization guarantee for \eqref{opt alpha} under these assumptions.
\vs
\begin{theorem}\label{thm val} Suppose Assumptions \ref{algo lip}\&\ref{loss assump} hold. Let $\bah$ be an approximate minimizer of the empirical validation risk per \eqref{opt alpha} and set $\defz:=\deff\log(\bC L\Gamma \nv/\deff)$. Also assume $\nv\geq \defz+\tau$ for some $\tau> 0$. Then, with probability at least $1-2e^{-\tau}$,
\vs\vs\begin{align}
\sup_{\bal\in\Bal}|\Lc(\ft_{\bal})- \Lch_{\Vc}(\ft_\bal)|\leq&\sqrt{\frac{C(\defz+\tau)}{\nv}}\label{eval1},\\
\Lc(\ft_\bah)\leq \min_{\bal\in \Bal}\Lc(\ft_\bal)+2&\sqrt{\frac{C(\defz+\tau)}{\nv}}+\delta.\label{excess risk val}
\end{align}\vs
Suppose also Assumptions \ref{algo lipp}\& \ref{loss assumpp} hold and $\nv\geq h+\defz+\tau$ for some $\tau>0$. Then, with probability at least $1-2e^{-\tau}$, the hyper-gradient of the validation risk converges uniformly. That is, 
\begin{align}
\sup_{\bal\in\Bal}\tn{\nabla\Lch_\Vc(\ft_\bal)-\nabla\Lc(\ft_\bal)}\leq \sqrt{\frac{C(h+\defz+\tau)}{\nv}}.\label{hyper eq2}
\end{align}\vs
\end{theorem}\vs\vs
This result shows that as soon as the size of the validation data exceeds the effective number of hyperparameters $\nv\gtrsim \deff$ (up to log factors) then (1) as evident per \eqref{eval1} the test error is close to the validation error (i.e.~validation error is indicative of the test error) and (2) per \eqref{excess risk val} the optimization over validation is guaranteed to return a hypothesis on par with the best choice of hyperparameters in $\Bal$. Theorem \ref{thm val} has two key distinguishing features, over the prior art on cross-validation \cite{kearns1997experimental,kearns1999algorithmic}, which makes it highly relevant for modern learning problems. The first distinguishing contribution of this result is that it applies to continuous hyperparameters and bounds the size of $\Bal$ via the refined notion of effective dimension, establishing a logarithmic dependence on problem parameters. This is particularly important for the Lipschtizness parameter $L$ which can be rather large in practice. The second distinguishing factor is that besides the loss function, per \eqref{hyper eq2} we also establish the uniform convergence of hyper-gradients. The reason the latter is useful is that if the validation loss satisfies favorable properties (e.g.~Polyak-Lojasiewicz condition), one can obtain generalization guarantees based on the stationary points of validation risk via \eqref{hyper eq2} (see \cite{foster2018uniform,sattar2020non}). We defer a detailed study of such gradient-based bilevel optimization guarantees to future work. Finally, we note that \eqref{hyper eq2} requires at least $h$ samples - which is the ambient dimension and greater than $\deff$. This is unavoidable due to the vectorial nature of the (hyper)-gradient and is consistent with related results on uniform gradient concentration \cite{mei2018landscape}.

Theorem \ref{thm val} is the simplest statement of our results and there are a variety of possible extensions that are more general and/or require less restrictive assumptions (see Theorem \ref{thm valFB} in Appendix \ref{app functional} for further detail). First, the loss function or gradient can be viewed as special cases of functionals of the loss function and as long as such functionals are Lipschitz with subexponential behavior, they will concentrate uniformly. Second, while Theorem \ref{thm val} aims to highlight our ability to handle continuous hyperparameters via the Lipschitzness of the algorithm $\Ac$, Assumption \ref{algo lip} can be replaced with a much weaker (see Assumption \ref{algo lipFB} in the Appendix). In general, $\Ac$ can be discontinuous as long as it is approximately locally-Lipschitz over the set $\Bal$. This would allow for discrete $\Bal$ (requiring $\nv\propto\log|\Bal|$ samples). Additionally, when analyzing neural nets, we indeed prove approximate Lipschitzness (rather than exact Lipschitzness).

Finally, we note that the results above do not directly imply good generalization as they do not guarantee that the validation error ($\min_{\bal}\Lch_{\Vc}(\ft_\bal)$) or the generalization error ($\min_{\bal\in \Bal}\Lc(\ft_\bal)$) of the model trained with the best hyperparameters is small. This is to be expected as when there are very few training data one can not hope for the model $\ft_\bal$ to have good generalization even with optimal hyperparameters. However, whether the training phase is successful or not, the validation phase returns approximately the best hyperparameters even with a bad model! In the next section we do in fact show that with enough training data the validation/generalization of the model trained with the best hyperparameter is indeed small allowing us to establish an end-to-end generalization bound.

\subsection{{End-to-end generalization with Train-Validation Split}}\label{end2end}
We begin by briefly discussing the role of the training data which is necessary for establishing an end-to-end bound. To accomplish this, we need to characterize how the population loss of the algorithm $\Ac$ scales with the training data $\nt$. To this aim, let us consider the limiting case $\nt\rightarrow +\infty$ and define the corresponding model for a given set of hyperparameters $\bal$ as
\[
\fs_\bal:=\Ac(\bal,\Dc):=\lim_{\nt\rightarrow\infty} \Ac(\bal,\Tc).
\]
Classical learning theory results typically bound the difference between the population loss/risk of a model that is trained with finite training data ($\Lc(\ft_\bal)$) and the loss achieved by the idealized infinite data model ($\Lc(\fs_\bal)$) in terms of an appropriate complexity measure of the class and the size of the training data. In particular, for a specific choice of the hyperparameter $\bal$, based on classical learning theory \cite{bartlett2002rademacher}) a typical behavior is to have
\vs\begin{align}
\Lc(\ft_\bal)\leq \Lc(\fs_\bal)+\frac{\Cc_\bal^\Tc+C_0\sqrt{t}}{\sqrt{\nt}},\label{unif alll}
\vs\end{align}
with probability at least $1-e^{-t}$. Here, $\Cc_\bal^\Tc$ is a dataset-dependent complexity measure for the hypothesis set of the lower-level problem and $C_0$ is a positive scalar. We are now ready to state our end-to-end bound which ensures a bound of the form \eqref{unif alll} holds simultaneously for all choices of hyperparameters $\bal\in\Bal$. 
\vs
\begin{proposition} [Train-validation bound] \label{unif excess lem} Consider the setting of Theorem \ref{thm val} and for any fixed $\bal\in\Bal$ assume \eqref{unif alll} holds. Also assume $\fs_\bal$ (in $\|\cdot\|_{\Xc}$ norm) and $\Cc_\bal^\Tc$ have $\order{L}$-bounded Lipschitz constants with respect to $\bal$ over $\Bal$. Then with probability at least $1-3e^{-t}$ over the train $\Tc$~and~validation $\Vc$ datasets
\vs\begin{align}
\Lc(\ft_\bah)\leq \min_{\bal\in\Bal}\left(\Lc(\fs_\bal)+\frac{\Cc_\bal^\Tc}{\sqrt{\nt}}\right)+\sqrt{\frac{\ordet{\deff+t}}{\nv}}+\delta.\nn%\label{abcdef}
\end{align}\vs
%\red{Additionally, let $\ftv_\bah=\Ac(\bah,\TVc)$ be the model trained on the union of train and validation. With probability at least $1-e^{-\tau}$, it obeys the bound
%\[
%\Lc(\ftv_\bah)\leq \Lc(\fs_\bah)+\frac{\Cc_\bah^{\TVc}+\ordet{\sqrt{\deff+t}}}{\sqrt{\nt+\nv}}.
%\]}
\end{proposition}\vs
In a nutshell, the above bound shows that the generalization error of a model trained with train-validation split is on par with the best train-only generalization achievable by picking the best hyperparameter $\bal\in\Bal$. The only loss incurred is an extra $\sqrt{\deff/\nv}$ term which is vanishingly small as soon as the validation data is sufficiently larger than the effective dimension of the hyperparameters. We note that the Lipschitzness condition on $\fs_\bal$ and $\Cc_\bal^\Tc$ can be relaxed. For instance, Proposition \ref{unif excess lem22}, stated in Appendix \ref{secB3}, provides a strict generalization where the Lipschitz property is only required to hold over a subset of the search space $\Bal$. 

We note that classical literature on this topic \cite{kearns1996bound,kearns1999algorithmic} typically use model selection to select the complexity from a nested set of growing hypothesis spaces by using explicit regularizers of a form similar in spirit to $\Cc_\bal^\Tc$. Instead, Proposition \ref{unif excess lem} aims to implicitly control the model capacity via the complexity measure $\Cc_\bal^\Tc$ of the outcome of the lower-level algorithm $\Ac$. This implicit capacity control is what we will utilize in Theorem \ref{one layer nas supp} (via norm-based generalization) which is of interest in practical NAS settings\footnote{Indeed, to obtain meaningful bounds in the regime \eqref{roi eq}, $\Cc_\bal^\Tc$ should not be dimension-dependent (as \# of params $p\gtrsim \nt$).}. This is because capacity of modern deep nets are rarely controlled explicitly and in fact, larger capacity often benefits generalization ability. In the next section, we demonstrate how one can utilize this end-to-end guarantee within specific problems.

\subsection{{Adapting the Generalization Bound for Transfer Learning}}\label{trans}
In this section, we consider a transfer/meta learning setting where the training and validation distribution is allowed to differ. This has the interpretation where training data is the source task and validation data is the target task. Assume $\Tc$ is drawn i.i.d.~from $\Dc'$ whereas $\Vc$ is drawn i.i.d.~from $\Dc$. On the training data, we train a hypothesis $f=[\bar{f},\tilde{f}]\in\Fc$ where $\bar{f}$ is part of the hypothesis used by the validation phase (such as feature representation) whereas $\tilde{f}$ is not necessarily used by the validation phase (e.g.~header specific to the source task). We also assume that $\bal$ is allowed to parameterize $f$ as well as the loss function (e.g.~via loss function, regularization, data augmentation). We solve the following bilevel ERM
\begin{align}
&\bah=\arg\min_{\bal\in\Bal}\Lch_\Vc(f_\bal):=\frac{1}{\nv}\sum_{i=1}^{\nv} \ell(\yt_i,f^\Tc_\bal(\xt_i))\label{trans prob}\\
&f^\Tc_\bal=[\bar{f}^\Tc_\bal,\tilde{f}^\Tc_\bal]=\underset{f=[\bar{f},\tilde{f}]\in \Fc_\bal}{\arg\min}\Lch_\Tc(f;\bal):=\frac{1}{\nt}\sum_{i=1}^{\nt} \ell(y_i,f(\x_i);\bal)
\end{align}
Here, to highlight that $f$ is parameterized by $\bal$, on the right hand side, we use the $\Fc_\bal$ notation to denote the dependence of the hypothesis set on $\bal$. %However we opted to not use explicit $f_\bal$ to keep notation simpler. 
\begin{definition}[Source-Target Distribution Mismatch] Let $\Dc,\Dc'$ be validation/target and training/source distributions respectively. Define $\ell^\st_\bal(\Dc)=\underset{f\in \Fc_\bal}{\min}\E_{(\x,y)\sim\Dc}[\ell(y,f(\x);\bal)]$ and define the suboptimality of a model $f\in\Fc$ as
\[
S^\bal_{\Dc}(f)=\E_{(\x,y)\sim\Dc}[\ell(y,f(\x);\bal)]-\ell^\st_\bal(\Dc).
\] 
%Fix sub-optimality tolerance $\eps>0$. and consider the $\eps$ near-optimal source/training models defined as
%\begin{align}
%\Fc^\eps_\bal=\{f \in\Fc_\bal\bgl \E_{(\x,y)\sim\Dc'}[\ell(y,f(\x);\bal)]\leq \ell^\st_\bal(\Dc')+\eps\}.
%\end{align}
Fix sub-optimality tolerance $\eps>0$. The source-target mismatch is then defined as the sub-optimality of the target/validation model induced by using distribution $\Dc'$ (rather than $\Dc$) as follows
\[
\text{DM}_{\Dc'}^\Dc(\bal,\eps)=\sup_{S^\bal_{\Dc'}(f)\leq \eps} S_{\Dc}(f)-S_{\Dc'}(f).%\Lc_\Dc(f_\bal^{\Dc'})-\Lc_\Dc(f_\bal^{\Dc}).
\]
\end{definition}
In words, $\text{DM}_{\Dc'}^\Dc(\bal,\eps)$ assesses the degradation in the validation risk due to the distributional mismatch in the infinite sample regime. The tolerance $\eps$ restricts our attention to the near-optimal models with accuracy close to the population minima. Let us also define the global mismatch by setting $\eps=\infty$ and declaring $\text{DM}_{\Dc'}^\Dc(\bal):=\text{DM}_{\Dc'}^\Dc(\bal,\infty)$. Let $(\eps_i)_{i=1}^{\nt}$ be Rademacher random variables. Define Rademacher complexity of a function class $\Fc$ as
\begin{align}
\Rc_{\nt}(\Fc)&=\frac{1}{\nt}\E_{\Tc,\eps_i}\Bigg[\sup_{f\in\Fc}\sum_{i=1}^{\nt}\eps_if(\x_i)\Bigg].
\end{align}

Following this, we can state the following generalization bound that jointly addresses transfer learning and hyperparameter tuning bound which is a direct corollary of Proposition \ref{unif excess lem}. 

\begin{corollary} [Transfer+Hyperparameter Tuning] \label{unif trans prop} Consider the setting of Theorem \ref{unif excess lem}. Also assume $\ell^\st_\bal(\Dc)$, $\text{DM}_{\Dc'}^\Dc(\bal)$ and $\Rc_{\nt}(\Fc_\bal)$ have $\order{L}$-bounded Lipschitz constants with respect to $\bal$ over $\Bal$. Then with probability at least $1-3e^{-t}$ over the train $\Tc$~and~validation $\Vc$ datasets
\vs\begin{align}
\Lc(\ft_\bah)\leq \min_{\bal\in\Bal}\left(\ell^\st_\bal(\Dc)+\text{DM}_{\Dc'}^\Dc(\bal)+4\Gamma \Rc_{\nt}(\Fc_\bal)\right)+\sqrt{\frac{\ordet{\deff+t}}{\nv}}+\delta.\nn%\label{abcdef}
\end{align}\vs
%\red{Additionally, let $\ftv_\bah=\Ac(\bah,\TVc)$ be the model trained on the union of train and validation. With probability at least $1-e^{-\tau}$, it obeys the bound
%\[
%\Lc(\ftv_\bah)\leq \Lc(\fs_\bah)+\frac{\Cc_\bah^{\TVc}+\ordet{\sqrt{\deff+t}}}{\sqrt{\nt+\nv}}.
%\]}
\end{corollary}
This corollary decomposes the excess risk into a transfer learning risk $\text{DM}_{\Dc'}^\Dc(\bal)$, minimum population risk $\ell^\st_\bal(\Dc)$, and Rademacher complexity $\Rc_{\nt}(\Fc_\bal)$ terms. Importantly, model selection guarantees the best combination of these risk sources. We remark that Lipschitzness requirement can be relaxed to Lipschitzness of an upper bound to these quantities or Lipschitzness over partitioning of $\Bal$ as in Assumption \ref{algo lipFB}.

\begin{proof} 
%Applying Theorem \ref{thm val}, we find that
%\begin{align}
%\Lc(\ft_\bah)\leq \min_{\bal\in \Bal}\Lc(\ft_\bal)+2&\sqrt{\frac{C(\defz+\tau)}{\nv}}+\delta.%\label{excess risk val}
%\end{align}
%where $\Lc$ is the population risk with respect to $\Dc$. What remains is controlling $\min_{\bal\in \Bal}\Lc(\ft_\bal)$ which obeys
%\begin{align}
%\min_{\bal\in \Bal}\Lc_{\Dc}(\ft_\bal)&=\min_{\bal\in \Bal}\{\ell^\st_\bal(\Dc)+S^\bal_{\Dc}(\ft_\bal)\}\\
%&\leq \min_{\bal\in \Bal}\{\ell^\st_\bal(\Dc)+\text{DM}_{\Dc'}^\Dc(\bal,\eps)+S^\bal_{\Dc'}(\ft_\bal)\}.
%%&\leq \min_{\bal\in \Bal}\{\Lc_{\Dc'}(\ft_\bal)+[\Lc_{\Dc}(\ft_\bal)-\Lc_{\Dc'}(\ft_\bal)]\}
%\end{align}
%Observe that
%\[
%\Lc_{\Dc}(\ft_\bal)-\Lc_{\Dc'}(\ft_\bal)\leq 
%\]
Applying Lemma \ref{rad comp lem} for the source/training distribution, we conclude: For a fixed $\bal$, with probability $1-2e^{-t}$, for all $f\in \Fc_\bal$, we have that 
$|\Lc_{\Dc'}(f;\bal)-\Lch_\Tc(f;\bal)|\leq 2\Gamma \Rc_{\nt}(\Fc_\bal)+C\sqrt{\frac{t}{\nt}}$.
Using optimality of $\ft_\bal$, this implies
$
S^\bal_{\Dc'}(\ft_\bal)=\Lc_{\Dc'}(\ft_\bal;\bal)-\ell^\st_\bal(\Dc')\leq 4\Gamma \Rc_{\nt}(\Fc_\bal)+2C\sqrt{\frac{t}{\nt}}.
$
Thus, using the definition of mismatch, we obtain the target/validation bound
\[
\Lc_{\Dc}(\ft_\bal)\leq \ell^\st_\bal(\Dc)+\text{DM}_{\Dc'}^\Dc(\bal)+4\Gamma \Rc_{\nt}(\Fc_\bal)+2C\sqrt{\frac{t}{\nt}}.
\]
Observe that, choosing Algorithm $\Ac$ as the empirical risk minimization, we have $\Lc(\fs_\bal)=\ell^\st_\bal(\Dc)$. Thus, the condition \eqref{unif alll} holds with $\frac{\Cc_\bal^\Tc}{\sqrt{\nt}}=\text{DM}_{\Dc'}^\Dc(\bal)+4\Gamma \Rc_{\nt}(\Fc_\bal)$ concluding the proof.
\end{proof}

\vs\vs
\section{Feature Maps and Shallow Networks}

\label{sec:FMAP}
In this section and Section \ref{sec deep}, we provide our main results on neural architecture/activation search which will utilize the generalization bounds provided above. Towards understanding the NAS problem, we first introduce the feature map selection problem \cite{khodak2019weight}. This problem is similar in spirit to the multiple kernel learning \cite{gonen2011multiple} problem which aims to select the best kernel for a learning task. This problem can be viewed as a simplified linear NAS problem where the hyperparameters control a linear combination of features and the parameters of the network are shared across all hyperparameters. Building on our findings on feature maps/kernels, Section \ref{sec:NAS} will establish our main results on activation search for shallow networks.
 \vs\vs
 \subsection{Feature map selection for kernel learning}\label{sec:FMAPsec}
Below, the hyperparameter vector $\bal\in\R^{\h+1}$ controls both the choice of the feature map and the ridge regularization coefficient.

\vs\begin{definition}[Optimal Feature Map Regression]\label{fmap def} Suppose we are given $\h$ feature maps $\phi_i:\Xc\rightarrow \R^p$. Define the superposition $\phi_\bal(\cdot)=\sum_{i=1}^\h\bal_i\phi_i(\cdot)$. Given training data $\Tc$, the algorithm $\Ac$ solves the ridge regression with feature matrix $\bPhi_\bal:=\bPhi_\bal^\Tc$ via
\begin{align}
&\bt_\bal=\arg\min_{\bt}\tn{\y-\bPhi_\bal\bt}^2+\bal_{\h+1}\tn{\bt}^2\quad\label{ridge regg}\\
&\text{where}\quad \bPhi_\bal=[\phi_\bal(\x_1)~\phi_\bal(\x_2)~\dots~\phi_\bal(\x_{\nt})]^T.\label{fmap mat}
\end{align}
Here $\bal_{h+1}\in[\la_{\min},\la_{\max}]\subset\R^+\cup\{0\}$ controls the regularization strength. We then solve for optimal choice $\bah$ via \eqref{opt alpha} with hypothesis $\ft_\bal(\vb)=\vb^T\bt_\bal$.
\end{definition}
\vs
The main motivation behind studying problems of the form \eqref{ridge regg}, is obtaining the best linear superposition of the feature maps minimizing the validation risk. This is in contrast to building $\h$ individual models and then applying ensemble learning, which would correspond to a linear superposition of the $h$ kernels induced by these feature maps. Instead this problem formulation models weight-sharing which has been a key ingredient of state-of-the-art NAS algorithms \cite{pham2018efficient,li2020geometry} as same parameter $\bt$ has compatible dimension with all feature maps. In essence, for NAS, this parameter will correspond to the (super)network's weights and the feature maps will be induced by different architecture choices so that the formulation above can be viewed as the simplest of NAS problems with linear networks. Nevertheless, as we will see in the forthcoming sections this analysis serves as a stepping stone for more complex NAS problems. To apply Theorem \ref{thm val} to the optimal feature map regression problem we need to verify its assumptions/characterize $\defz$. 
\begin{lemma} \label{ridge thm}Suppose the feature maps and labels are bounded i.e.~$\sup_{\x\in\Xc,1\leq i\leq \h}\tn{\phi_i(\x)}^2\leq B$ and $|y|\leq 1$. Also assume the loss $\ell$ is bounded and $1$-Lipschitz w.r.t.~the model output. Set $\laz=\la_{\min}+\inf_{\bal\in\Bal}\sigma_{\min}^2(\bPhi_\bal)>0$. Additionally let $\Bal$ be a convex set with $\ell_1$ radius $R\geq 1$. Then, Theorem \ref{thm val} holds with $\defz=(h+1)\log(\liptwo)$.
\end{lemma}\vs
An important component of the proof of this lemma is that we show that when $\laz>0$, $f_\bal$ is a Lipschitz function of $\bal$ and Theorem \ref{thm val} applies. Thus per \eqref{opt alpha} in this setting one can provably and jointly find the optimal feature map and the optimal regularization strength as soon as the size of the validation exceeds the number of hyperparameters.

We note that there are two different mechanisms by which we establish Lipschitzness w.r.t.~$\bal$ in the above theorem. When $\la_{\min}>0$, the lower problem is strongly-convex with respect to the model parameters. As we show in the next lemma, this is more broadly true for any training procedure which is based on minimizing a loss which is strongly convex with respect to the model parameters.\vs
\begin{lemma} \label{convex smooth}Let $\Bal$ be a convex set. Suppose $f_\bal$ is parameterized by $\bt_\bal$ where $\bt_\bal$ is obtained by minimizing a loss function $\bar{\Lc}_\Tc(\bal,\bt):\Bal\times \R^p\rightarrow\R$. Suppose $\bar{\Lc}_\Tc(\bal,\bt)$ is $\mu$ strongly-convex in $\bt$ and $\bar{L}$ smooth in $\bal$. Then $\bt_\bal$ is $\sqrt{\bar{L}/\mu}$-Lipschitz in $\bal$.
\end{lemma}\vs

Importantly, Lemma \ref{ridge thm} can also operate in the ridgeless regime ($\la_{\min}=0$) even when the training loss is not strongly convex. This holds as long as the feature maps are not poorly-conditioned in the sense that
\vs\begin{align}
\inf_{\bal\in\Bal}\sigma_{\min}\left(\bPhi_\bal \bPhi_\bal^T\right)=\laz>0.\label{low bound}
\end{align}\vs
We note the exact value of $\laz$ is not too important as the effective number of hyperparameters only depends logarithmically on this quantity. Such a ridgeless regression setting has attracted particular interest in recent years as deep nets can often generalize well in an overparameterized regime without any regularization despite perfectly interpolating the training data. In the remainder of the manuscript, we focus on ridgeless regression problems with an emphasis on neural nets (thus we drop the $h+1$'th index of $\bal$).

Our next result utilizes Proposition \ref{unif excess lem} to provide an end-to-end generalization bound for feature map selection involving both training and validation sample sizes. Below we assume that \eqref{low bound} holds with high probability over $\Tc$.
\vs\begin{theorem}[End-to-end generalization for feature map selection] \label{e2e bound}Consider the setup in Definition \ref{fmap def} with $\bal_{h+1}=0$. Set $R=\sup_{\bal\in\Bal}\tone{\bal}$ and assume $\sup_{\x\in\Xc,1\leq i\leq \h}\tn{\phi_i(\x)}^2\leq B$ and $\ell$ in \eqref{val loss} is $\Gamma$-Lipschitz and bounded by a constant. Suppose \eqref{low bound} holds with probability at least $1-p_0$. Also $p\geq \nt\geq \nv\gtrsim \h\log(M)$ with $M=\lipsum$. Furthermore, let $\yT=[y_1~y_2~\dots~y_{\nt}]$. Then with probability at least $1-4e^{-t}-p_0$, the population risk (over $\Dc$) obeys
\[
\Lc(f_{\bah})\leq  \min_{\bal\in\Bal}2\Gamma \sqrt{\frac{B\y^T\Kb_\bal^{-1}\y}{\nt}}+C\sqrt{\frac{\h\log(M)+\tau}{\nv}}+\delta.
\]
\end{theorem}\vs
In this result, the excess risk term $\sqrt{\frac{\y^T\Kb_\bal^{-1}\y}{\nt}}$ becomes smaller as the kernel induced by $\bal$ becomes better aligned with the labeling function e.g.~when $\yT$ lies on the principal eigenspace of $\Kb_\bal$. This theorem shows that for the optimal feature map regression problem, bilevel optimization via a train-validation split returns a generalization guarantee on par with that of the best feature map (minimizing the excess risk) as soon as the size of the validation data exceeds the number of hyperparameters.

\vs\vs
\subsection{Activation search for shallow networks}\label{sec:NAS}
In this section we focus on an \emph{activation search} problem where the goal is to find the best activation among a parameterized family of activations for training a shallow neural networks based on a train-validation split. To this aim we consider a one-hidden layer network of the form $\x\mapsto\ff(\x)=\vb^T\sigma(\W\x)$ and focus on a binary classification task with $y\in\{-1,+1\}$ labels. Here, $\sigma:\R\rightarrow\R$ denotes the activation, $\W\in\R^{k\times d}$ input-to-hidden weights, and $\vb\in\R^d$ hidden-to-output weights. We focus on the case where the activation belongs to a family of activations of the form $\sigma_\bal=\sum_{i=1}^\h \bal_i\sigma_i$ with $\bal\in\Bal$ denoting the hyperparameters. Here, $\{\sigma_i\}_{i=1}^h$ are a list of candidate activation functions (e.g.,~ReLU, sigmoid, Swish).  The neural net with hyperparameter $\bal$ is thus given by
$
\fF{\bal}(\x)=\vb^T\sigma_\bal(\W\x).
$
For simplicity of exposition in this section we will only use the input layer for training thus the training weights are $\W$ with dimension $p=\text{dim}(\W)=k\times d$ and fix $\vb$ to have $\pm \sqrt{\cz/k}$ entries (roughly half of each) with a proper choice of $\cz>0$. In Section \ref{sec deep} we further discuss how our results can be extended to NAS beyond activation search and to deeper networks where all the layers are trained. 

\begin{tcolorbox}[
                  standard jigsaw,
                  opacityback=0,
                  boxsep=4pt,
                  left=0pt,right=0pt,top=0pt,bottom=0pt,
                  ]
\vs\noindent\textbf{Bilevel optimization for shallow activation search:} We now explain the specific gradient-based algorithm we consider for the lower-level optimization problem. For a fixed hyperparameter $\bal$, the lower-level optimization aims to minimize a quadratic loss over the training data of the form
\vs\[
\Lch_{\Tc}(\W)=\frac{1}{2}\sum_{i=1}^{\nt}(y_i-\fF{\bal}(\x_i,\W))^2.
\]
To this aim, for a fixed hyperparameter $\bal\in\Bal$, starting from a random initialization of the form $\W_0\distas\Nn(0,1)$ we run gradient descent updates of the form $\W_{\tau+1}=\W_\tau-\eta \nabla \Lch_\Tc(\W_\tau)$  for $T$ iterations. Thus, the lower algorithm $\Ac$ returns the model
\vs\[
\ft_\bal(\x)=\vb^T\sigma_\bal(\W_T\x).
\]
We then solve for the $\del$-approximate optimal activation $\bah$ via \eqref{opt alpha}  by setting $\ell$ in \eqref{val loss} to be the hinge loss.
\end{tcolorbox}
 
To state our end-to-end generalization guarantee, we need a few definitions. First, we introduce neural feature maps induced by the Neural Tangent Kernel (NTK) \cite{jacot2018neural}.
\vs\begin{definition}[Neural feature maps \& NTK]\label{def ntk} Let $\fF{\bal}(\cdot,\bt)$ be a neural net parameterized by weights $\bt\in\R^p$ and architecture $\bal$. Define $\phi_\bal(\x)=\frac{\pa \fF{\bal}(\x)}{\pa \bt_0}$ to be the \emph{neural feature map} at the random initialization $\bt_0\sim\Dci$. Define the neural feature matrix $\bPhi_\bal=[\phi_\bal(\x_1)~\dots~\phi_\bal(\x_{\nt})]^T\in\R^{\nt\times p}$ as in \eqref{fmap mat} i.e.
\vs\begin{align}
\bPhi_\bal=\left[\frac{\pa \fF{\bal}(\x_1)}{\pa \bt_0}~\dots~\frac{\pa \fF{\bal}(\x_n)}{\pa \bt_0}\right]^T\label{neural fmap}.
\end{align} 
We define the gram matrix as $\Kbh_\bal=\bPhi_\bal\bPhi_\bal^T\in\R^{\nt\times \nt}$ with $(i,j)$th entry equal to $\li\phi_\bal(\x_i),\phi_\bal(\x_j)\ri$ and NTK matrix is as $\Kb_\bal=\E_{\bt_0}[\Kbh_\bal]$.
\end{definition}\vs

Neural feature maps are in general nonlinear function of $\bal$ (cf.~Sec.~\ref{sec deep}). However, in case of shallow networks, it is nicely additive and obeys $\phi_\bal(\x_i)=\sum_{i=1}^\h\bal_i \phi_1(\x_i)$ regardless of random initialization $\bt_0$ establishing a link to Def.~\ref{fmap def}. The next assumption ensures the expressivity of the NTK to interpolate the data and enables us to analyze regularization-free training. \vs
\begin{assumption}[Expressive Neural Kernels]\label{ntk assump} There exists $\laz>0$ such that for any $\bal\in\Bal$, the NTK matrix $\Kb_\bal\succeq \la_0\Iden_{\nt}$. 
\end{assumption}\vs
This assumption is similar to \eqref{low bound} but we take expectation over random $\bt_0$. Assumptions in a similar spirit to this are commonly used for the optimization/generalization analysis of neural nets, especially in the interpolating regime \cite{arora2019fine,chizat2018lazy,cao2019generalization,mei2019generalization}. For fixed $\bal$, $\Kb_\bal\succ 0$ as long as no two training inputs are perfectly correlated and $\phi_\bal$ is analytic and not a polynomial \cite{du2019gradient}. The key aspect of our assumption is that we require the NTK matrices to be lower bounded for all $\bal$. Later in Theorem \ref{gen thm generic} of \S\ref{sec deep} we shall show how to circumvent this assumption with a small ridge regularization.

With these definitions in place we are now ready to state our end-to-end generalization guarantee for Shallow activation search where the lower-level problem is optimized via gradient descent. The reader is referred to Theorem \ref{one layer nas supp2} for the precise statement. Note that $\laz$ and $\Kb_\bal$ scales linearly with initialization variance $\cz$. To state a result invariant to initialization, we will state our result in terms of the normalized eigen lower bound $\blaz=\laz/\cz$ and kernel matrix $\Kbb_\bal=\Kb_\bal/\cz$.
\begin{theorem}[Neural activation search]\label{one layer nas supp} Suppose input features have unit Euclidean norm i.e.~$\tn{\x}= 1$ and labels take values in $\{-1,1\}$. Pick $\Bal$ to be a subset of the unit $\ell_1$ ball. Suppose Assumption \ref{ntk assump} holds for $\bt_0\leftrightarrow\W_0$ and the candidate activations have first two derivatives ($|\sigma'_i|,|\sigma''_i|$) upper bounded by $B>0$. Furthermore, fix $\vb$ with half $\sqrt{\cz/k}$ and half $-\sqrt{\cz/k}$ entries for a proper $\cz$ (see supplementary). Define the normalized lower bound $\blaz=\laz/\cz$ and kernel matrix $\Kbb_\bal=\Kb_\bal/\cz$. Also assume the network width obeys
\vs\[
k\gtrsim \text{poly}(\nt,\blaz^{-1},\eps^{-1}).
\]
for a tolerance level $1>\eps>0$ and the size of the validation data obeys $\nv\gtrsim \ordet{\h}$. Following the aforementioned bilevel optimization scheme with a proper $\eta>0$ choice and any choice of number of iterations obeying $T\gtrsim \ordet{\frac{\nt}{\blaz}\log(\eps^{-1})}$, the classification error (0-1 loss) on the data distribution $\Dc$ obeys
\vs\[
\Lcz(\ft_\bah)\leq  \min_{\bal\in\Bal}2B\sqrt{\frac{\yT^T\Kbb_\bal^{-1}\yT}{\nt}}+C\sqrt{\frac{\ordet{\h}+t}{\nv}}+\eps+\delta,
\]
with probability at least $1-4(e^{-t}+\nt^{-3}+e^{-10h})$ (over the randomness in $\W_0,\Tc,\Vc$). Here, $\yT=[y_1~y_2~\dots~y_{\nt}]$. On the same event, for all $\bal\in\Bal$, the training classification error obeys $\Lczh_{\Tc}(\ft_\bal)\leq \eps$.
\end{theorem}
For a fixed $\bal$, the norm-based excess risk term $\sqrt{\frac{\y^T\Kbb_\bal^{-1}\y}{\nt}}$ quantifies the alignment between the kernel and the labeling function (which is small when $\yT$ lies on the principal eigenspace of $\Kb_\bal$). This generalization bound is akin to expressions that arise in norm-based NTK generalization arguments such as \cite{arora2019fine}. Critically, however, going beyond a fixed $\bal$, our theorem establishes this for all activations uniformly to conclude that the minimizer of the validation error also achieves minimal excess risk. The final statement of the theorem shows that the training error is arbitrarily small (essentially zero as $T\rightarrow \infty$) over all activations uniformly. Together, these results formally establish the pictorial illustration in Figures \ref{overview fig}(a) \& (b).

The proof strategy has two novelties with respect to standard NTK arguments. First, it requires a subtle uniform convergence argument on top of the NTK analysis to show that certain favorable properties that are essential to the NTK proof hold \emph{uniformly} for all activations (i.e.~choices of the hyperparameters) simultaneously with the same random initialization $\W_0$. Second, since neural nets may not obey Assumption \ref{algo lip}, to be able to apply our generalization bounds we need to construct a uniform Lipschitz approximation via its corresponding linearized feature map ($\fln{\bal}(\x)=\x^T\phi_\bal(\x)$) and bound the neural net's risk over train-validation procedure in terms of this proxy. This uniform approximation is in contrast to pointwise approximation results of \cite{arora2019exact}. Sec.~\ref{sec deep} provides further insights on our strategy. Finally, we mention a few possible extensions. One can (i) use logistic loss in the algorithm $\Ac$ rather than least-squares loss --which is often more suitable for classification tasks-- \cite{zou2020gradient} or (ii) use regularization techniques such as early stopping or ridge penalization \cite{li2020gradient} or (iii) refine the excess risk term using further technical arguments such as local Rademacher complexity \cite{bartlett2005local}.

\vs\vs
\section{Extension to NAS for Deep Architectures}\label{sec deep}

In this section, we provide further discussion extending our results in Sec.~\ref{sec:NAS} to multi-layer networks and general NAS beyond simple activation search. Our intention is to provide the core building blocks for an NTK-based NAS argument over a continuous architecture space $\Bal$. Recall the neural feature map introduced in Definition \ref{def ntk} where $\fF{\bal}(\cdot,\bt)$ can be any architecture induced by hyperparameters $\bal$. For instance, in DARTS, the architecture is a directed acyclic graph where each node $\x^{(i)}$ is a latent representation of the raw input $\x$ and $\bal$ dictates the operations $\sigma^{(i,j)}$ on the edges that transform $\x^{(i)}$ to obtain $\x^{(j)}$.

Recall the matrices $\Kb_\bal,\Kbh_\bal\in\R^{\nt\times \nt}$ from Definition \ref{def ntk} which are constructed from the neural feature map $\phi_\bal(\x):=\frac{\pa \fF{\bal}(\x)}{\pa \bt_0}$. For infinite-width networks (with proper initialization), NTK perfectly governs the training and generalization dynamics and the architectural hyperparameters $\bal$ controls the NTK kernel matrix $\Kb_\bal\in\R^{\nt\times \nt}$ (associated to the training set $\Tc$). A critical challenge for the general architectures is that the relation between the NTK kernel $\Kb_\bal$ and $\bal$ can be highly nonlinear. Here, we introduce a generic result for NTK-based generalization where we assume that $\Kb_\bal$ is possibly nonlinear but Lipschitz function of $\bal$. Recall that Theorem \ref{thm val} achieves logarithmic dependency on the Lipschitz constant thus unless the Lipschitz constant is extremely large, good generalization bounds are achievable. Our arguments will utilize this fact.

\begin{assumption}[Lipschitz Kernel] \label{lip ker}$\Kb_\bal,\Kbh_\bal\in\R^{n\times n}$ are $L$-Lipschitz functions of $\bal$ in spectral norm.
\end{assumption}\vs

To be able to establish generalization bounds for learning with generic neural feature maps in connection to NTK (Thm~\ref{gen thm generic} below) we need to ensure that wide networks converge to their infinite-width counterparts uniformly over $\Bal$. Let $\wi$ be a width parameter associated with the network. For instance, for a fully-connected network, $\wi$ can be set to the minimum number of hidden units across all layers. Similar to random features, it is known that, the network at random initialization converges to its infinite width counterpart exponentially fast. We now formalize this assumption.
\begin{assumption}[Neural Feature Concentration] \label{nn conc}
Recall Def.~\ref{def ntk}. There exists a width parameter $\wi>0$ and scalar $\kz>0$ such that, for any fixed $\bal\in\Bal$, at initialization $\bt_0\sim\Dci$, we have
\vs\begin{align}
\Pro\Big\{\|\Kbh_\bal-\Kb_\bal\|\geq \sqrt{\kz t/\wi}\Big\}\geq \e^{-t}.\label{K conc}
\end{align}
\end{assumption}\vs
For deep ResNets, fully-connected deep nets and DARTS architecture space (with zero, skip, conv operations),  this assumption holds with proper choice of $\kz>0$ (cf.~Theorem E.1 of \cite{du2019gradient} and Lemma 22 of \cite{zhou2020theory} which sets $\kz\propto n^2$).

Assumptions \ref{lip ker} and \ref{nn conc}, allows us to establish uniform convergence to the NTK over a continuous architecture space $\Bal$. Specifically, given tolerance $\eps>0$, for $\wi\gtrsim \ordet{\eps^{-2}\kz\deff\log(L)}$, with high probability (over initialization), $\|\Kb_\bal-\Kbh_\bal\|\leq \eps$ holds for all $\bal\in\Bal$ uniformly (cf.~Lemma \ref{lem sup simple2}).

Our generalization result for generic architectures is provided below and establishes a bound similar to Theorem \ref{one layer nas supp} by training a linearized model with neural feature maps. However, unlike Theorem \ref{one layer nas supp}, here we employ a small ridge regularization to promote Lipschitzness which helps us circumvent Assumption \ref{ntk assump}. 
\begin{theorem} \label{gen thm generic}Suppose Assumptions \ref{lip ker} and \ref{nn conc} hold and for all $\bal$ and some $B>0$ neural feature maps obey $\tn{\frac{\pa \fF{\bal}(\x)}{\pa \bt_0}}^2\leq B$ almost surely. We solve feature map regression (Def.~\ref{fmap def}) with neural feature maps and fixed ridge penalty $\la$. Fix some eigen-cutoff $\laz>0$ and tolerance $\eps>0$. Set $0<\la\leq \frac{\eps\laz^2}{4\sqrt{B\nt}}$ and $\defz=\ordet{\deff\log(\la^{-2}L\nt^3)}$. Finally define the $\la_0$-positive set
\vs\begin{align*}
\Bal_0=\{\bal\in\Bal\bgl \Kb_\bal\geq \laz\}.
\end{align*}
Also assume $\wi\gtrsim \ordet{\eps^{-4}\la_0^{-4}\kz^2\deff\log(L)}$ and $\nv\gtrsim\defz$. Finally, set $\ell$ in \eqref{val loss} to be the hinge loss. Then, with probability at least $1-5e^{-t}$, for some constant $C>0$, the binary classification error obeys
\vs\begin{align}\nn%\label{bound for cond}
\Lcz(\ft_\bah)\leq \min_{\bal\in\Bal_0}2\sqrt{\frac{B\y^T\Kb_\bal^{-1}\y}{\nt}}+C\sqrt{\frac{{\defz+t}}{{\nv}}}+\eps+\del.
\end{align}
\end{theorem}\vs\vs
Here the set $\Bal_0$ is the set of architectures that are favorable in the sense that the NTK associated with their infinite-width network can rather quickly fit the data as its kernel matrix has a large minimum eigenvalue. This also ensures that the neural nets trained on $\Tc$ are Lipschitz with respect to $\bal$ over $\Bal_0$. In essence, the result states that the bilevel procedure \eqref{opt alpha} is guaranteed to return a model at least as good as the best model over $\Bal_0$. Importantly, one can enlarge the set $\Bal_0$ by reducing the penalty $\la$ at the cost of a larger $\defz$ term which grows logarithmically in $1/\la$. The above theorem also relates to optimal kernel selection using validation data which connects NAS to the multiple kernel learning (MKL) \cite{gonen2011multiple} problem. However in the MKL literature, the kernel $\Kb_\bal$ is typically a linear function of $\bal$ whereas in NAS the dependence is more involved and nonconvex, especially for realistic search spaces. 

\vs\noindent\textbf{Deep activation search.} As a concrete example, we briefly pause to demonstrate that the above assumption holds for a deep multilayer activation search problem. In this case we will show that $L$ is at most exponential in depth $D$. To be precise, following Section \ref{sec:NAS}, fix a pre-defined set of activations $\{\sigma_j\}_{j=1}^h$. For a feedforward depth $D$ network, set $\bal\in\R^{D h}$ to be the concatenation of $D$ subvectors $\{\bl{i}\}_{i=1}^D\subset\R^h$. The layer $\ell$ activation is then given via 
\[
\sigma_{\bl{i}}(\cdot)=\sum_{j=1}^\h \bl{i}_j\sigma_j(\cdot).
\]
Now, given input $\x=\xx{0}$ and weight matrices $\bt=\{\Ww{i}\}_{i=1}^{D+1}$, define the corresponding feedforward network $\fF{\bal}(\x,\bt):\R^d\rightarrow\R=\Ww{D+1}\xx{D}$ where the hidden features $\xx{i}$ are defined as
\[
\xx{i}=\sigma_{\bl{i}}(\Ww{i} \xx{i-1})\quad\text{for}\quad 1\leq i\leq D.
\]
The lemma below shows that in this case the Lipschitzness $L$ with respect to the hyperparameters is at most exponential in $D$. The result is stated for a fairly flexible random Gaussian initialization. The precise statement is deferred to Lemma \ref{lem rand lip bound}.
\begin{lemma} \label{lemma deep act} Suppose for all $1\leq i\leq h$, $|\phi_i(0)|,|\phi_i'(x)|,|\phi_i''(x)|$ are bounded. Input features are normalized to ensure $\tn{\x}\lesssim\sqrt{d}$. Let layer $i$ have $k_i$ neurons and $\W^{(i)}\in\R^{k_i\times k_{i-1}}$. Suppose the aspect ratios $k_i/k_{i-1}$ are bounded for layers $i\ge 2$. For the first layer, denote $k_0=d$ and $k_1=k$. Each layer $\ell$ is initialized with i.i.d.~$\Nn(0,c_\ell)$ entries satisfying
\[
c_\ell\leq\begin{cases}\cc\quad\text{if}\quad i=1\\\cc/k_{i-1}\quad\text{if}\quad \ell\geq 2\end{cases}\quad\text{for some constant}\quad\cc>0.
\] 
 Then, Assumption \ref{lip ker} holds (with high probability for $\Kbh_\bal$) with $\log(L)\lesssim D+\log(k+d+\nt)$.
\end{lemma}
We remark that this initialization scheme corresponds to Xavier/He initializations ($1/\text{fan$\_$in}$ variance) as well as our setting in Theorem \ref{one layer nas supp}. We suspect that the exponential dependence on $D$ can be refined by enforcing normalization schemes during initialization to ensure that hidden features don't grow exponentially with depth. Recall that sample complexity in Theorem \ref{thm val} depends logarithmically on $L$, which grows at most linearly in $D$ up to log factors. Furthermore, as stated earlier Assumption \ref{nn conc} is known to hold in this setting as well (cf.~discussion above). Thus for a depth $D$ network, using the above Lipschitzness bound, Theorem \ref{gen thm generic} allows for good generalization with a validation sample complexity of $\nv\propto \ordet{\deff\log(L)}=\ordet{\DD\times\deff}$. \clr{Finally, note that $\log(L)$ also exhibits a logarithmic dependence on $k$. Thus as long as network width is not exponentially large (which holds for all practical networks), our excess risk bounds remain small leading to meaningful generalization guarantees. We do remark that results can also be extended to infinite-width networks (which are of interest due to NTK). Here, the key idea is constructing a Lipschitz approximation to the infinite-width problem via a finite-width problem. In similar spirit to Lemma \ref{nn conc}, such approximation can be made arbitrarily accurate with a polynomial choice of $k$ \cite{zhou2020theory,arora2019exact}. As discussed earlier, Theorem \ref{thm val2} in the appendix provides a clear path to fully formalize this and the proof of Theorem \ref{one layer nas supp} already employs such a Lipschitz approximation scheme to circumvent imperfect Lipschitzness of nonlinear neural net training.}

\vs\vs
\section{Algorithmic Guarantees via Connection to Low-rank Matrix Learning}\label{sec rank 1 factor}
The results stated so far focus on generalization and are not fully algorithmic in nature in the sense that they assume access to an approximately optimal hyperparameter of the upper-level problem per \eqref{opt alpha} based on the validation data. In this section we wish to investigate whether it is possible to provably find such an approximate solution with a few validation samples and a computationally tractable algorithm. To this aim, in this section, we establish algorithmic connections between our activation/feature-map search problems of Section \ref{sec:FMAP} to a rank-1 matrix learning problem. In Def.~\ref{fmap def} --instead of studying $\bPhi_\bal$ given $\bal$-- let us consider the matrix of feature maps 
\vs\[
\X=[\phi_1(\x)~\phi_2(\x)~\dots~\phi_\h(\x)]^T\in\R^{\h\times p}
\] for a given input $\x$. Then, the population squared-loss risk of a $(\bal,\bt)$ pair predicting $\bt^T\phi_\bal(\x)$ is given by 
\vs\[
\Lc(\bal,\bt):=\E[(y-\bal^T\X\bt)^2]=\E[(y-\langle\X,\bal\bt^T\rangle)^2].
\] Thus, the search for the optimal model pair $(\bas,\bts)$ is simply a rank-1 matrix learning task with $\M_\st=\bas\bts^T$. Can we learn the right matrix with a tractable algorithm in the regime \eqref{roi eq}?

This is a rather subtle question as in the regime \eqref{roi eq} there is not enough samples to reconstruct $\M_\st$ as anything algorithm regardless of computational tractability requires $\nt+\nv\gtrsim p+h$! But this of course does not rule out the possibility of finding an approximately optimal hyperparameter close to $\bas$.  To answer this --rather tricky question-- we study a discriminative data model commonly used for modeling low-rank learning. Consider a realizable setup $y=\bas^T\X\bts$ where we ignore noise for ease of exposition, see supplementary for details. We also assume that the feature matrix $\X$ has i.i.d.~$\Nn(0,1)$ entries. Suppose we have $\Tc=(y_i,{\X}_i)_{i=1}^{\nt=n},\Vc=(\yt_i,\tilde{\X}_i)_{i=1}^{\nv=n}$ datasets with equal sample split $n=\nt=\nv$. If we combine these datasets into $\Tc$ and solve ERM, when $2n\leq p$, for \emph{any choice of} $\bal$, weights $\bt\in\R^p$ can perfectly fit the labels. Instead, we propose the following two-stage algorithm to achieve a near-optimal learning guarantee. Set $\hat{\M}=\sum_{i=1}^n \yt_i\tilde{\X}_i$.
\begin{align} 
&1.~\textbf{Spectral estimator on $\Vc$}:~~~~\text{Set}~\bah=\text{top\_eigen\_vec}(\hat{\M}\hat{\M}^T).\label{spect est}\\
&2.~\textbf{Solve~ERM on $\Tc$}:~~~~~~~~~~~~~~~~\text{Set}~\bth=\arg\min_{\bt}\sum_{i=1}^n (y_i-\bah^T\X_i\bt)^2\label{erm prob}.
\end{align}
We have the following guarantee for this procedure.

\vs\begin{theorem}[Low-rank learning with $p>n$]\label{thm low-rank} Let $(\X_i,\tilde{\X}_i)_{i=1}^n$ be i.i.d.~matrices with i.i.d.~$\Nn(0,1)$ entries. Let $y_i=\bas^T\X_i\bts$ for unit norm $\bal\in\R^\h,\bt\in\R^p$. Consider an asymptotic setting where $p,n,\h$ grow to infinity with fixed ratios given by $\bp=p/n>1$, $\bh=\h/n<1$ and consider the asymptotic performance of $(\bah,\bth)$. 

Let $1\geq \rho_{\bas,\bah}\geq 0$ be the absolute correlation between $\bah,\bal$ i.e. $\rho_{\bas,\bah}=|\bal^T\bah|$. Suppose $\bp\bh\leq 1/6$. We have that
\vs\begin{align}
&\lim_{n\rightarrow\infty}\rho^2_{\bas,\bah}\geq 1-64\bp\bh\label{high corr}
\end{align}
Additionally, if $\bp\bh\leq c$ for sufficiently small constant $c>0$,
\vs\begin{align}
\lim_{n\rightarrow\infty}\Lc(\bah,\bth)\leq \underbrace{1-\frac{1}{\bp}}_{\text{risk}(\bas)}+\underbrace{\frac{200\bh}{1-1/\bp}}_{\text{estimating}~\bas}.\label{high dim linear}
\end{align}
\end{theorem}\vs
A few remarks are in order. First, the result applies in the regime $p\gg n$ as long as --the rather surprising condition-- $ph\lesssim{n^2}$ holds (see \eqref{high corr}). Numerical experiments in Section \ref{sec:exp} verify that (specifically Figure \ref{fig both}) this condition is indeed necessary. Here, $\text{risk}(\bas)=1-n/p$ is the \emph{exact asymptotic risk} one would achieve by solving ERM with the knowledge of optimal $\bas$. Our result shows that one can approximately recover this optimal $\bas$ up to an error that scales with $ph/n^2$. Our second result achieves a near-optimal risk via $\bah$ without knowing $\bas$. Since $1-1/\bp$ is essentially constant, the risk due to $\bas$-search is proportional to $\bh=h/n$. This rate is consistent with Theorem \ref{thm val} which would achieve a risk of $1-n/p+\ordet{\sqrt{h/n}}$. Remarkably, we obtain a slightly more refined rate ($h/n\leq \sqrt{h/n}$) using a spectral estimator with a completely different mathematical machinery based on high-dimensional learning. To the best of our knowledge, our spectral estimation result \eqref{high corr} in the $p>n$ regime is first of its kind (with a surprising $ph\lesssim{n^2}$ condition) and might be of independent interest. Finally, while this result already provides valuable algorithmic insights, it would be desirable to extend this result to general feature distributions to establish algorithmic guarantees for the original activation/feature map search problems.

\begin{figure}\centering
	\subfigure[\normalsize{Input layer stability for a one-hidden layer network}] { \label{fig:input_distance}
		\begin{tikzpicture}
			\node at (0,0) [scale=0.45]{\includegraphics{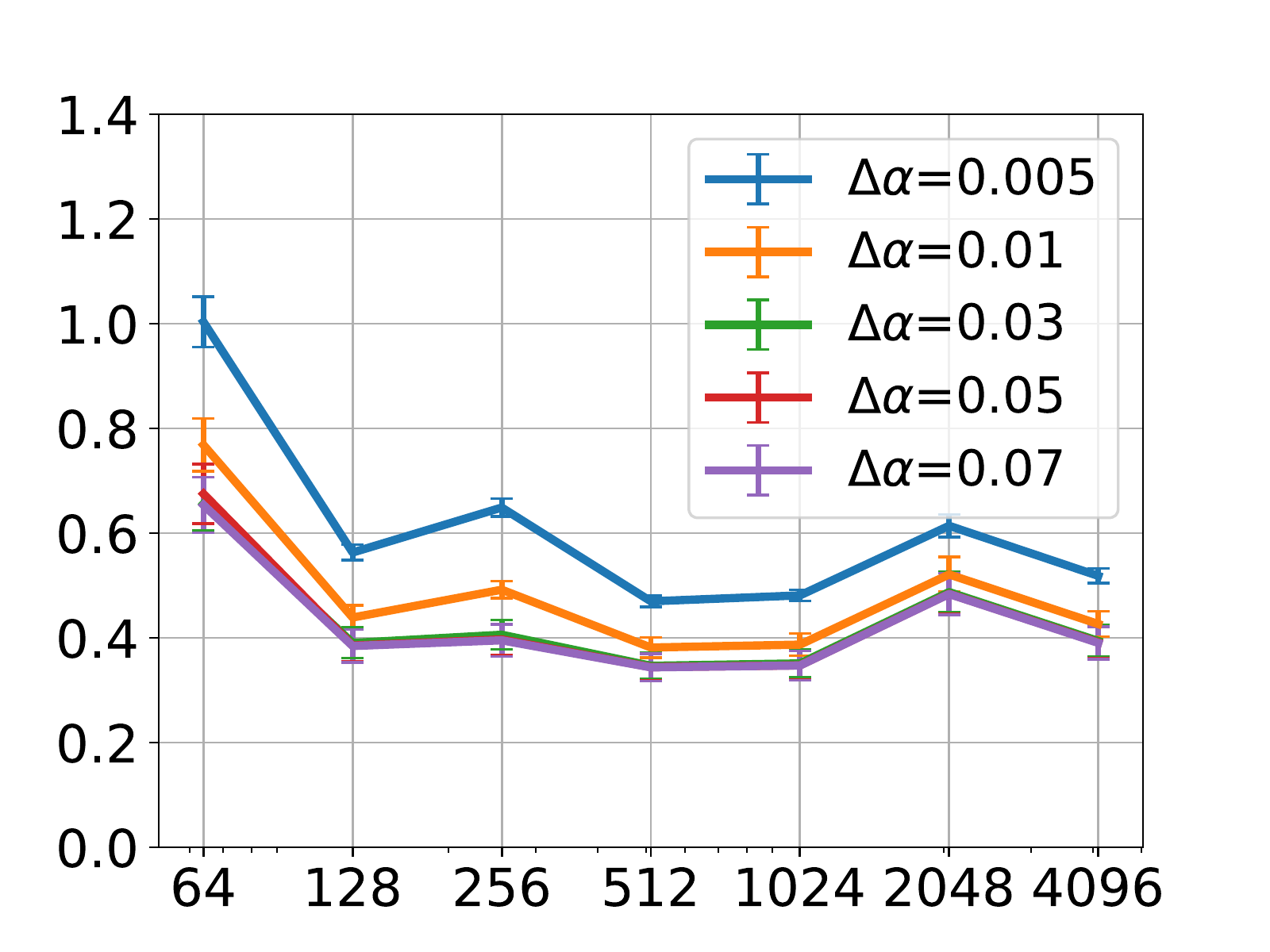}};
			\node at (0,-2.8) [scale=0.7] {\Large{Model width $k$}};
			\node at (-3.65,0) [rotate=90,scale=0.7] {\Large{Normalized distance}};
		\end{tikzpicture}
	}\hspace{2pt}
	\subfigure[\normalsize{Stability of weights of a deeper four layer network}] { \label{fig:deep_distance}
		\begin{tikzpicture}
			\node at (0,0) [scale=0.45]{\includegraphics{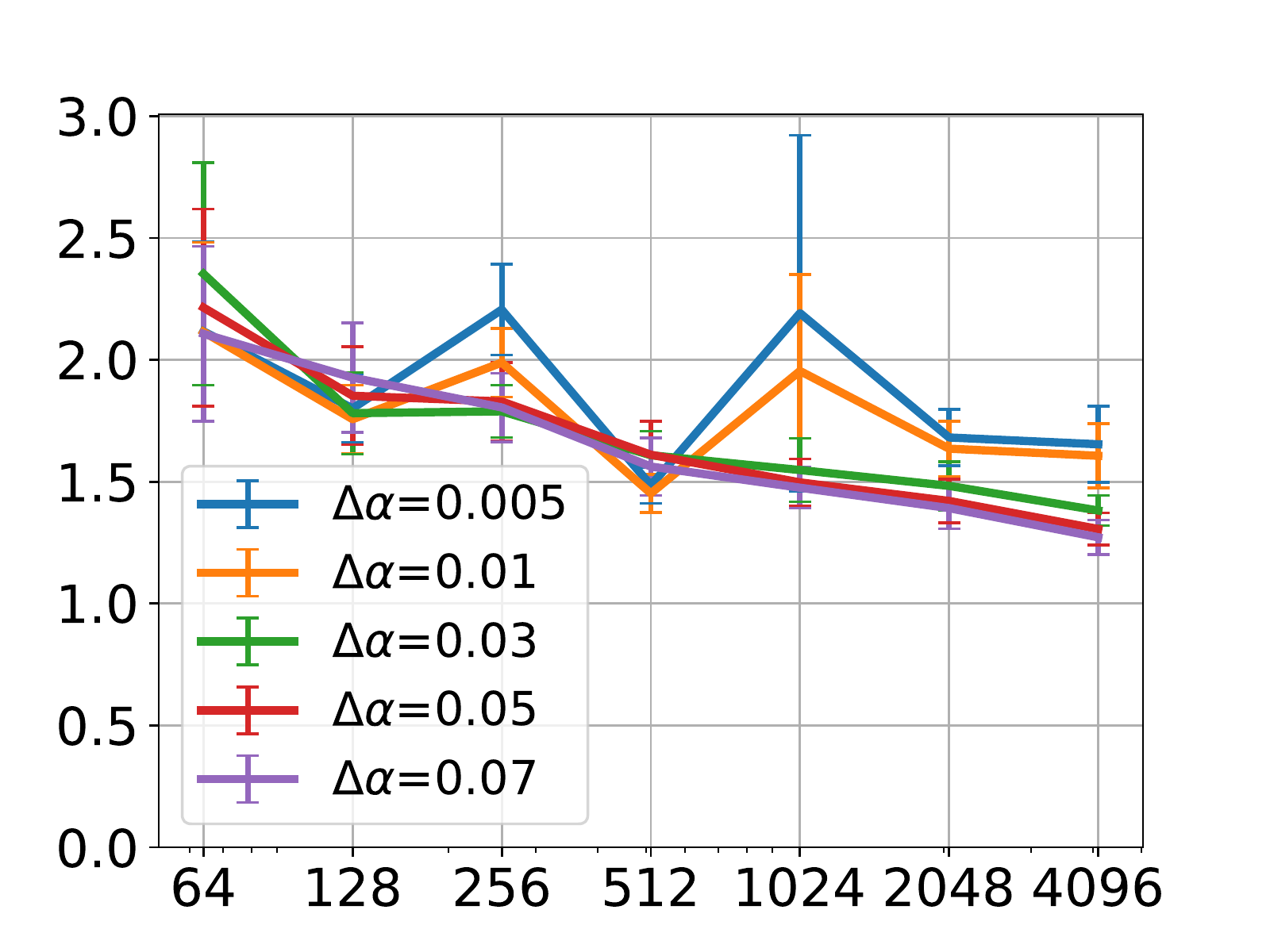}};
			\node at (0,-2.8) [scale=0.7] {\Large{Model width $k$}};
			\node at (-3.65,0) [rotate=90,scale=0.7] {\Large{Normalized distance}};
		\end{tikzpicture}% in Assumption \ref{algo lip}
	}\vspace{-8pt}\caption{\small{We visualize the Lipschitzness of the algorithm when $\mathcal{A}(\cdot)$ is stochastic gradient descent. We train networks with activation parameters $\alpha$ and $\alpha+\Delta\alpha$ and display the normalized distances $\tn{\bt_\alpha-\bt_{\alpha+\Delta\alpha}}/\Delta\alpha$ for different perturbation strengths $\Delta\alpha$.}}\label{fig distance}\vspace{-13pt}
\end{figure}
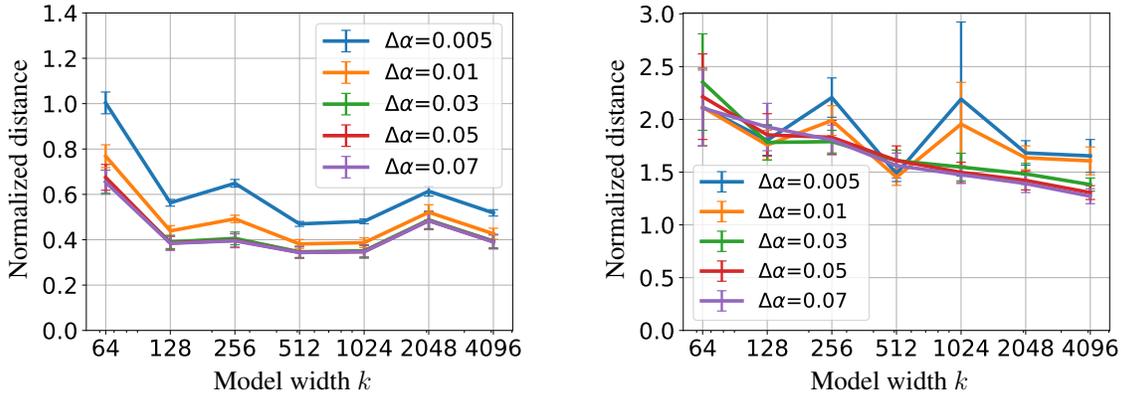

\section{Numerical Experiments}\label{sec:exp}
To verify our theory, we provide three sets of experiments. First, to test Theorem \ref{one layer nas supp}, we verify the (approximate) Lipschitzness of trained neural nets to perturbations in the activation function. Second, to test Theorem \ref{thm val}, we will study the test-validation gap for DARTS search space. Finally, we verify our claims on 

\vs\noindent\textbf{a.~~Lipschitzness of Trained Networks.}
First, we wish to verify Assumption \ref{algo lip} for neural nets by demonstrating their Lipschitzness under proper conditions. In these experiments, we consider a single hyperparameter $\alpha\in\R$ to control the activation via a combination of ReLU and Sigmoid i.e.~$\sigma_\alpha(x)=(1-\alpha)\text{ReLU}(x)+\alpha\cdot\text{Sigmoid}(x)$. Training the network weights $\bt$ with this activation from the same random initialization leads to the weights $\bt_\alpha$. We are interested in testing the stability of these weights to slight $\alpha$ perturbations by studying the normalized distance $\tn{\bt_\alpha-\bt_{\alpha+\Delta\alpha}}/\Delta\alpha$. This in turn ensures the Lipschitzness of the model output via a standard bound (see supp. for numerical experiments). Fig.~\ref{fig distance} presents our results on both shallow and deeper networks on a binary MNIST task which uses the first two classes with squared loss. This setup is in a similar spirit to our theory. In Fig.~\ref{fig:input_distance} we train input layer of a shallow network $f_\alpha(x)=\vb^T\sigma_\alpha(\W\X)$ where $\W\in\R^{k\times 784}$. In Fig.~\ref{fig:deep_distance}, a deeper fully connected network with 4 layers is trained. Here, the number of neurons from input to output are $k$, $k/2$, $k/4$ and $1$ and the same activation $\sigma_\alpha(\X)$ is used for all layers. Finally, we initialize the network with He initialization and train the model for 60 epochs with batch size 128 with SGD optimizer and learning rate 0.003. For each curve and width level, we average 20 experiments where we first pick 20 random $\alpha \in [0,1]$ and their perturbation $\alpha+\Delta\alpha$. We then compute the average of normalized distances $\tn{\bt_\alpha-\bt_{\alpha+\Delta\alpha}}/\Delta\alpha$.

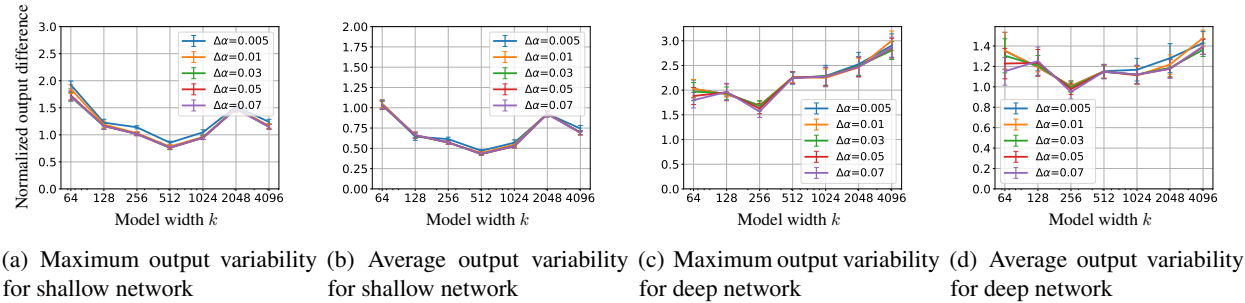
\begin{figure}\centering
	
	\subfigure[{Maximum output variability for shallow network}] { \label{fig:shallow_output_max}
		\begin{tikzpicture}
			\node at (0,0) [scale=0.23]{\includegraphics{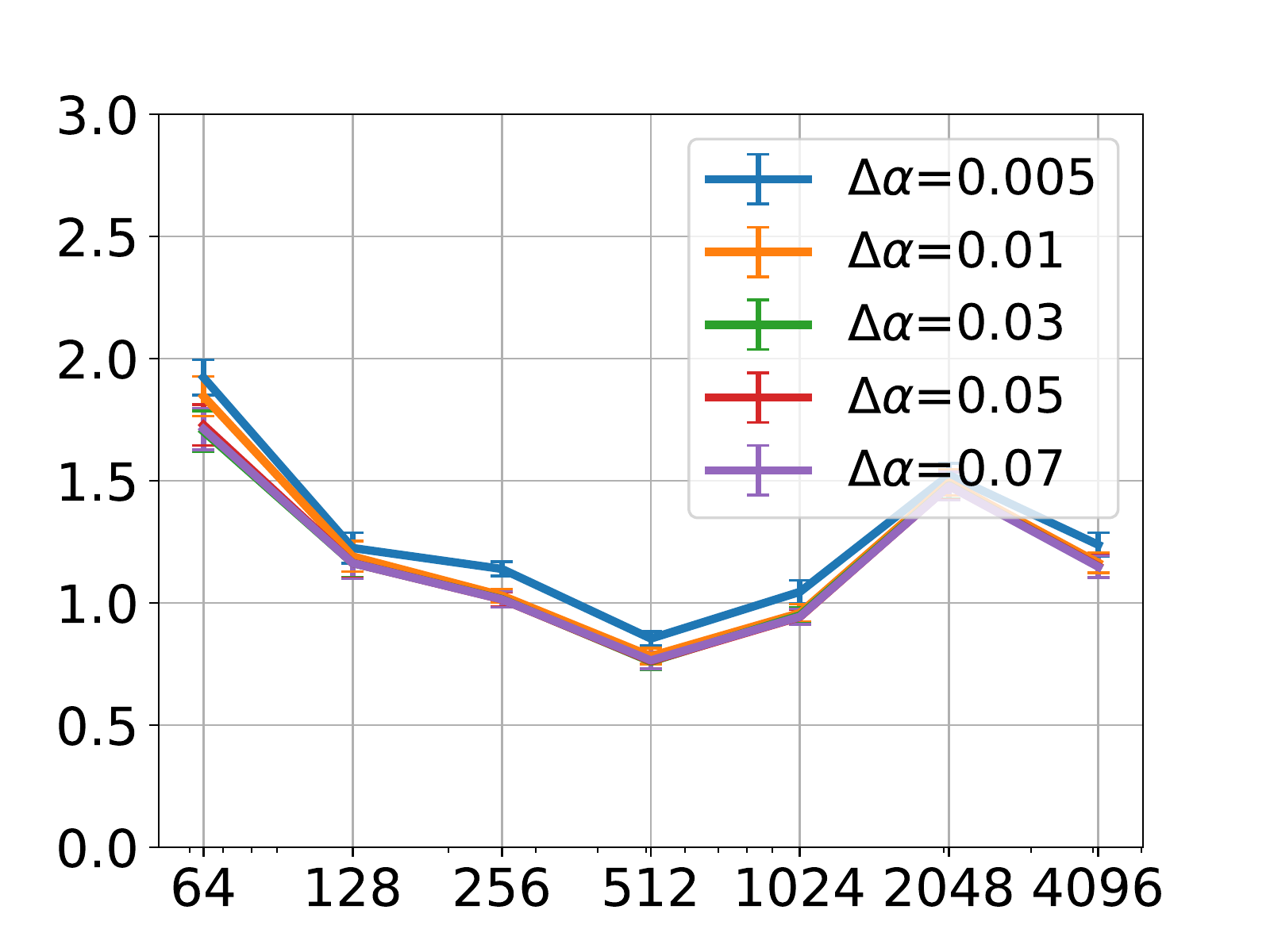}};
			\node at (0,-1.5) [scale=0.6] {Model width $k$};
			\node at (-1.9,0) [rotate=90,scale=0.6] {Normalized output difference};
		\end{tikzpicture}
	}\hspace{1.5pt}
	\subfigure[{Average output variability for shallow network}] { \label{fig:shallow_output_avg}
		\begin{tikzpicture}
			\node at (0,0) [scale=0.23]{\includegraphics{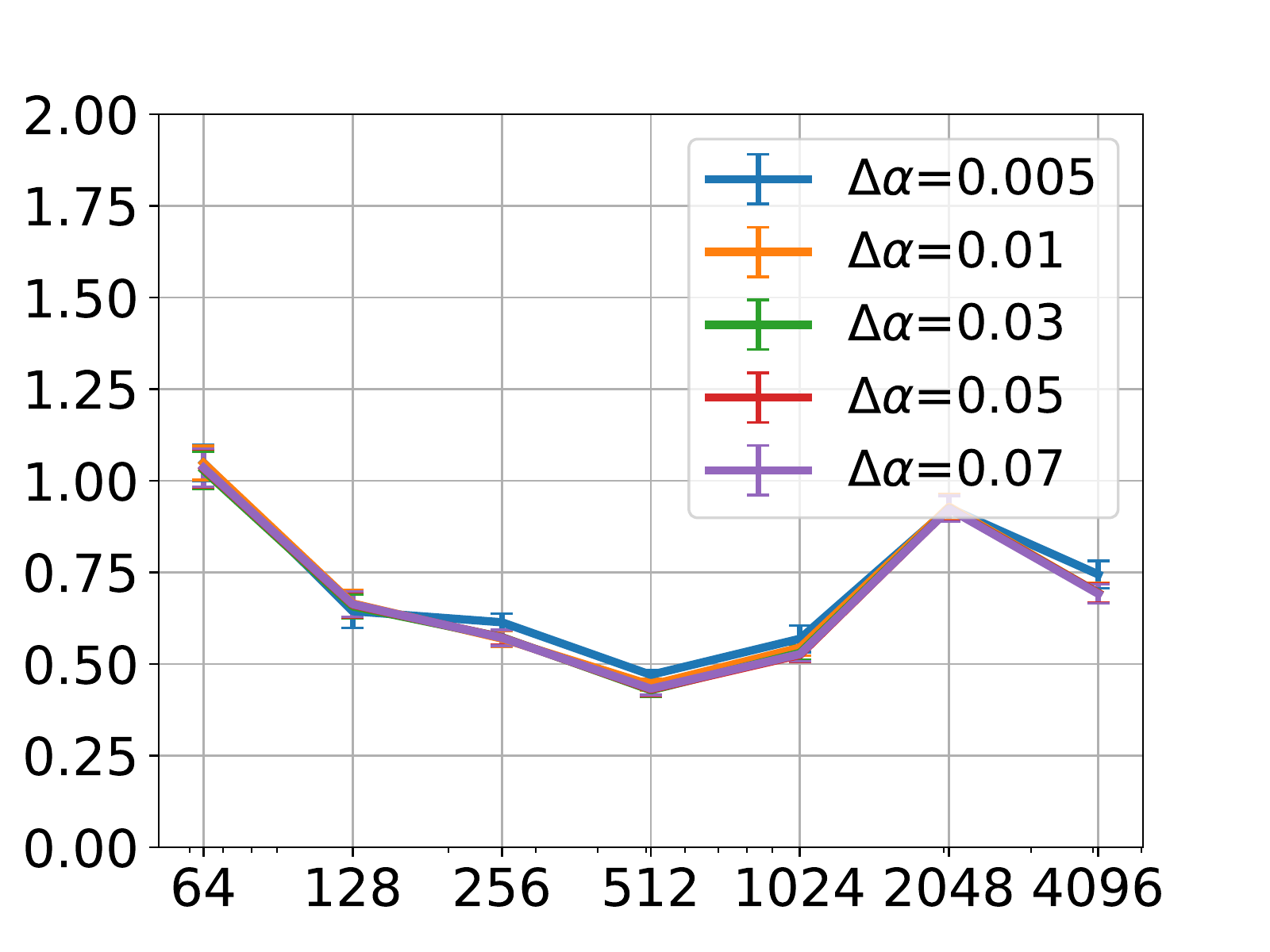}};
			\node at (0,-1.5) [scale=0.6] {Model width $k$};
		\end{tikzpicture}
	}\hspace{1.5pt}
	\subfigure[{Maximum output variability for deep network}] { \label{fig:deep_output_max}
		\begin{tikzpicture}
			\node at (0,0) [scale=0.23]{\includegraphics{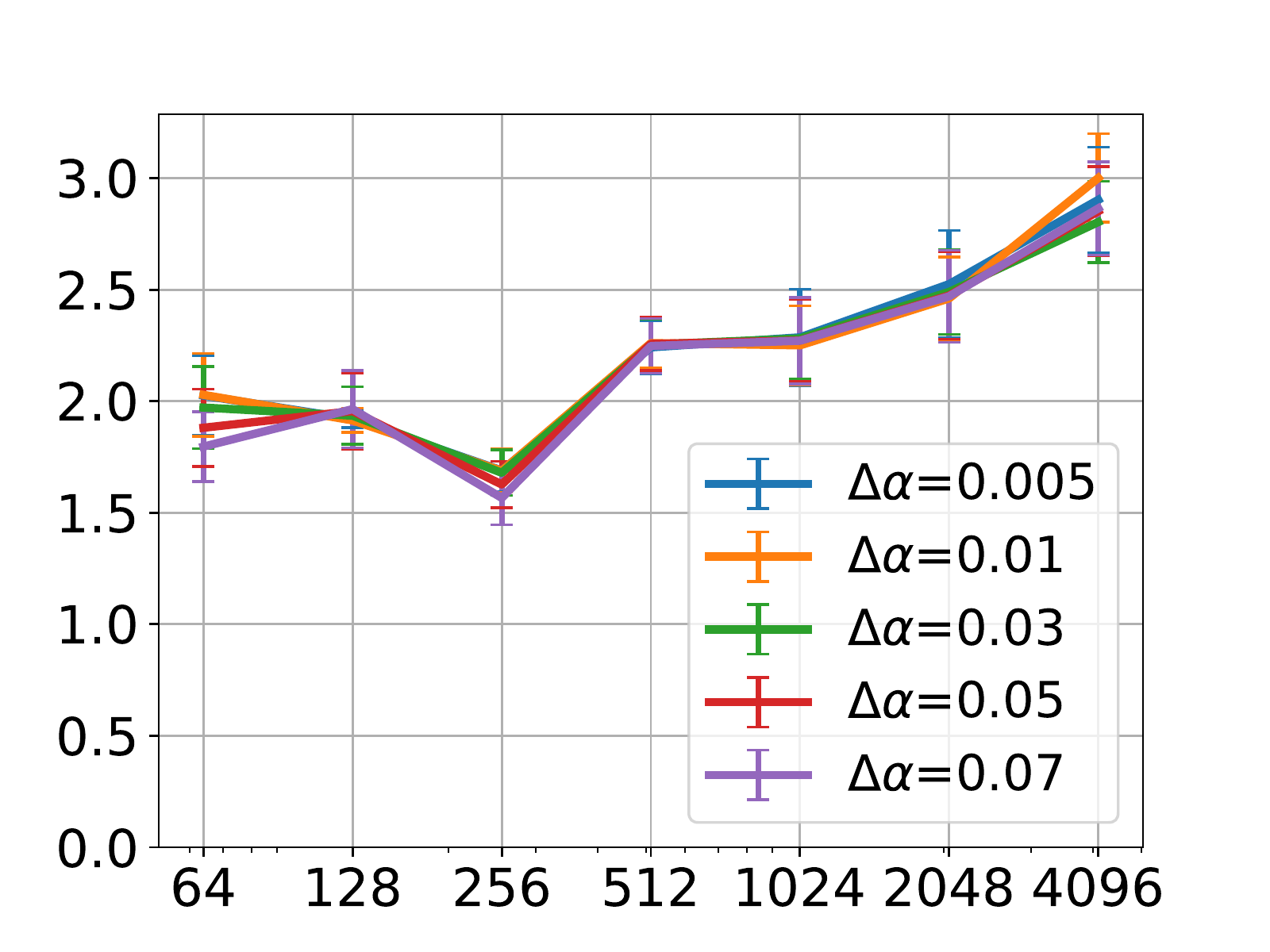}};
			\node at (0,-1.5) [scale=0.6] {Model width $k$};
		\end{tikzpicture}
	}\hspace{1.5pt}
	\subfigure[{Average output variability for deep network}] { \label{fig:deep_output_avg}
		\begin{tikzpicture}
			\node at (0,0) [scale=0.23]{\includegraphics{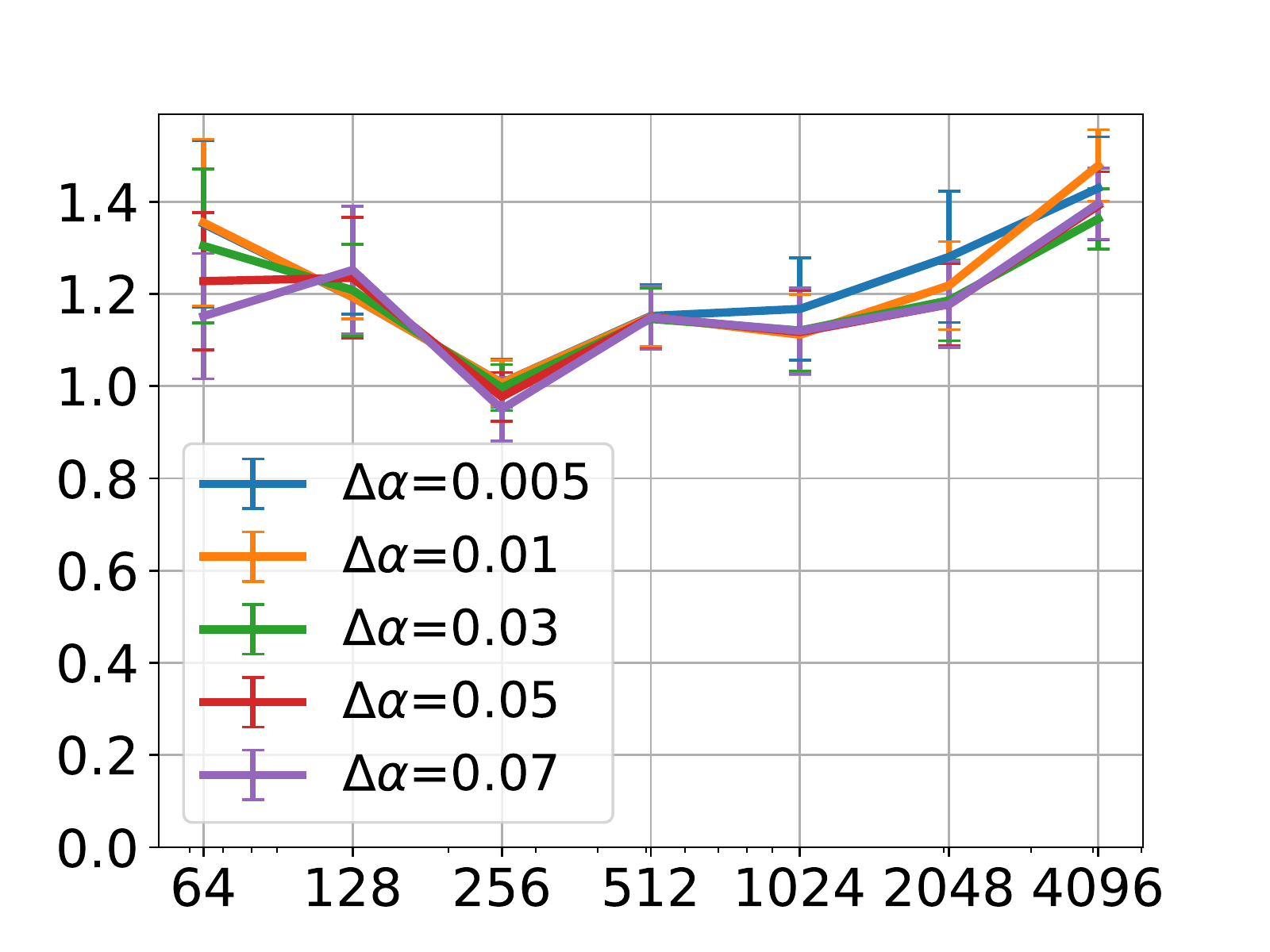}};
			\node at (0,-1.5) [scale=0.6] {Model width $k$};
		\end{tikzpicture}}
		\caption{\small{We visualize the Lipschitzness of the algorithm when $\mathcal{A}(\cdot)$ is stochastic gradient descent. In Figure~\ref{fig:shallow_output_max}~and~\ref{fig:shallow_output_avg}, we train the input layer of 2-layer shallow networks with activation parameters $\alpha$ and $\alpha+\Delta\alpha$ which have the same setup as Figure~\ref{fig:input_distance}. Then we display the maximum and average output variability defined in \eqref{max first} and \eqref{avg output} respectively for different perturbation strengths $\Delta\alpha$ and model width $k$. In Figure~\ref{fig:deep_output_max}~and~\ref{fig:deep_output_avg}, we train 4-layer deep fully connected networks with activation parameters $\alpha$ and $\alpha+\Delta\alpha$ and also display the maximum and average output variability for different perturbation strengths $\Delta\alpha$. }}\label{fig output}
\end{figure}

We now provide further experiments to better verify Assumption~\ref{algo lip}. Let $\Ttc$ be the test data (of MNIST) which provides a proxy for the input domain $\Xc$~\footnote{We note that these values are calculated over the test data however we found the behavior over the training data to be similar.}. Our goal is to assess the Lipschitzness of the network prediction over $\Ttc$ which exactly corresponds to the setup of Assumption~\ref{algo lip}. Specifically, we will evaluate two quantities as a function of the activation perturbation $\Delta\alpha$
\begin{align}
&\text{Maximum output variability:}~\lit{f,\Delta\alpha,\Ttc}=\frac{\sup_{\x\in \Ttc}{|f(\bt_\alpha,\x)-f(\bt_{\alpha+\Delta\alpha},\x)|}}{\Delta\alpha},\label{max first}\\
&\text{Average output variability:}~\lia{f,\Delta\alpha,\Ttc}=\frac{1}{\nt\Delta\alpha}\sum_{\x\in\Ttc}|f(\bt_\alpha,\x)-f(\bt_{\alpha+\Delta\alpha},\x)|.\label{avg output}
\end{align}
Here, the maximum output variability is the most relevant quantity as it directly corresponds to the Lipschitzness of the function over the input domain. We keep the same setup in Figure~\ref{fig:shallow_output_max}~and~\ref{fig:shallow_output_avg} as Figure~\ref{fig:input_distance}, however, in Figure~\ref{fig:shallow_output_max}~and~\ref{fig:shallow_output_avg}, instead of computing the distance between trained models, we plot maximum output variability~\eqref{max first} and average output variability~\eqref{avg output} respectively for different perturbation level $\Delta_\alpha$ and model width. Figure~\ref{fig:deep_output_max}~and~\ref{fig:deep_output_avg} demonstrate the output variability on 4-layer neural networks which has the same setup as \ref{fig:deep_distance}.

All figures support our theory and show that, the normalized distance is indeed stable to the perturbation level $\Delta\alpha$ across different widths and only mildly changes. Note that $\Delta\alpha\in\{0.01,0.005\}$ result in a slightly larger normalized distance compared to larger perturbations. Such behavior for small $\Delta\alpha$ is not surprising and is likely due to the imperfect Lipschitzness of the network (especially with ReLU activation). Fortunately, our theory allows for this as it only requires an approximate Lipschitz property (recall the discussion below Theorem \ref{thm val}).

\begin{figure}\centering
	\subfigure[{The test-validation gap for the continuously parameterized architecture during the search phase of DARTS.}] { \label{fig:darts_gap}
		\begin{tikzpicture}
			\node at (0,0) [scale=0.45]{\includegraphics[width=1\textwidth,height=0.8\textwidth]{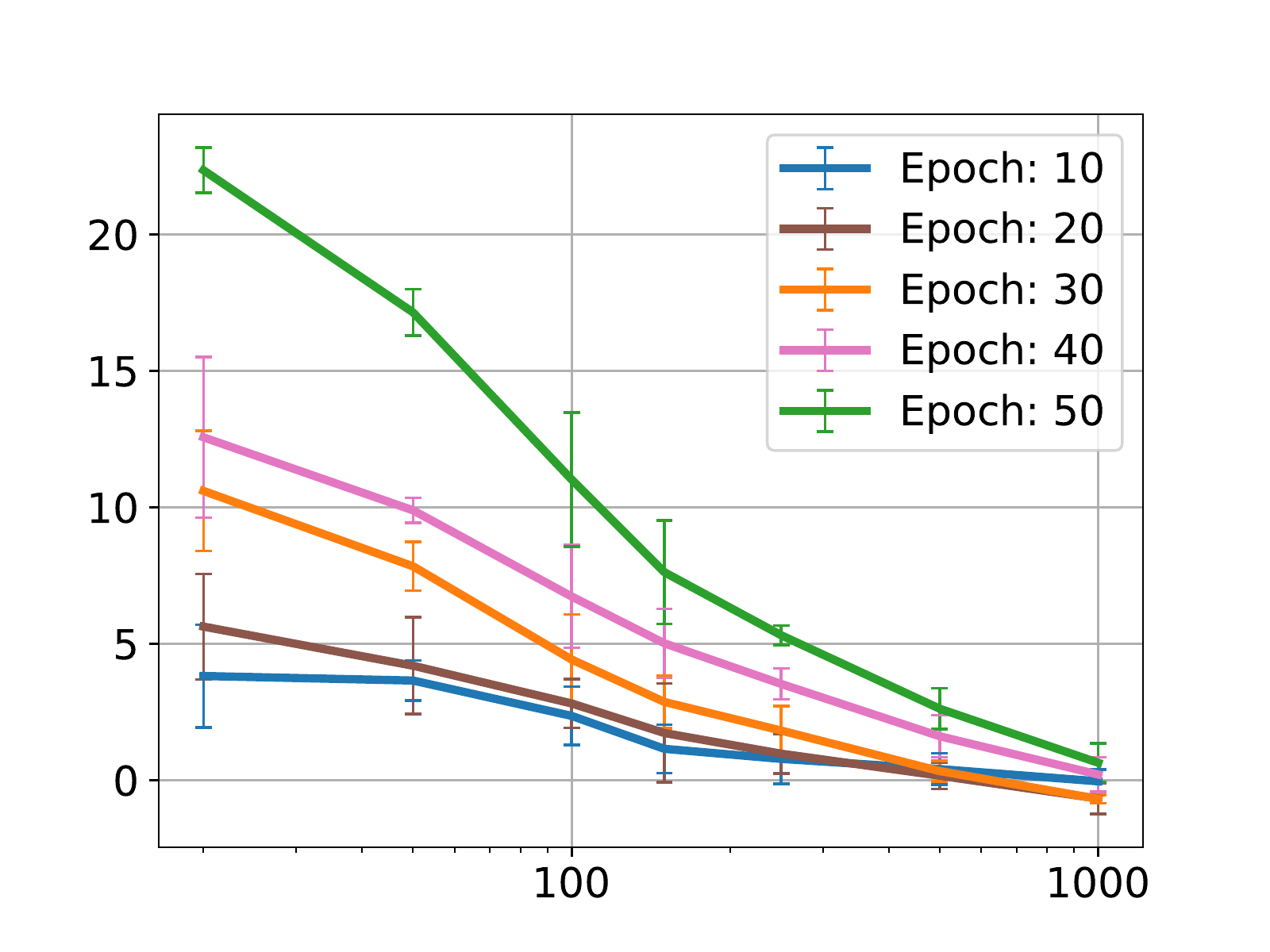}};
			\node at (0,-3.1) [scale=0.7] {\Large{Validation size}};
			\node at (-3.65,0) [rotate=90,scale=0.7] {\Large{Test-validation gap}};
		\end{tikzpicture}
	}\hspace{2pt}
	\subfigure[{\clr{The train (dotted curves), validation (dashed curves) and test (solid curves) errors during the search phase of DARTS training.}}] { \label{fig:darts_val_test_train}
		\begin{tikzpicture}
			\node at (0,0) [scale=0.45]{\includegraphics[width=1\textwidth,height=0.8\textwidth]{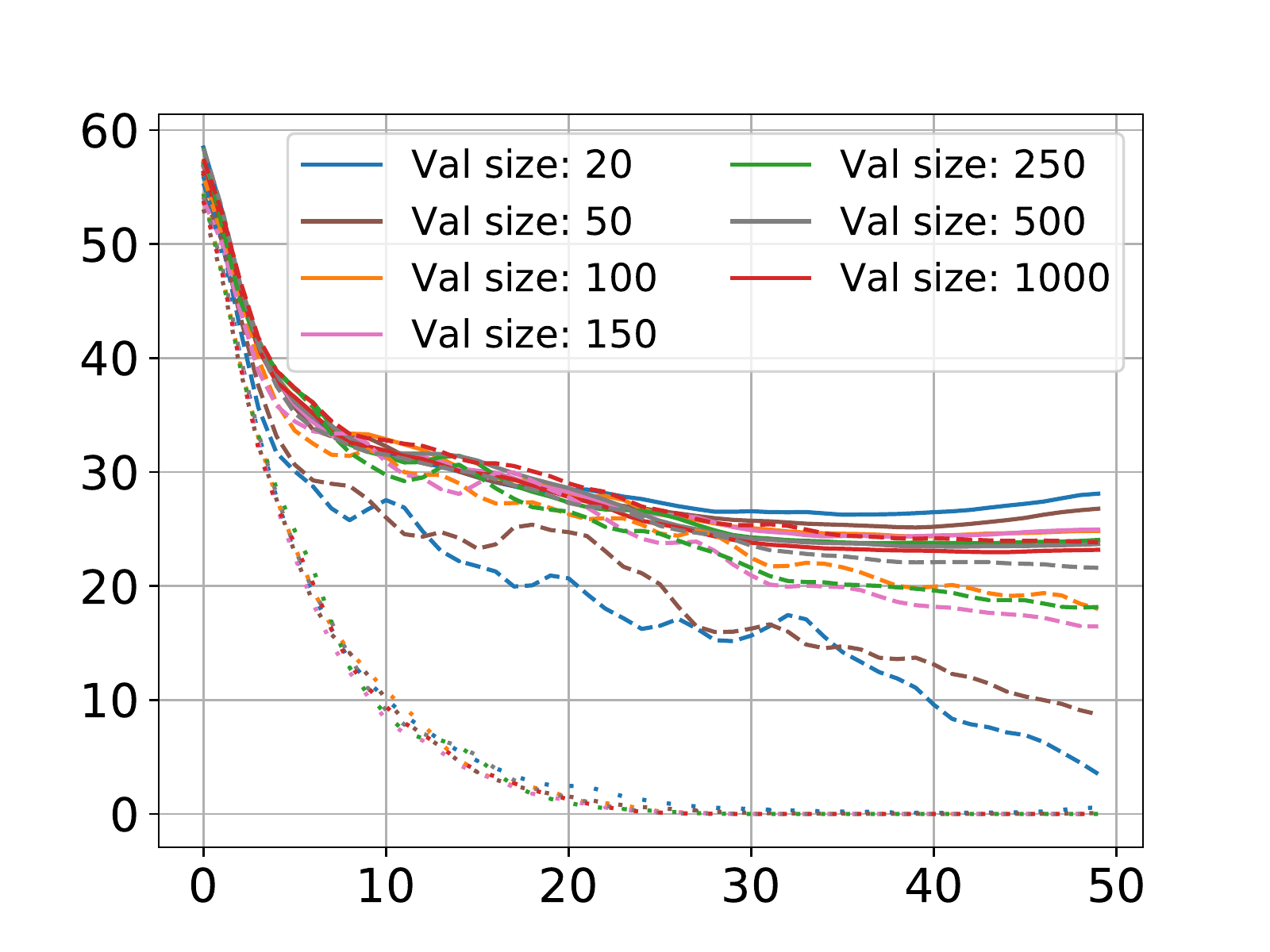}};
			\node at (0,-3.1) [scale=0.7] {\Large{Epoch}};
			\node at (-3.65,0) [rotate=90,scale=0.7] {\Large{Test / Val / Train errors}};
		\end{tikzpicture}% in Assumption \ref{algo lip}
	}\vspace{-8pt}\caption{\normalsize{The test-validation gap and errors for the continuously parameterized architecture during the search phase of DARTS. \clr{The result is average of 5 different runs.} Evaluations are for different validation sample sizes and epoch checkpoints. \clr{Figure~\ref{fig:darts_val_test_train} shows that all models overfit to training samples quickly. The DARTS runs with less validation examples also trend to overfit the validation dataset which leads to the increase in test error (most visibly for~$\nv\in \{20,50\}$).}}}\label{darts fig}\vspace{-13pt}
\end{figure}
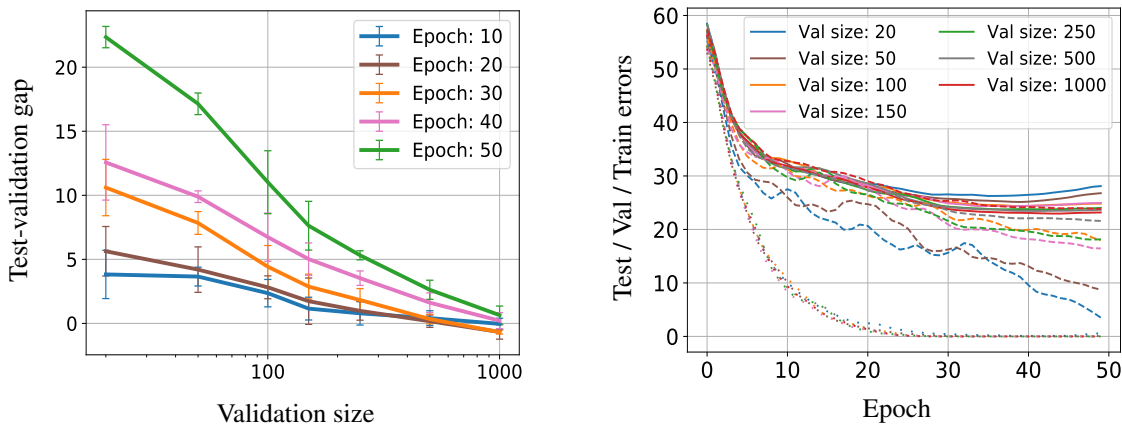

\vs\noindent\textbf{b.~~Test-Validation Gap for DARTS.} In this experiment, we study a realistic architecture search space via DARTS algorithm \cite{liu2018darts} over CIFAR-10 dataset using 10k training samples. We only consider the search phase of DARTS and train for 50 epochs using SGD. This phase outputs a \emph{continuously parameterized architecture}, which can be computed on DARTS' supernet. \clr{Each operation on the edges of the final architecture is a linear superposition of eight predefined operations (e.g.~conv3x3, zero, skip).} The curves are obtained by averaging three independent runs. In Fig.~\ref{darts fig}, we assess the gap between the test and validation errors while varying validation sizes from $20$ to $1000$. Our experiments reveal two key findings via Figure \ref{fig:darts_gap}. First, the train-validation gap indeed decreases rapidly as soon as the validation size is only mildly large, e.g.~around $\nv=250$ --much smaller than the typical validation size used in practice. This is consistent with Theorem \ref{thm val} as the architecture has 224 hyperparameters. On the other hand, there is indeed a potential of overfitting to validation for $\nv\leq 100$. We also observe that the gap noticeably increases with more epochs. The small gaps at initial epochs may be due to insufficient training i.e.~network does not yet achieve zero training loss. For later epochs, since early-stopping (i.e.~using earlier epoch checkpoints) has a ridge regularization effect, we suspect that widening gap may be due to the growing Lipschitz constant with respect to the architecture choice. Such behavior would be consistent with Thm \ref{gen thm generic} (smaller ridge penalty leads to more excess validation risk). \clr{Figure \ref{fig:darts_val_test_train} displays the train/validation/test errors by epoch for different validation sample sizes. This figure is also consistent with our core setup and expectations \eqref{roi eq}. The training loss/error quickly goes down to zero. Validation contains much fewer samples but it is difficult to overfit (despite continuously parameterized architecture). However, as discussed above, below a certain threshold ($\nv\leq 100$), differentiable search indeed overfits to the validation leading to deteriorating test risk.}

\begin{figure}\centering
	\subfigure[{Correlation as a function of the left singular dimension $\h$ (corresponding to \# of hyperparameters)}] { \label{fig:h}
		\begin{tikzpicture}
			\node at (0,0) [scale=0.45]{\includegraphics[width=0.9\textwidth,height=0.65\textwidth]{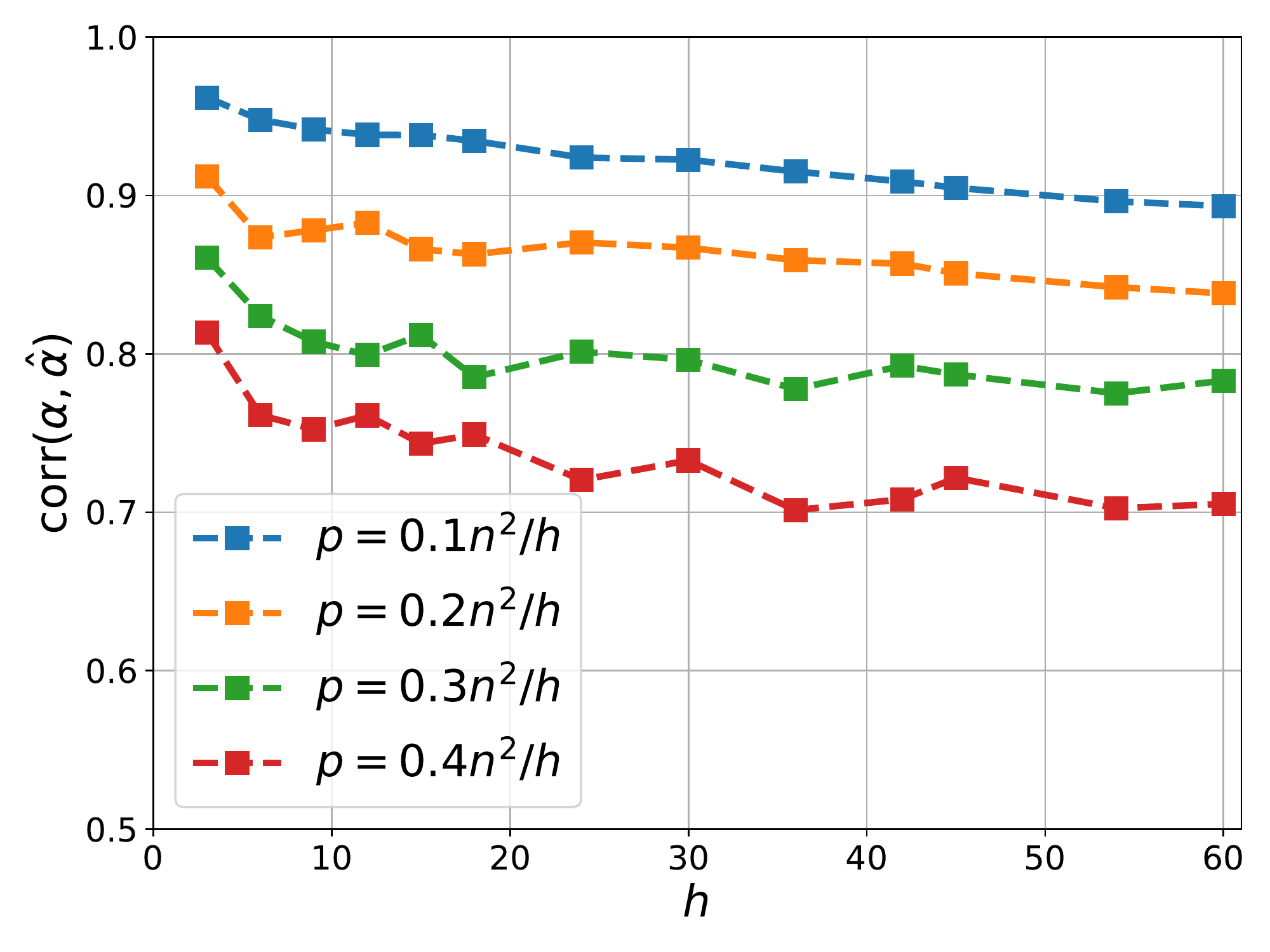}};
		\end{tikzpicture}
	}\hspace{2pt}
	\subfigure[{Correlation as a function of overparameterization level (left-dim+right-dim / sample size)}] { \label{fig:overh}
		\begin{tikzpicture}
			\node at (0,0) [scale=0.45]{\includegraphics[width=0.9\textwidth,height=0.65\textwidth]{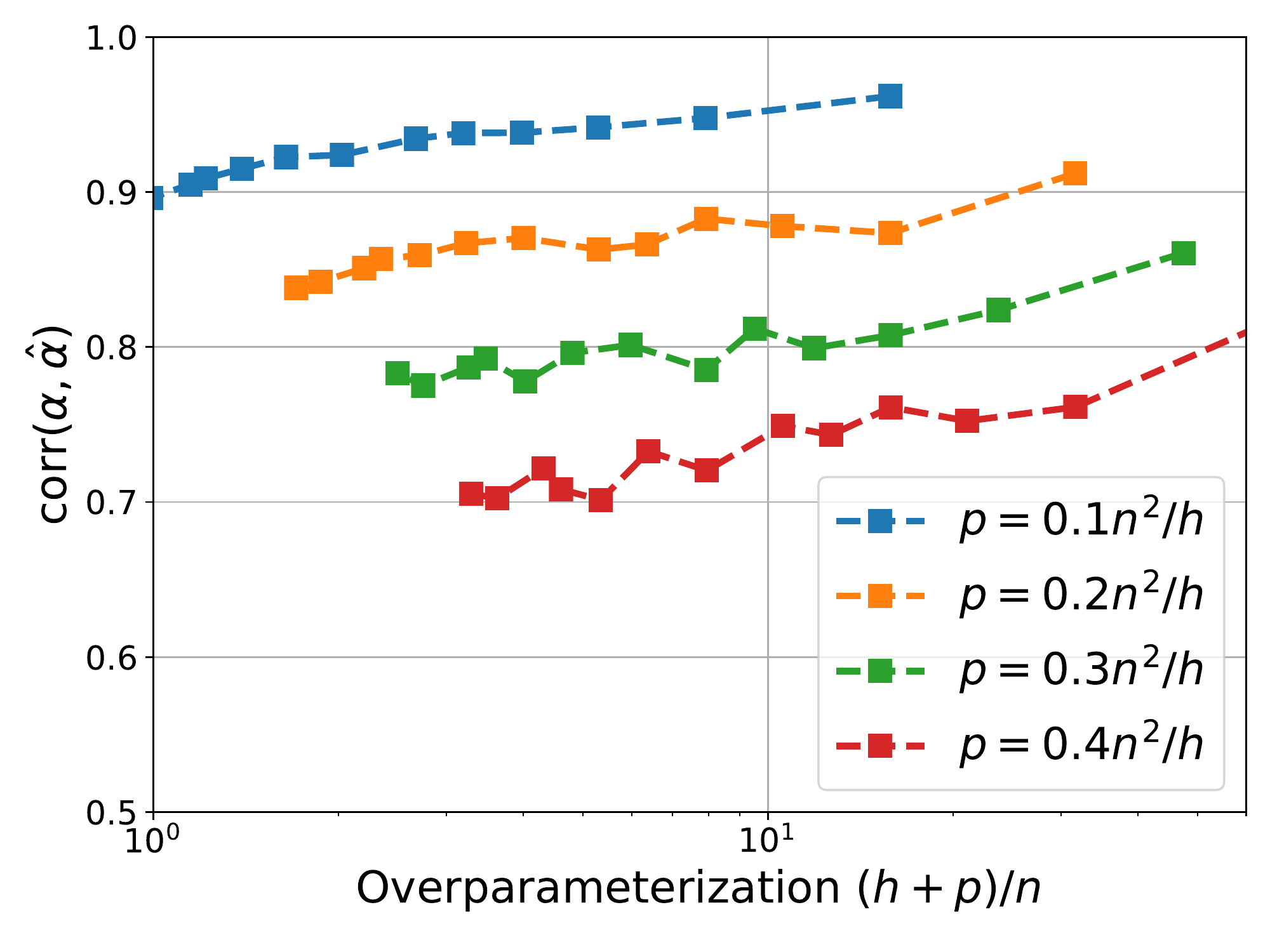}};
		\end{tikzpicture}
	}\vspace{-8pt}\caption{\normalsize{Overparameterized rank-1 learning setup of Sec.~\ref{sec rank 1 factor}. The (absolute value of) the correlation coefficient between $\bas$ and the estimate $\bah$ as a function of $\h$. Each curve (with distinct color) corresponds to a fixed $ph/n^2$ choice. \clr{Here we kept the notation consistent with Sec.~\ref{sec rank 1 factor} and set $n=\nt=\nv$.}}}\label{fig both}\vspace{-8pt}
\end{figure}

\vs\noindent\textbf{c.~~Overparameterized Rank-1 Learning.} We now aim to verify the theoretical claims of Sec.~\ref{sec rank 1 factor} on rank-1 learning. Specifically, we will verify our claim \eqref{high corr} and empirically demonstrate that recovery of the ground-truth hyperparameter $\bas$ requires $hp\lesssim n^2$ where we set $n=\nt=\nv$. This is in contrast to the arguably more intuitive $\h\lesssim n$ requirement. Figure \ref{fig both} summarizes our numerical results. Our experiment is constructed as follows. We generate an $\h \times p$ rank-1 matrix $\M=\bas\bts^T$ with left and right singular vectors $\bas,\bts$ generated as i.i.d.~Gaussians normalized to unit norm. We collect $n$ noiseless labels via $y_i=\bas^T\X_i\bts$ where $\X_i\distas\Nn(0,1)$ as in Theorem \ref{thm low-rank} and apply the spectral estimator \eqref{spect est} to estimate the $\bas$ vector. In our experiments, we vary $\h$ between $0$ to $60$ and we set $p=\gamma n^2/h$. $\gamma$ is also varied from $0.1$ to $0.4$. Figure \ref{fig:h} displays the (absolute value of) the correlation coefficient between $\bas$ and the estimate $\bah$ as a function of $\h$. Each curve (with distinct color) corresponds to a fixed $ph/n^2$ choice. Observe that these curves remain constant even if $\h$ is varying more than a factor of 20. This indicates that correlation indeed depends on $\h p$ rather than solely $\h$. When we increase $\gamma$, $p$ increases and correlation noticeably decreases as we move from a higher curve to a lower curve again indicating the dependence on $p$. Finally, Figure \ref{fig:overh} displays the exact same information but the x-axis is the total number of parameters normalized by the total data (used for spectral initialization). This shows that, just as predicted by Theorem \ref{thm low-rank}, hyperparameter $\bas$ can be learned even when $p$ is much larger than $n$ as long as $hp\lesssim n^2$.

\section{Related Works}\label{sec related}

Our work establishes generalization guarantees for architecture search and is closely connected to the literature on deep learning theory, statistical learning, and hyperparameter optimization / NAS. These connections are discussed below.

\textbf{Statistical learning:} The statistical learning theory provide rich tools for analyzing test performance of algorithms \cite{bartlett2002rademacher,vapnik2006estimation}. Our discussion on learning with bilevel optimization and train-validation split connects to the model selection literature \cite{kearns1996bound,kearns1997experimental,vuong1989likelihood} as well as the more recent works on architecture search \cite{khodaksimple,khodak2019weight}. The model selection literature is mostly concerned with controlling the model complexity (e.g.~via nested hypothesis), which is not directly applicable to high-capacity deep nets. The latter two works are closer to us and also establish connections between feature maps and NAS. However, there are key distinctions. First, we operate on continuous hyperparameter spaces whereas these consider discrete hyperparameters which are easier to analyze. Second, their approaches do not directly apply to neural nets as they have to control the space of all networks with zero training loss which can be large. In contrast, we analyze tractable lower-level algorithms such as gradient-descent and study the properties of a single model returned by the algorithm. \cite{guyon1997scaling} discuss methods for determining train-validation split ratios. Favorable learning theoretic properties of (cross-)validation are studied by \cite{kearns1999algorithmic,xu2020rademacher}. These works either apply to specific scenarios such as tuning lasso penalty or do not consider hyperparameters. We also note that algorithmic stability of \cite{bousquet2001algorithmic} utilizes stability of an algorithm to changes in the training set. In contrast, we consider the stability with respect to hyperparamters. \cite{bai2020important} discusses the importance of train-validation split in meta-learning problems, which also accept a bilevel formulation. Finally, \cite{wang2020guarantees} explores tuning the learning rate for improved generalization. They focus on a simple quadratic objective using hyper-gradient methods and characterize when train-validation split provably helps.

\textbf{Generalization in deep learning:} The statistical study of neural networks can be traced back to 1990's \cite{anthony2009neural, bartlett1998almost,bartlett1998sample}. With the success of deep learning, the generalization properties of deep networks received a renewed interest in recent years \cite{dziugaite2017computing,arora2018stronger,neyshabur2017exploring,golowich2018size}. \cite{bartlett2017spectrally,neyshabur2017pac} establish spectrally normalized risk bounds for deep networks and \cite{nagarajan2018deterministic} provides refined bounds by exploiting inter-layer Jacobian. \cite{arora2018stronger} proposes tighter bounds using compression techniques. These provide a solid foundation on the generalization ability of deep nets. More recently, \cite{jacot2018neural} has introduced the neural tangent kernel which enables the analysis of deep nets trained with gradient-based algorithms: With proper initialization, wide networks behave similarly to kernel regression. NTK has received significant attention for analyzing the optimization and learning dynamics of wide networks \cite{du2018gradient,allen2019convergence,chizat2018note,zhang2019training,nitanda2019refined,zou2018stochastic,wang2020global,oymak2019overparameterized}. Closer to us, \cite{cao2019generalization,arora2019fine,ma2019comparative,oymak2019generalization,allen2018learning,arora2019fine} provide generalization bounds for gradient descent training. A line of research implements neural kernels for convolutional networks and ResNets ~\cite{yang2019scaling,shankar2020neural,li2019enhanced,huang2020deep}. Related to us \cite{arora2019exact} mention the possibility of using NTK for NAS and recent work by \cite{park2020towards} shows that such an approach can indeed produce good results and speed up NAS. In connection to these, \S\ref{sec:FMAP} and \ref{sec deep} establish the first provable guarantees for NAS and also provide a rigorous justification of the NTK-based NAS by establishing data-dependent bounds under verifiable assumptions.

\textbf{Neural Architecture Search and Bilevel Optimization:} HPO and NAS find widespread applications in machine learning. These are often formulated as a bilevel optimization problem, which seeks the optimal hyperparameter at the upper-level optimization minimizing a validation loss. There are variety of NAS approaches employing reinforcement learning, evolutionary search, and Bayesian optimization \cite{zoph2016neural,baker2016designing,zhong2020blockqnn}. Recently differentiable optimization methods have emerged as a powerful tool for NAS (and HPO) problem \cite{liu2018darts,xie2018snas,cai2018proxylessnas,xu2019pc,li2020geometry,pham2018efficient,dong2020autohas} which use continuous relaxations of the architecture. Specifically, DARTS proposed by \cite{liu2018darts} uses a continuous relaxation on the upper-level optimization and weight sharing in the lower level. The algorithm optimizes the lower and upper level simultaneously by gradient descent using train-validation split. \cite{xie2018snas,xu2019pc,pham2018efficient,dong2020autohas} provide further improvements on differentiable NAS. The success of differentiable HPO/NAS methods has further raised interest in large continuous hyperparameter spaces and accelerating bilevel optimization \cite{franceschi2018bilevel,li2020geometry,lorraine2020optimizing}. In this paper we initiate some theoretical understanding of this exciting and expanding empirical literature.

\textbf{High-dimensional learning:} In \S\ref{sec rank 1 factor}, we use ideas from high-dimensional learning to establish algorithmic results. Closest to us are the works on spectral estimators. The recent literature utilizes spectral methods for low-rank learning problems such as phase-retrieval and clustering \cite{von2007tutorial}. Spectral algorithm is used to initialize the algorithm within a basin of attraction for a subsequent method such as convex optimization or gradient descent. This is in similar flavor to our Theorem \ref{thm low-rank} which employs a two-stage algorithm. \cite{lu2020phase,luo2019optimal,mondelli2020optimal} provide asymptotic/sharp analysis for spectral methods for phase retrieval. However, unlike our problem, these works all focus symmetric matrices and operate in the low-dimensional regime where sample size is more than the parameter size. While not directly related, we remark that sparse phase retrieval and sparse PCA problems \cite{jaganathan2013sparse,zou2006sparse} do lead to a high-dimensional regime (sample size less than parameter size) due to the sparsity prior on the parameter.

\vs\vs\section{Conclusion and Future Directions}

In this paper, we explored theoretical aspects of the NAS problem. We first provided statistical guarantees when solving bilevel optimization with train-validation split. We showed that even if the lower-level problem overfits --which is common in deep learning-- the upper-level problem can guarantee generalization with a few validation data. We applied these results to establish guarantees for the optimal activation search problem and extended our theory to generic neural architectures. These formally established the high-level intuition in Figure \ref{overview fig}. We also showed interesting connections between the activation search and a novel low-rank matrix learning problem and provided sharp algorithmic guarantees for the latter.

There are multiple exciting directions for future research. First, one can develop variants of Theorems \ref{one layer nas supp} and \ref{gen thm generic} by studying other lower-level algorithms (e.g.~different loss functions, incorporate regularization) and, most importantly, developing a better understanding of the architecture search spaces and architectural operations. Second, our results are established for the NTK (i.e. lazy training) regime and it would be desirable to obtain similar results for other learning regimes such as the mean-field regime \cite{mei2018mean}. Finally, it would be interesting to study both computational and statistical aspects of the gradient-based solutions to the upper-level problem \eqref{opt alpha}. To this aim, our Theorem \ref{thm val} established that gradient of the validation loss (i.e.~hyper-gradient) uniformly converges under mild assumptions.  However, besides this, we require a deeper understanding of the population landscape (i.e.~as $\nv\rightarrow\infty$) of the the upper-level validation problem (even for the shallow NAS problem of Section \ref{sec:FMAP}) which might necessitate new advances in bilevel optimization theory.

\section*{Acknowledgements}
S.O. and M.L.~are supported by NSF-CNS under award \#1932254 and by an NSF-CAREER under award \#2046816. M.S.~is supported by the Packard Fellowship in Science and Engineering, a Sloan Research Fellowship in Mathematics, an NSF-CAREER under award \#1846369, the Air Force Office of Scientific Research Young Investigator Program (AFOSR-YIP) under award \# FA9550-18-1-0078, DARPA Learning with Less Labels (LwLL) and FastNICS programs, and NSF-CIF awards \#1813877 and \#2008443.

\begin{comment}
\section*{Acknowledments}
The works of S.O. and M.L.~were partially supported in part by the NSF under Award
CNS-1932254 and by the NSF-CAREER Award 2046816. The work of M.S.~was supported in part by the
Packard Fellowship in Science and Engineering, in part by the Sloan Research
Fellowship in Mathematics, in part by NSF-CAREER under Award 1846369,
in part by the Air Force Office of Scientific Research Young Investigator
Program under Award FA9550-18-1-0078, in part by the DARPA Learning
with Less Labels and Fast Network Interface Cards Programs, in part by
NSF-CIF under Award 1813877.
\end{comment}

\let\OLDbib\thebibliography
\renewcommand\thebibliography[1]{
  \OLDbib{#1}
  \setlength{\parskip}{0pt}
  \setlength{\itemsep}{5pt plus 0.3ex}
}
\bibliographystyle{plain}
\bibliography{refs}

\begin{thebibliography}{10}

\bibitem{allen2018learning}
Zeyuan Allen-Zhu, Yuanzhi Li, and Yingyu Liang.
\newblock Learning and generalization in overparameterized neural networks,
  going beyond two layers.
\newblock {\em arXiv preprint arXiv:1811.04918}, 2018.

\bibitem{allen2019convergence}
Zeyuan Allen-Zhu, Yuanzhi Li, and Zhao Song.
\newblock A convergence theory for deep learning via over-parameterization.
\newblock In {\em International Conference on Machine Learning}, pages
  242--252, 2019.

\bibitem{anthony2009neural}
Martin Anthony and Peter~L Bartlett.
\newblock {\em Neural network learning: Theoretical foundations}.
\newblock cambridge university press, 2009.

\bibitem{arora2019fine}
Sanjeev Arora, Simon Du, Wei Hu, Zhiyuan Li, and Ruosong Wang.
\newblock Fine-grained analysis of optimization and generalization for
  overparameterized two-layer neural networks.
\newblock In {\em International Conference on Machine Learning}, pages
  322--332. PMLR, 2019.

\bibitem{arora2019exact}
Sanjeev Arora, Simon~S Du, Wei Hu, Zhiyuan Li, Ruslan Salakhutdinov, and
  Ruosong Wang.
\newblock On exact computation with an infinitely wide neural net.
\newblock {\em arXiv preprint arXiv:1904.11955}, 2019.

\bibitem{arora2018stronger}
Sanjeev Arora, Rong Ge, Behnam Neyshabur, and Yi~Zhang.
\newblock Stronger generalization bounds for deep nets via a compression
  approach.
\newblock In {\em International Conference on Machine Learning}, pages
  254--263. PMLR, 2018.

\bibitem{bai2020important}
Yu~Bai, Minshuo Chen, Pan Zhou, Tuo Zhao, Jason~D Lee, Sham Kakade, Huan Wang,
  and Caiming Xiong.
\newblock How important is the train-validation split in meta-learning?
\newblock {\em arXiv preprint arXiv:2010.05843}, 2020.

\bibitem{baker2016designing}
Bowen Baker, Otkrist Gupta, Nikhil Naik, and Ramesh Raskar.
\newblock Designing neural network architectures using reinforcement learning.
\newblock {\em arXiv preprint arXiv:1611.02167}, 2016.

\bibitem{bartlett1998sample}
Peter~L Bartlett.
\newblock The sample complexity of pattern classification with neural networks:
  the size of the weights is more important than the size of the network.
\newblock {\em IEEE transactions on Information Theory}, 44(2):525--536, 1998.

\bibitem{bartlett2005local}
Peter~L Bartlett, Olivier Bousquet, Shahar Mendelson, et~al.
\newblock Local rademacher complexities.
\newblock {\em The Annals of Statistics}, 33(4):1497--1537, 2005.

\bibitem{bartlett2017spectrally}
Peter~L Bartlett, Dylan~J Foster, and Matus~J Telgarsky.
\newblock Spectrally-normalized margin bounds for neural networks.
\newblock In {\em Advances in Neural Information Processing Systems}, pages
  6241--6250, 2017.

\bibitem{bartlett1998almost}
Peter~L Bartlett, Vitaly Maiorov, and Ron Meir.
\newblock Almost linear vc-dimension bounds for piecewise polynomial networks.
\newblock {\em Neural computation}, 10(8):2159--2173, 1998.

\bibitem{bartlett2002rademacher}
Peter~L Bartlett and Shahar Mendelson.
\newblock Rademacher and gaussian complexities: Risk bounds and structural
  results.
\newblock {\em Journal of Machine Learning Research}, 3(Nov):463--482, 2002.

\bibitem{bousquet2001algorithmic}
Olivier Bousquet and Andr{\'e} Elisseeff.
\newblock Algorithmic stability and generalization performance.
\newblock {\em Advances in Neural Information Processing Systems}, pages
  196--202, 2001.

\bibitem{cai2018proxylessnas}
Han Cai, Ligeng Zhu, and Song Han.
\newblock Proxylessnas: Direct neural architecture search on target task and
  hardware.
\newblock {\em arXiv preprint arXiv:1812.00332}, 2018.

\bibitem{cao2019generalization}
Yuan Cao and Quanquan Gu.
\newblock Generalization bounds of stochastic gradient descent for wide and
  deep neural networks.
\newblock {\em arXiv preprint arXiv:1905.13210}, 2019.

\bibitem{chang2020provable}
Xiangyu Chang, Yingcong Li, Samet Oymak, and Christos Thrampoulidis.
\newblock Provable benefits of overparameterization in model compression: From
  double descent to pruning neural networks.
\newblock {\em AAAI}, 2021.

\bibitem{chizat2018note}
Lenaic Chizat and Francis Bach.
\newblock A note on lazy training in supervised differentiable programming.
\newblock {\em arXiv preprint arXiv:1812.07956}, 2018.

\bibitem{chizat2018lazy}
Lenaic Chizat, Edouard Oyallon, and Francis Bach.
\newblock On lazy training in differentiable programming.
\newblock {\em arXiv preprint arXiv:1812.07956}, 2018.

\bibitem{cubuk2020randaugment}
Ekin~D Cubuk, Barret Zoph, Jonathon Shlens, and Quoc~V Le.
\newblock Randaugment: Practical automated data augmentation with a reduced
  search space.
\newblock In {\em Proceedings of the IEEE/CVF Conference on Computer Vision and
  Pattern Recognition Workshops}, pages 702--703, 2020.

\bibitem{dong2020autohas}
Xuanyi Dong, Mingxing Tan, Adams~Wei Yu, Daiyi Peng, Bogdan Gabrys, and Quoc~V
  Le.
\newblock Autohas: Efficient hyperparameter and architecture search.
\newblock {\em arXiv preprint arXiv:2006.03656}, 2020.

\bibitem{du2019gradient}
Simon Du, Jason Lee, Haochuan Li, Liwei Wang, and Xiyu Zhai.
\newblock Gradient descent finds global minima of deep neural networks.
\newblock In {\em International Conference on Machine Learning}, pages
  1675--1685. PMLR, 2019.

\bibitem{du2018gradient}
Simon~S Du, Xiyu Zhai, Barnabas Poczos, and Aarti Singh.
\newblock Gradient descent provably optimizes over-parameterized neural
  networks.
\newblock {\em arXiv preprint arXiv:1810.02054}, 2018.

\bibitem{dziugaite2017computing}
Gintare~Karolina Dziugaite and Daniel~M Roy.
\newblock Computing nonvacuous generalization bounds for deep (stochastic)
  neural networks with many more parameters than training data.
\newblock {\em arXiv preprint arXiv:1703.11008}, 2017.

\bibitem{finn2017model}
Chelsea Finn, Pieter Abbeel, and Sergey Levine.
\newblock Model-agnostic meta-learning for fast adaptation of deep networks.
\newblock In {\em International Conference on Machine Learning}, pages
  1126--1135. PMLR, 2017.

\bibitem{foster2018uniform}
Dylan~J Foster, Ayush Sekhari, and Karthik Sridharan.
\newblock Uniform convergence of gradients for non-convex learning and
  optimization.
\newblock In {\em Proceedings of the 32nd International Conference on Neural
  Information Processing Systems}, pages 8759--8770, 2018.

\bibitem{franceschi2018bilevel}
Luca Franceschi, Paolo Frasconi, Saverio Salzo, Riccardo Grazzi, and
  Massimiliano Pontil.
\newblock Bilevel programming for hyperparameter optimization and
  meta-learning.
\newblock In {\em International Conference on Machine Learning}, pages
  1568--1577. PMLR, 2018.

\bibitem{golowich2018size}
Noah Golowich, Alexander Rakhlin, and Ohad Shamir.
\newblock Size-independent sample complexity of neural networks.
\newblock In {\em Conference On Learning Theory}, pages 297--299. PMLR, 2018.

\bibitem{gonen2011multiple}
Mehmet Gonen and Ethem Alpaydin.
\newblock Multiple kernel learning algorithms.
\newblock {\em The Journal of Machine Learning Research}, 12:2211--2268, 2011.

\bibitem{guyon1997scaling}
Isabelle Guyon et~al.
\newblock A scaling law for the validation-set training-set size ratio.
\newblock {\em AT\&T Bell Laboratories}, 1(11), 1997.

\bibitem{hastie2019surprises}
Trevor Hastie, Andrea Montanari, Saharon Rosset, and Ryan~J Tibshirani.
\newblock Surprises in high-dimensional ridgeless least squares interpolation.
\newblock {\em arXiv preprint arXiv:1903.08560}, 2019.

\bibitem{heckel2020compressive}
Reinhard Heckel and Mahdi Soltanolkotabi.
\newblock Compressive sensing with un-trained neural networks: Gradient descent
  finds a smooth approximation.
\newblock In {\em International Conference on Machine Learning}, pages
  4149--4158. PMLR, 2020.

\bibitem{huang2020deep}
Kaixuan Huang, Yuqing Wang, Molei Tao, and Tuo Zhao.
\newblock Why do deep residual networks generalize better than deep feedforward
  networks?---a neural tangent kernel perspective.
\newblock {\em Advances in Neural Information Processing Systems}, 33, 2020.

\bibitem{jacot2018neural}
Arthur Jacot, Franck Gabriel, and Cl{\'e}ment Hongler.
\newblock Neural tangent kernel: Convergence and generalization in neural
  networks.
\newblock {\em arXiv preprint arXiv:1806.07572}, 2018.

\bibitem{jaganathan2013sparse}
Kishore Jaganathan, Samet Oymak, and Babak Hassibi.
\newblock Sparse phase retrieval: Convex algorithms and limitations.
\newblock In {\em 2013 IEEE International Symposium on Information Theory},
  pages 1022--1026. IEEE, 2013.

\bibitem{kearns1996bound}
Michael Kearns.
\newblock A bound on the error of cross validation using the approximation and
  estimation rates, with consequences for the training-test split.
\newblock {\em Advances in Neural Information Processing Systems}, pages
  183--189, 1996.

\bibitem{kearns1997experimental}
Michael Kearns, Yishay Mansour, Andrew~Y Ng, and Dana Ron.
\newblock An experimental and theoretical comparison of model selection
  methods.
\newblock {\em Machine Learning}, 27(1):7--50, 1997.

\bibitem{kearns1999algorithmic}
Michael Kearns and Dana Ron.
\newblock Algorithmic stability and sanity-check bounds for leave-one-out
  cross-validation.
\newblock {\em Neural computation}, 11(6):1427--1453, 1999.

\bibitem{khodak2019weight}
Mikhail Khodak, Liam Li, Maria-Florina Balcan, and Ameet Talwalkar.
\newblock On weight-sharing and bilevel optimization in architecture search.
\newblock {\em preprint available at
  https://openreview.net/forum?id=HJgRCyHFDr}, 2019.

\bibitem{khodaksimple}
Mikhail Khodak, Liam Li, Nicholas Roberts, Maria-Florina Balcan, and Ameet
  Talwalkar.
\newblock A simple setting for understanding neural architecture search with
  weight-sharing.
\newblock {\em ICML AutoML Workshop}, 2020.

\bibitem{li2020geometry}
Liam Li, Mikhail Khodak, Maria-Florina Balcan, and Ameet Talwalkar.
\newblock Geometry-aware gradient algorithms for neural architecture search.
\newblock {\em arXiv preprint arXiv:2004.07802}, 2020.

\bibitem{li2020gradient}
Mingchen Li, Mahdi Soltanolkotabi, and Samet Oymak.
\newblock Gradient descent with early stopping is provably robust to label
  noise for overparameterized neural networks.
\newblock In {\em International Conference on Artificial Intelligence and
  Statistics}, pages 4313--4324. PMLR, 2020.

\bibitem{li2019enhanced}
Zhiyuan Li, Ruosong Wang, Dingli Yu, Simon~S Du, Wei Hu, Ruslan Salakhutdinov,
  and Sanjeev Arora.
\newblock Enhanced convolutional neural tangent kernels.
\newblock {\em arXiv preprint arXiv:1911.00809}, 2019.

\bibitem{liu2018darts}
Hanxiao Liu, Karen Simonyan, and Yiming Yang.
\newblock Darts: Differentiable architecture search.
\newblock {\em arXiv preprint arXiv:1806.09055}, 2018.

\bibitem{lorraine2020optimizing}
Jonathan Lorraine, Paul Vicol, and David Duvenaud.
\newblock Optimizing millions of hyperparameters by implicit differentiation.
\newblock In {\em International Conference on Artificial Intelligence and
  Statistics}, pages 1540--1552. PMLR, 2020.

\bibitem{lu2020phase}
Yue~M Lu and Gen Li.
\newblock Phase transitions of spectral initialization for high-dimensional
  non-convex estimation.
\newblock {\em Information and Inference: A Journal of the IMA}, 9(3):507--541,
  2020.

\bibitem{luo2019optimal}
Wangyu Luo, Wael Alghamdi, and Yue~M Lu.
\newblock Optimal spectral initialization for signal recovery with applications
  to phase retrieval.
\newblock {\em IEEE Transactions on Signal Processing}, 67(9):2347--2356, 2019.

\bibitem{ma2019comparative}
Chao Ma, Lei Wu, et~al.
\newblock A comparative analysis of the optimization and generalization
  property of two-layer neural network and random feature models under gradient
  descent dynamics.
\newblock {\em arXiv preprint arXiv:1904.04326}, 2019.

\bibitem{mei2018landscape}
Song Mei, Yu~Bai, Andrea Montanari, et~al.
\newblock The landscape of empirical risk for nonconvex losses.
\newblock {\em Annals of Statistics}, 46(6A):2747--2774, 2018.

\bibitem{mei2019generalization}
Song Mei and Andrea Montanari.
\newblock The generalization error of random features regression: Precise
  asymptotics and double descent curve.
\newblock {\em arXiv preprint arXiv:1908.05355}, 2019.

\bibitem{mei2018mean}
Song Mei, Andrea Montanari, and Phan-Minh Nguyen.
\newblock A mean field view of the landscape of two-layer neural networks.
\newblock {\em Proceedings of the National Academy of Sciences},
  115(33):E7665--E7671, 2018.

\bibitem{mendelson2003few}
Shahar Mendelson.
\newblock A few notes on statistical learning theory.
\newblock In {\em Advanced lectures on machine learning}, pages 1--40.
  Springer, 2003.

\bibitem{mondelli2020optimal}
Marco Mondelli, Christos Thrampoulidis, and Ramji Venkataramanan.
\newblock Optimal combination of linear and spectral estimators for generalized
  linear models.
\newblock {\em arXiv preprint arXiv:2008.03326}, 2020.

\bibitem{nagarajan2018deterministic}
Prabhat Nagarajan, Garrett Warnell, and Peter Stone.
\newblock Deterministic implementations for reproducibility in deep
  reinforcement learning.
\newblock {\em arXiv preprint arXiv:1809.05676}, 2018.

\bibitem{neyshabur2017exploring}
Behnam Neyshabur, Srinadh Bhojanapalli, David McAllester, and Nathan Srebro.
\newblock Exploring generalization in deep learning.
\newblock {\em arXiv preprint arXiv:1706.08947}, 2017.

\bibitem{neyshabur2017pac}
Behnam Neyshabur, Srinadh Bhojanapalli, David McAllester, and Nathan Srebro.
\newblock A pac-bayesian approach to spectrally-normalized margin bounds for
  neural networks.
\newblock {\em arXiv preprint arXiv:1707.09564}, 2017.

\bibitem{nitanda2019refined}
Atsushi Nitanda and Taiji Suzuki.
\newblock Refined generalization analysis of gradient descent for
  over-parameterized two-layer neural networks with smooth activations on
  classification problems.
\newblock {\em arXiv preprint arXiv:1905.09870}, 2019.

\bibitem{oymak2018learning}
Samet Oymak.
\newblock Learning compact neural networks with regularization.
\newblock In {\em International Conference on Machine Learning}, pages
  3966--3975. PMLR, 2018.

\bibitem{oymak2019generalization}
Samet Oymak, Zalan Fabian, Mingchen Li, and Mahdi Soltanolkotabi.
\newblock Generalization guarantees for neural networks via harnessing the
  low-rank structure of the jacobian.
\newblock {\em arXiv preprint arXiv:1906.05392}, 2019.

\bibitem{oymak2019overparameterized}
Samet Oymak and Mahdi Soltanolkotabi.
\newblock Overparameterized nonlinear learning: Gradient descent takes the
  shortest path?
\newblock In {\em International Conference on Machine Learning}, pages
  4951--4960. PMLR, 2019.

\bibitem{oymak2020towards}
Samet Oymak and Mahdi Soltanolkotabi.
\newblock Towards moderate overparameterization: global convergence guarantees
  for training shallow neural networks.
\newblock {\em IEEE Journal on Selected Areas in Information Theory}, 2020.

\bibitem{park2020towards}
Daniel~S Park, Jaehoon Lee, Daiyi Peng, Yuan Cao, and Jascha Sohl-Dickstein.
\newblock Towards nngp-guided neural architecture search.
\newblock {\em arXiv preprint arXiv:2011.06006}, 2020.

\bibitem{pham2018efficient}
Hieu Pham, Melody Guan, Barret Zoph, Quoc Le, and Jeff Dean.
\newblock Efficient neural architecture search via parameters sharing.
\newblock In {\em International Conference on Machine Learning}, pages
  4095--4104. PMLR, 2018.

\bibitem{sattar2020non}
Yahya Sattar and Samet Oymak.
\newblock Non-asymptotic and accurate learning of nonlinear dynamical systems.
\newblock {\em arXiv preprint arXiv:2002.08538}, 2020.

\bibitem{shankar2020neural}
Vaishaal Shankar, Alex Fang, Wenshuo Guo, Sara Fridovich-Keil, Jonathan
  Ragan-Kelley, Ludwig Schmidt, and Benjamin Recht.
\newblock Neural kernels without tangents.
\newblock In {\em International Conference on Machine Learning}, pages
  8614--8623. PMLR, 2020.

\bibitem{vapnik2006estimation}
Vladimir Vapnik.
\newblock {\em Estimation of dependences based on empirical data}.
\newblock Springer Science \& Business Media, 2006.

\bibitem{vershynin2018high}
Roman Vershynin.
\newblock {\em High-dimensional probability: An introduction with applications
  in data science}, volume~47.
\newblock Cambridge university press, 2018.

\bibitem{von2007tutorial}
Ulrike Von~Luxburg.
\newblock A tutorial on spectral clustering.
\newblock {\em Statistics and computing}, 17(4):395--416, 2007.

\bibitem{vuong1989likelihood}
Quang~H Vuong.
\newblock Likelihood ratio tests for model selection and non-nested hypotheses.
\newblock {\em Econometrica: Journal of the Econometric Society}, pages
  307--333, 1989.

\bibitem{wang2020global}
Haoxiang Wang, Ruoyu Sun, and Bo~Li.
\newblock Global convergence and induced kernels of gradient-based
  meta-learning with neural nets.
\newblock {\em arXiv preprint arXiv:2006.14606}, 2020.

\bibitem{wang2020guarantees}
Xiang Wang, Shuai Yuan, Chenwei Wu, and Rong Ge.
\newblock Guarantees for tuning the step size using a learning-to-learn
  approach.
\newblock {\em arXiv preprint arXiv:2006.16495}, 2020.

\bibitem{xie2018snas}
Sirui Xie, Hehui Zheng, Chunxiao Liu, and Liang Lin.
\newblock Snas: stochastic neural architecture search.
\newblock {\em arXiv preprint arXiv:1812.09926}, 2018.

\bibitem{xu2020rademacher}
Ning Xu, Timothy~CG Fisher, and Jian Hong.
\newblock Rademacher upper bounds for cross-validation errors with an
  application to the lasso.
\newblock {\em arXiv preprint arXiv:2007.15598}, 2020.

\bibitem{xu2019pc}
Yuhui Xu, Lingxi Xie, Xiaopeng Zhang, Xin Chen, Guo-Jun Qi, Qi~Tian, and
  Hongkai Xiong.
\newblock Pc-darts: Partial channel connections for memory-efficient
  architecture search.
\newblock {\em arXiv preprint arXiv:1907.05737}, 2019.

\bibitem{yang2019scaling}
Greg Yang.
\newblock Scaling limits of wide neural networks with weight sharing: Gaussian
  process behavior, gradient independence, and neural tangent kernel
  derivation.
\newblock {\em arXiv preprint arXiv:1902.04760}, 2019.

\bibitem{zhang2019training}
Huishuai Zhang, Da~Yu, Wei Chen, and Tie-Yan Liu.
\newblock Training over-parameterized deep resnet is almost as easy as training
  a two-layer network.
\newblock {\em arXiv preprint arXiv:1903.07120}, 2019.

\bibitem{zhong2020blockqnn}
Zhao Zhong, Zichen Yang, Boyang Deng, Junjie Yan, Wei Wu, Jing Shao, and
  Cheng-Lin Liu.
\newblock Blockqnn: Efficient block-wise neural network architecture
  generation.
\newblock {\em IEEE transactions on pattern analysis and machine intelligence},
  2020.

\bibitem{zhou2020theory}
Pan Zhou, Caiming Xiong, Richard Socher, and Steven~CH Hoi.
\newblock Theory-inspired path-regularized differential network architecture
  search.
\newblock {\em arXiv preprint arXiv:2006.16537}, 2020.

\bibitem{zoph2016neural}
Barret Zoph and Quoc~V Le.
\newblock Neural architecture search with reinforcement learning.
\newblock {\em arXiv preprint arXiv:1611.01578}, 2016.

\bibitem{zou2018stochastic}
Difan Zou, Yuan Cao, Dongruo Zhou, and Quanquan Gu.
\newblock Stochastic gradient descent optimizes over-parameterized deep relu
  networks.
\newblock {\em arXiv preprint arXiv:1811.08888}, 2018.

\bibitem{zou2020gradient}
Difan Zou, Yuan Cao, Dongruo Zhou, and Quanquan Gu.
\newblock Gradient descent optimizes over-parameterized deep relu networks.
\newblock {\em Machine Learning}, 109(3):467--492, 2020.

\bibitem{zou2006sparse}
Hui Zou, Trevor Hastie, and Robert Tibshirani.
\newblock Sparse principal component analysis.
\newblock {\em Journal of computational and graphical statistics},
  15(2):265--286, 2006.

\end{thebibliography}

\newpage

\appendix
\section*{Organization of the Appendix}
\begin{enumerate}

\item We gather some useful statistical learning and concentration results in Section \ref{A}.

\item Proofs of our generic train-validation bounds and feature maps are provided in Section \ref{sec generic bound}.

\item Proofs for general architectures appear in Sections \ref{deep fmap} and \ref{deep j sec}.

\item Proofs for algorithmic guarantees (i.e.~overparameterized low-rank learning) are the subject of Section \ref{spectral thm} and \ref{thm low-rank sec}.

\item Finally Theorem \ref{one layer nas supp} (activation search for shallow networks) is proven in Sections \ref{shallow proof} and \ref{sec mingchen}.
\end{enumerate}\vspace{-5pt}

\section{Useful Statistical Learning and Concentration Results}\label{A}
Let $(\eps_i)_{i=1}^{\nt}$ be Rademacher random variables. Define Rademacher complexity of a function class $\Fc$ as
\begin{align}
\Rc_{\nt}(\Fc)&=\frac{1}{\nt}\E_{\Tc,\eps_i}\Bigg[\sup_{f\in\Fc}\sum_{i=1}^{\nt}\eps_if(\x_i)\Bigg].
\end{align}
\begin{lemma} \label{rad comp lem}Suppose the loss function $\ell$ is bounded in $[0,C]$ and $\Gamma$-Lipschitz in the second argument. Also define 
\[
\hat{f}=\arg\min_{f\in\Fc}\Lch_\Tc(f)\quad\text{where}\quad \Lch_\Tc(f)=\frac{1}{\nt}\sum_{i=1}^{\nt}\ell(y_i,f(\x_i)).
\]
Then with probability at least $1-\e^{-t}$, for all $f\in\Fc$ we have
\[
%\sup_{f\in\Fc}
\Lc_{\Dc}(f)\leq \Lch_\Tc(f)+ 2\Gamma \Rc_{\nt}(\Fc)+C\sqrt{\frac{t}{\nt}}.
\]
\end{lemma}

The corollary below is a standard generalization result for linear models. 
\begin{corollary}[Linear models] \label{corr linear}Let $\Fcl_R=\{f\bgl f(\x)=\bt^T\phi(\x),~\tn{\bt}\leq R\}$. Suppose $\tn{\phi(\x)}^2\leq B$ for all $\x\in\Xc$. Then, with probability at least $1-e^{-t}$, for all $f\in\Fcl_R$, we have 
\[
\Lc_{\Dc}(f)\leq \Lch_\Tc(f)+\frac{2\Gamma R\sqrt{B}+C\sqrt{t}}{\sqrt{\nt}}.
\]
Define $\bPhi=[\phi(\x_1)~\dots~\phi(\x_{\nt})]^T$. Set $\Kb=\bPhi\bPhi^T$. Suppose $n>p=\text{dim}(\bt)$ and $\sigma_{\min}(\bPhi)>0$. Consider the min Euclidean norm estimator $\bth=\bPhi^\dagger \y$ and $\hat{f}(\x)=\bth^T\phi(\x)$. Noting that $\tn{\bth}=\tn{\bPhi^\dagger \y}=\sqrt{\y^T\Kb^{-1}\y}$, we arrive at
\begin{align}
\Lc_{\Dc}(\hat{f})\leq \frac{2\Gamma \sqrt{B\y^T\Kb^{-1}\y}+C\sqrt{t}}{\sqrt{\nt}}.\label{corr unif concent}
\end{align}
\end{corollary}
\begin{proof} 
The Rademacher complexity of $\Fcl_R$ is bounded as
\begin{align*}
\Rc_{\nt}(\Fcl_R)&=\frac{1}{\nt}\E_{\Tc,\eps_i}\Bigg[\sup_{\tn{\bt}\leq R}\sum_{i=1}^{\nt}\eps_i\bt^T\phi(\x_i)\Bigg]\\
&=\frac{R}{\nt}\E\Bigg[\twonorm{\sum_{i=1}^{\nt}\eps_i\phi(\x_i)}\Bigg]\leq \frac{R\sqrt{B}}{\sqrt{\nt}}.
\end{align*}
To finish the proof observe that $\hat{f}\subset \Fcl_R$ with $R=\sqrt{\y^T\Kb^{-1}\y}$.
\end{proof}

\begin{lemma}[Moment concentration with Gaussian tail]\label{moment gauss} Let $X$ be nonnegative satisfy the tail bound $\Pro(X\geq E+t)\leq e^{-t^2/2}$ for some $E\geq 0$. Then 
\[
\E[|X|^k]\leq 2^{k+1}\max(E,k+2)^k,
\]
which implies
\[
\E[|X|^k]^{1/k}\leq 2\max(E,k+2).
\]
\end{lemma}
\begin{proof} If $E\leq k+2$, we will simply use the bound $\Pro(X\geq k+2+t)\leq e^{-t^2/2}$. The tail condition implies 
\[
\Pro(X\geq 2Et)\leq \begin{cases}1~if~t\leq 1\\e^{-Et^2/2}~~\text{if}~~t\geq 1\end{cases},
\]
which implies that
\[
\Pro(X^k\geq 2^ktE^k)\leq \begin{cases}1~if~t\leq 1\\e^{-Et^{2/k}/2}~~\text{if}~~t\geq 1\end{cases},
\] 
Let $f,Q$ be the pdf and tail of $(X/2E)^k$. Then
\begin{align}
\frac{\E[X^k]}{2^kE^k}&=\int_0^\infty f(t)tdt=-\int_0^\infty tdQ(t)=-[Q(t)t]_0^\infty +\int_0^\infty Q(t)dt\\
&=\int_0^\infty Q(t)dt\leq 1+\int_{1}^\infty e^{-Et^{2/k}/2}dt= 1+\sum_{i=0}^\infty \int_{e^{ik/2}}^{e^{(i+1)k/2}} e^{-Et^{2/k}/2}dt\\
&\leq 1+\sum_{i=0}^\infty e^{(i+1)k/2} e^{-Ee^i/2}\\
&\leq 1+\sum_{i=0}^\infty e^{-i-1}\leq 2.
\end{align}
For the final line to hold, we simply need $e^{Ee^{i}/2}\geq e^{i+1}e^{(i+1)k/2}$. Taking logs of both sides, this reduces to $Ee^{i}/2\geq (i+1)(k+2)/2$ which is the same as 
\[
Ee^{i}\geq (i+1)(k+2).
\]
Clearly, this holds for all $i$ when $E\geq k+2$.
\end{proof}
\section{Generalizations and Proofs for Learning with Train-Validation Split\\(Theorems \ref{thm val}, \ref{e2e bound} \& Proposition \ref{unif excess lem})}\label{sec generic bound}
\subsection{Uniform Convergence of Loss Functionals}\label{app functional}

We will show uniform concentration of an $m$-dimensional functional $\FB$ of the loss function i.e.~we shall first study $\FB\ell(y,f(\x))\in\R^m$. To obtain the results on loss function or its gradient, we can set $\FB$ to be identity and $\nabla$ operator respectively. More generally, $\FB$ can also represent the Hessian or projection of the gradient to proper subspaces of interest. Associated with the functional $\FB$ we define the $\FB$ distance metric to be
\[
\lif{f_1,f_2}=\sup_{y\in\Yc,\x\in\Xc}\tn{\FB\ell(y,f_1(\x))-\FB\ell(y,f_2(\x))}.
\]
Note that, for simplicity of exposition in our notation we dropped the dependence on $\Xc,\Yc,\ell$. This distance will serve as a proxy for the $\lix{\cdot}$-norm in the generalized analysis. We also loosen Assumption \ref{algo lip} to relax global Lipschitzness condition so that $\Ac$ can instead be \textbf{approximately locally-Lipschitz}. Thus, the following provides a generalization of Assumption \ref{algo lip}. 
\begin{assumption} \label{algo lipFB}Suppose there exists a partitionining $\Pc=(\Bal_i)_{i=1}^P$ of $\Bal$ such that $\log(P)\leq \deff \log(\bar{C})$ and the algorithm $\Ac$ satisfies the following. Over each $\Bal_i\subset\Pc$, $\Ac(\cdot)$ is an $(L,\eps_0)$-Lipschitz approximable function of $\bal$ in $\lif{\cdot,\cdot}$ distance. That is, there exists a function $g_\bal$ such that, $\lif{\ft_\bal,g_\bal}\leq \eps_0$ over $\Bal_i$ and $g_\bal$ is $L$-Lipschitz function of $\bal$ in $\lif{\cdot,\cdot}$ that is $\lif{g_{\bal_1},g_{\bal_2}}\leq L\tn{\bal_1-\bal_2}$.
\end{assumption}

\begin{assumption}\label{loss assump2}$\FB\ell(y,\ft_\bal(\x))-\E[\FB\ell(y,\ft_\bal(\x))]$ has subexponential ($\|\cdot\|_{\psi_1}$) norm bounded by some $S\geq 1$ with respect to the randomness in $(\x,y)\sim\Dc$. 
\end{assumption}
The following lemma is obtained as a corollary of Lemma D.7 of \cite{oymak2018learning} by specializing it to the unit Euclidean ball.
\begin{lemma} \label{subexplem}Consider the empirical and population functionals $\FB\Lch_\Vc(\ft_\bal)=\frac{1}{\nt}\sum_{i=1}^{\nt} \FB\ell(y_i,\ft_\bal(\x_i))$ and $\FB\Lc(\ft_\bal)=\E[\FB\ell(y,\ft_\bal(\x))]$ respectively. Suppose $\nv\geq m$. Then
\[
\Pro\Bigg\{\twonorm{\FB\Lch_\Vc(\ft_\bal)-\FB\Lc(\ft_\bal)}\gtrsim S\frac{\sqrt{m+t^2}}{\sqrt{\nv}}\Bigg\}\leq 2e^{-\min(t^2,t\sqrt{\nv})}.
\]
\end{lemma}
With this lemma in place we are now ready to state our uniform concentration result.
\begin{theorem}[Uniform Concentration over Validation] \label{thm valFB} Suppose Assumptions \ref{algo lipFB} and \ref{loss assump2} hold. Fix $\tau>0$ and set 
\[
\defz:=\deff\log\left(\bC L \sqrt{\frac{\nv}{\deff}}\right).
\]
Then, as long as $\nv\geq \defz+m+\tau$, with probability at least $1-2\e^{-\tau}$, $\FB\Lch_\Vc$ uniformly converges as follows
\begin{align}
\sup_{\bal\in\Bal}\twonorm{\FB\Lch_\Vc(\ft_\bal)-\FB\Lc(\ft_\bal)}\leq S\sqrt{\frac{C(\defz+m+\tau)}{\nv}}+4\eps_0,\label{functional eq}
\end{align}\vs
with $C>0$ a fixed numerical constant.
\end{theorem}
\begin{proof} Let $\Cc^i_\eps$ be an $\eps/L$-cover of $\Bal_i$ in $\ell_2$ norm and $\Cc_\eps=\bigcup_{i=1}^P\Cc^i_\eps$. Recall from Definition \ref{cover assump} that the covering bound for $\Bal_i\subset\Bal$ obeys $ |\Cc^i_\eps|\leq \deff\log (\bC L/\eps)$. Using the bound on $P$ (via Assumption \ref{algo lipFB}), we arrive at
\[
\log |\Cc_\eps|\leq \log(P)+\deff\log (\bC L/\eps)\leq 2\deff\log (\bC L/\eps).
\]
Let $F=\{f_\bal \bgl \bal\in\Bal\}$. Let $G=\{g_\bal\bgl \bal\in\Bal\}$ be the set of (locally) Lipschitz functions $g_\bal$ within $\eps_0$ neighborhood of $F$. Following Assumption \ref{algo lipFB}, let $G_\eps$ be the $\lif{\cdot,\cdot}$ $\eps$-cover of $G$ induced by $\Cc_\eps$. Additionally, let $F_\eps\subset F$ be a set of hypothesis with same cardinality as $G_\eps$ that are within $\eps_0$ distance to their counterparts in $G_\eps$. For a fixed $f\in F_\eps$, using sub-exponentiality of the loss functional (Assumption \ref{loss assump2}) and subexponential concentration provided in Lemma \ref{subexplem}, for a proper choice of constant $C>0$, with probability $1-2\e^{-\min(t\sqrt{n},t^2)}$, we have that
\[
\twonorm{\FB\Lch_\Vc(f)-\FB\Lc(f)}\leq \frac{S\sqrt{C}\sqrt{m+t^2}}{2\sqrt{\nv}}.
\]
Recall that we choose $\nv\geq \deff\log (\bC L/\eps)+m+\tau$. Also set the short-hand notation $\deft=\deff\log (\bC L/\eps)$. Thus setting $t=\sqrt{2\deft+\tau}\leq \sqrt{\nv}$ ensures
\[
\min(t\sqrt{n},t^2)\geq t^2\geq \log |\Cc_\eps|+\tau.
\]
Thus, union bounding over $F_\eps$, we find that with probability $1-2\e^{-\tau}$, for all $f\in F_\eps$
\begin{align}
\tn{\FB\Lch_\Vc(f)-\FB\Lc(f)}\leq S\sqrt{C}\frac{\sqrt{m+\deft+\tau}}{\sqrt{\nv}}.\label{unifFB}
\end{align}
Now, fix any $f\in F$. Pick $g\in G$ and $g'\in G_\eps$ such that $\lif{g,g'}\leq \eps$ and $\lif{g,f}\leq \eps_0$. Additionally pick $f'\in F_\eps$ which is $\eps_0$-neighbor of $g'\in G_\eps$. This implies that, for all feasible $(\x,y)$ 
\begin{align}
&\tn{\FB\ell(y,f(\x))-\FB\ell(y,f'(\x))}\nn\\
&~~~\leq\tn{\FB\ell(y,f(\x))-\FB\ell(y,g(\x))}+\tn{\FB\ell(y,g(\x))-\FB\ell(y,f'(\x))}+\tn{\FB\ell(y,g'(\x))-\FB\ell(y,f'(\x))}\nn\\
&~~~\leq \eps+2\eps_0,\nn
\end{align}
via Assumption \ref{algo lipFB}. This further implies the same bound for population and empirical functionals
\[
\tn{\FB\Lch_\Vc(f)-\FB\Lch_\Vc(f')},\tn{\FB\Lc(f)-\FB\Lc(f')}\leq \eps+2\eps_0.
\]
Combining this with \eqref{unifFB} leads to the following uniform convergence bound for all $f\in F$
\[
\sup_{f\in F}\tn{\FB\Lc(f)-\FB\Lch_\Vc(f')}\leq S\frac{\sqrt{C(\deft+m+\tau)}}{\sqrt{\nv}}+2(\eps+2\eps_0).
\]
To proceed, select $\eps=\sqrt{\frac{\deff}{\nv}}$ and set $\defz=\deff\log(\bC L \sqrt{\nv/\deff})$. Thus,
\begin{align}
\sup_{f\in F}|\Lc(f_\bal)-\Lch_\Vc(f_\bal)|\leq S\sqrt{\frac{C(\defz+m+\tau)}{\nv}}+4\eps_0,
\end{align}
concluding the proof of the bound.
\end{proof}

\subsection{Proof of Theorem \ref{thm val}}
Let us state slight generalization of Assumptions \ref{algo lip} and \ref{algo lipp} which will be utilized in our proof. 
\begin{assumption} \label{algo lip2} There exists a function $g_\bal$ such that, for all pairs $\bal_1,\bal_2\in \Bal$, $\lix{g_{\bal_1}-\ft_{\bal_1}}\leq \eps_0$ and $\lix{g_{\bal_1}-g_{\bal_2}}\leq L\tn{\bal_1-\bal_2}$.
\end{assumption}

\begin{assbis}{algo lip2}\label{algo lip2p}
For some $R\geq 1$ and all $\bal_1,\bal_2\in\Bal$ and $\x\in\Xc$, hyper-gradient obeys $\tn{\nabla_\bal\ft_{\bal_1}(\x)}\leq R$, $\tn{\nabla_\bal\ft_{\bal_1}(\x)-\nabla_\bal\ft_{\bal_2}(\x)}\leq RL\tn{\bal_1-\bal_2}$ and $\tn{\nabla_\bal g_{\bal_1}(\x)-\nabla_\bal\ft_{\bal_1}(\x)}\leq R\eps_0$.
\end{assbis}

The proof will be a corollary of the generalized result Theorem \ref{thm valFB}. Let us first state a lemma to show that Assumptions \ref{loss assump} and \ref{algo lip2} imply Assumptions \ref{algo lipFB} and \ref{loss assump2}.
\begin{lemma} \label{lem corollary}Let $C>0$ be a proper choice of constant. 
\begin{itemize}
\item Assumptions \ref{loss assump} and \ref{algo lip2} imply Assumptions \ref{algo lipFB} and \ref{loss assump2} hold for the loss function (setting $\FB=\text{Identity}$), with $m=1$, $L\rightarrow L\Gamma$, $\eps_0\rightarrow \Gamma\eps_0$, and $S=C$.
\item Assumptions \ref{loss assumpp} and \ref{algo lip2p} imply Assumptions \ref{algo lipFB} and \ref{loss assump2} hold for the gradient (setting $\FB=\nabla$), with $m=h$, $L\rightarrow 2RL\Gamma$, $\eps_0\rightarrow 2R\Gamma\eps_0$, and $S=C$.
\end{itemize}
\end{lemma}
\begin{proof} For the first statement, we conclude that Assumption \ref{algo lipFB} holds with $L\rightarrow L\Gamma$ via %note that %First let us show Lipschitzness. 
\begin{align}
\sup_{y\in\Yc,\x\in\Xc}|\ell(y,f_{\bal_1}(\x))-\ell(y,f_{\bal_2}(\x))|\leq \Gamma \sup_{\x\in\Xc}|f_{\bal_1}(\x)-f_{\bal_2}(\x)|\leq L\Gamma\tn{\bal_1-\bal_2}.\nn
\end{align}
Following the same argument, we plug in $\eps_0\rightarrow \Gamma\eps_0$. To prove the result for the gradient mapping note that if Assumptions \ref{loss assumpp} \& \ref{algo lip2p} hold then, using the fact that $\nabla_\bal\ell(y,f_{\bal}(\x))=\ell'(y,f_{\bal}(\x))\nabla f_{\bal}(\x)$ we have
\begin{align}
\tn{\nabla\ell(y,f_{\bal_1}(\x))-\nabla\ell(y,f_{\bal_2}(\x))}&\leq |\ell'(y,f_{\bal_1}(\x))-\ell'(y,f_{\bal_2}(\x))|\tn{\nabla f_{\bal_1}(\x)}+\Gamma \tn{\nabla f_{\bal_1}(\x)-\nabla f_{\bal_2}(\x)}\nn\\
&\leq \Gamma |f_{\bal_1}(\x)-f_{\bal_2}(\x)|R+\Gamma RL\tn{\bal_1-\bal_2}\nn\\
&\leq 2\Gamma LR\tn{\bal_1-\bal_2}.\nn
\end{align}
Following the same argument, we plug in $\eps_0\rightarrow 2R\Gamma\eps_0$. Subexponentiality with $S=C$ follows from the boundedness condition in Assumption \ref{loss assump}.
\end{proof}

Theorem \ref{thm val} directly follows by setting $\eps_0=0$ and plugging in the looser bound $\nv/\deff\geq \sqrt{\nv/\deff}$ in the definition of $\defz$ term. Note that, the theorem below is a corollary of the more general result in Theorem \ref{thm valFB}.
\begin{theorem}[Learning with Validation -- Lipschitz Approximation] \label{thm val2} Suppose Assumptions \ref{loss assump} \& \ref{algo lip2} hold. Let $\bah$ be the minimizer of the empirical risk \eqref{opt alpha} over validation. Fix $\tau>0$ and set $\defz:=\deff\log(\bC L\Gamma \sqrt{\nv/\deff})$. There exists a constant $C>0$ such that, whenever $\nv\geq \defz+\tau$, with probability at least $1-2\e^{-\tau}$, $f_{\bah}$ achieves the risk bound
\begin{align}
\sup_{\bal\in\Bal}|\Lc(\ft_{\bal})- \Lch_{\Vc}(\ft_\bal)|\leq&\sqrt{\frac{C(\defz+\tau)}{\nv}}+4\Gamma\eps_0\label{eval12},\\
\Lc(\ft_\bah)\leq \min_{\bal\in \Bal}\Lc(\ft_\bal)+2&\sqrt{\frac{C(\defz+\tau)}{\nv}}+8\Gamma\eps_0+\delta.\label{excess risk val2}
\end{align}\vs
Furthermore, suppose Assumptions \ref{loss assumpp} \& \ref{algo lip2p} hold as well. Set $\defg:=h+\deff\log(2\bC RL\Gamma \sqrt{\nv/\deff})$. Whenever $\nv\geq \defg+\tau$, with probability at least $1-2\e^{-\tau}$, 
\begin{align}
\sup_{\bal\in\Bal}\tn{\nabla\Lch_\Vc(\ft_\bal)-\nabla\Lc(\ft_\bal)}\leq \sqrt{\frac{C(\defg+\tau)}{\nv}}+8R\Gamma\eps_0.\label{hyper eq3}
\end{align}\vs
\end{theorem}
\begin{proof} Applying Theorem \ref{thm valFB} and plugging in the first statement of Lemma \ref{lem corollary}, with probability $1-2e^{-\tau}$, we obtain the statement \eqref{eval12}. Here we used the fact that $\defz+\tau+m\leq 2(\defz+\tau)$ and factor $2$ can be subsumed in the constant $C$. 

We obtain the advertised bound \eqref{excess risk val2} via (a) observing that the bound is valid for $\bah$ and the optimal population hypothesis $\bal^\st=\arg\min_{\bal\in\Bal}\Lc(f_\bal)$ and (b) using the fact that
\[
\Lch_\Vc(f_\bah)\leq \inf_{\bal\in\Bal}\Lch_\Vc(f_\bal)+\del\leq \Lch_\Vc(f_{\bal^\st})+\del.
\] 
Following \eqref{eval12}, we first have $\Lc(f_{\bah})\leq \min_{\bal}\Lch_{\Vc}(f_\bal)+\sqrt{\frac{C(\defz+\tau)}{\nv}}+4\eps_0\Gamma+\del$. Using a triangle inequality with $\bal^\st$, we find 
\begin{align}
&\Lc(f_{\bah})\leq \inf_{\bal\in \Bal}\Lc(f_\bal)+ 2\sqrt{\frac{C(\defz+\tau)}{\nv}}+8\eps_0\Gamma+\del\label{stronger eq}
\end{align}

Applying Theorem \ref{thm valFB} and plugging in the second statement of Lemma \ref{lem corollary}, with probability $1-2e^{-\tau}$, we obtain the statement \eqref{hyper eq3}. Here, we used the fact that $m=h$ and $\defz=\deff\log(2\bC RL\Gamma \sqrt{\nv/\deff})$. We then set $\defg=\defz+h$ to conclude.
\end{proof}

\subsection{Proof of Proposition \ref{unif excess lem}}\label{secB3}

For the purpose of our neural network analysis assuming $\Cc_\bal$ to be Lipschitz might be too restrictive. As a result, we will state a slight generalization which allows for Lipschitzness of $\Cc_\bal$ and $\fs_\bal$ over a smaller domain $\Bal_\st$ which provides more flexibility.
\begin{assumption} \label{assump TV}Over a domain $\Bal_\st\subset \Bal$: (1) $\fs_\bal$ is $L$-Lipschitz function of $\bal$ in $\lix{\cdot}$ norm and (2) the excess risk term $\Cc_\bal$ is $\kappa L\sqrt{\nt}$-Lipschitz for some $\kappa>0$. 
\end{assumption}
Note that, the first condition is equivalent to Assumption \ref{algo lip} holding over $\Bal_\st$ in population (i.e.~$\nt\rightarrow\infty$). We intentionally parameterized the Lipschitz constant by the same notation to simplify exposition. The following result is a slight generalization of Proposition \ref{unif excess lem} where we allow for non-Lipschitzness of $\Bal$ by instead assuming it holds over a smaller subset $\Bal_\st$.

\begin{proposition} [Train-Validation Bounds] \label{unif excess lem22} Consider the setting of Theorem \ref{thm val} and for any fixed $\bal\in\Bal$ assume \eqref{unif alll} holds. Additionally suppose Assumption \ref{assump TV} holds. Then with probability at least $1-e^{-\tau}$, 
\begin{align}
\min_{\bal\in\Bal}\Lc(\ft_\bal)\leq \min_{\bal\in\Bal_\st}\left(\Lc(\fs_\bal)+\frac{\Cc_\bal^\Tc}{\sqrt{\nt}}\right)+2C_0\sqrt{\frac{\deff\log(2(\Gamma+\kappa)\nt\bC L/\deff)+\tau}{\nt}}.\label{abcdef}
\end{align}
Using Theorem \ref{thm val}, this in turn implies that with probability at least $1-3e^{-\tau}$
\begin{align}
\Lc(\ft_\bah)\leq \min_{\bal\in\Bal_\st}\left(\Lc(\fs_\bal)+\frac{\Cc_\bal^\Tc}{\sqrt{\nt}}\right)+2C_0\sqrt{\frac{\deff\log(2(\Gamma+\kappa)\nt\bC L/\deff)+\tau}{\nt}}+\sqrt{\frac{C\deff\log(2\bC L\Gamma \nv/\deff)}{\nv}}+\delta.\label{final final}
\end{align}
Observe that \eqref{final final} implies Proposition \ref{unif excess lem} as the sample size setting of the paper is $\nt\geq\nv$ via \eqref{roi eq}.
\end{proposition}
\begin{proof} The proof of \eqref{abcdef} uses a covering argument. Create an $\eps/L$ covering $\Bal_\eps$ of the set $\Bal_\st$ of size $\log|\Bal_\eps|\leq \deff \log(\bC L/\eps)$. For each $\bal\in\Bal_\st$, setting $t={\deff\log(\bC L/\eps)+\tau}$ we have that for all $\bal_\eps\in\Bal_\eps$, the bound \eqref{unif alll} holds with probability $1-e^{-\tau}$. To conclude set $\eps=\frac{C_0\sqrt{\deff/\nt}}{2\Gamma+2\kappa}\geq \frac{C_0\deff/\nt}{2\Gamma+2\kappa}$ and $\defz=\deff\log(2(\Gamma+\kappa)\nt\bC L/\deff)+\tau$. Also observe, via Assumption \ref{assump TV}, that
\[
|\Lc(\fs_{\bal_\eps})-\Lc(\fs_{\bal})|+\frac{|\Cc_{\bal_\eps}^\Tc-{\Cc_{\bal}^\Tc}|}{\sqrt{\nt}}\leq \Gamma \eps+ \kappa\eps\leq 0.5C_0\sqrt{\deff/\nt}.
\]
To proceed, via Assumption \ref{algo lip} we also have $|\Lc(\fs_{\bal_\eps})-\Lc(\fs_{\bal})|\leq \Gamma\eps\leq 0.5C_0\sqrt{\deff/\nt}$. Together, using triangle inequality, these imply for all $\bal\in\Bal_\st$
\begin{align}
&\Lc(\ft_{\bal})\leq \Lc(\ft_{\bal_\eps})+0.5C_0\sqrt{\deff/\nt}+C_0\sqrt{\frac{\defz}{\nt}}\leq \Lc(\fs_{\bal})+\frac{\Cc_{\bal}^\Tc}{\sqrt{\nt}}+2C_0\sqrt{\defz/\nt},\nn\\
\implies&\Lc(\ft_{\bal})\leq  \Lc(\fs_{\bal})+\frac{\Cc_{\bal}^\Tc}{\sqrt{\nt}}+2C_0\sqrt{\frac{\defz}{\nt}}.\label{indiv T bound}
\end{align}
Finally to conclude with \eqref{abcdef} notice the fact that $\min_{\bal\in\Bal}\Lc(\ft_\bal)\leq \min_{\bal\in\Bal_\st}\Lc(\ft_\bal)$.

To conclude with the final statement \eqref{final final}, we apply Theorem \ref{thm val} which bounds $\Lch_\Vc(\ft_\bah)$ in terms of $\min_{\bal\in\Bal}\Lc(\ft_{\bal})$. This results in an overall success probability of $1-3e^{-\tau}$.
\begin{comment}
To proceed with the final result, applying \eqref{indiv T bound} with the union of train and validation data (i.e.~set $\Tc\gets \TVc$), with probability $1-e^{-\tau}$, we find that for all $\bal\in\Bal_\st$
\begin{align*}
\Lc(\ftv_\bal)\leq \left(\Lc(\fs_\bal)+\frac{\Cc_\bal^{\TVc}}{\sqrt{\nt+\nv}}\right)+2C_0\sqrt{\frac{\defz+\tau}{\nt+\nv}},
\end{align*}
with $\defz=\deff\log(2(\Gamma+\kappa)(\nt+\nv)\bC L/\deff)$. To proceed, for $\bah$, note that
\begin{align}
\Lc(\ftv_\bah)&\leq \left(\Lc(\fs_\bah)+\frac{\Cc_\bah^{\TVc}}{\sqrt{\nt+\nv}}\right)+2C_0\sqrt{\frac{\defz+\tau}{\nt+\nv}},\\
&\leq \left(\Lc(\fs_\bah)+\frac{\Cc_\bah^{\Tc}}{\sqrt{\nt}}\right)+\left(\frac{\Cc_\bah^{\TVc}}{\sqrt{\nt+\nv}}-\frac{\Cc_\bah^{\Tc}}{\sqrt{\nt}}\right)+2C_0\sqrt{\frac{\defz+\tau}{\nt+\nv}},
\end{align}
\end{comment}
\end{proof}

\subsection{Proof of Theorem \ref{e2e bound}}\label{proof e2e bound}

\begin{proof} Below $C>0$ is an absolute constant. First under the provided conditions (which include \eqref{low bound} with probability $1-p_0$), applying Lemma \ref{lip fmap lemma}, we find that Assumption \ref{algo lip} ($\Ac$ is Lipschitz) holds with Lipschitz constant $L=\lips$. Assumption \ref{loss assump} holds automatically and Assumption \ref{cover assump} holds with $\deff=\h$ and $\bC=3$. Thus, applying Theorem \ref{thm val2} (with $\eps_0=0$), with probability at least $1-2e^{-\tau}$
\[
\Lc(f_{\bah})\leq \inf_{\bal\in \Bal}\Lc(f_\bal)+ C\frac{\sqrt{\h\log(6RL\Gamma \sqrt{\nv/\h})+\tau}}{\sqrt{\nv}}+\del.
\]
The remaining task is bounding the $\inf_{\bal\in \Bal}\Lc(f_\bal)-\inf_{\bal\in \Bal}\Lc(\fs_\bal)$ term. We do this via Lemma \ref{lemma excess} which yields that with probability $1-2e^{-t}$, for all $\bal\in\Bal$ \eqref{excess bound} holds. Together, these imply that with probability $1-4e^{-\tau}-p_0$
\begin{align*}
\Lc(f_{\bah})-\del&\leq \min_{\bal\in\Bal}C\frac{\sqrt{\h\log(6R\times\lips\Gamma \sqrt{\nv/h})+\tau}}{\sqrt{\nv}}\\
&\quad\quad\quad+\frac{2\Gamma \sqrt{B\y^T\Kb_\bal^{-1}\y}+C\sqrt{\h\log(\lipf)+\tau}}{\sqrt{\nt}}\\
&\leq \min_{\bal\in\Bal}\frac{C\sqrt{\h\log(M)+\tau}}{\sqrt{\nv}}+\frac{2\Gamma \sqrt{B\y^T\Kb_\bal^{-1}\y}+C\sqrt{\h\log(M)+\tau}}{\sqrt{\nt}}\\
&\leq \min_{\bal\in\Bal}\frac{2\Gamma \sqrt{B\y^T\Kb_\bal^{-1}\y}}{\sqrt{\nt}}+\frac{C\sqrt{\h\log(M)+\tau}}{\sqrt{\min(\nt,\nv)}}
\end{align*}
where $M=\lipsum$.
\end{proof}

\subsection{Uniform Concentration of Excess Risk for Feature Maps}
\begin{lemma} \label{lemma excess} Consider the setup of Definition \ref{fmap def}. Let $\sup_{\x\in\Xc,1\leq i\leq \h}\tn{\phi_i(\x)}^2\leq B$. 
Declare $\Kb_\bal=\bPhi_\bal\bPhi_\bal^T$. Suppose \eqref{low bound} holds with probability $1-p_0$ over the training data. Assume $\ell$ is a $\Gamma$ Lipschitz loss bounded by $C\geq 1$. With probability at least $1-p_0-\e^{-t}$, we have that for all $\bal\in\Bal$
\begin{align}
\Lc(\bt_\bal)\leq \frac{2\Gamma \sqrt{B\y^T\Kb_\bal^{-1}\y}+2C\sqrt{\h\log(\lipf)+\tau}}{\sqrt{\nt}}\label{excess bound}
\end{align}
\end{lemma}
\begin{proof} The proof strategy follows that of Proposition \ref{unif excess lem}. Using \eqref{corr unif concent} of Corollary \ref{corr linear}, we know that, for any choice of $\bal\in\Bal$ with probability $1-\e^{-t}$ we have that
\begin{align}
\Lc_{\Dc}(\hat{f})\leq \frac{2\Gamma \sqrt{B\y^T\Kb_\bal^{-1}\y}+C\sqrt{t}}{\sqrt{\nt}}.\label{interpolate bound}
\end{align}
where $\Kb_\bal=\bPhi_\bal\bPhi_\bal^T$. To proceed, we will apply a covering argument to the population loss of the empirical solutions. This will require Lipschitzness of the population loss. Observe that 
\[
\sqrt{\y^T\Kb_\bal^{-1}\y}=\tn{\bt_\bal}=\tn{\bPhi_\bal^\dagger\y}.
\]
Lemma \ref{lip fmap lemma} shows the Lipschitzness of $\bt_\bal$ with $L=\lipp$ which implies that
\[
\tn{\bt_\bal-\bt_{\bal'}}\leq L\tn{\bal-\bal'}.
\]
To proceed, let $\Cc_\eps$ be an $\eps$ cover of $\Bal$ of size $\log|\Cc_\eps|\leq \h\log(3R/\eps)$. Setting $t=\h\log(3R/\eps)+\tau$, with probability $1-\e^{-\tau}$, we find that all $\bal\in \Cc_\eps$ obeys \eqref{interpolate bound} with this choice of $t$. Observe that, with probability at least $1-p_0$, via the lower bound \eqref{low bound}, we have $\tn{\bt_{\bal}},\tn{\bt_{\bal'}}\leq \tn{\y}/\sqrt{\laz}$. To proceed, for any $\bal$ pick $\bal'\in\Cc_\eps$ with $\tn{\bal-\bal'}\leq \eps$ and observe that
\begin{align*}
\Lc(f_{\bal})-\Lc(f_{\bal'})&=\E[\ell(y,\bt_{\bal}^T\phi_\bal(\x)] -\E[\ell(y,\bt_{\bal'}^T\phi_{\bal'}(\x)]\\
&\leq \Gamma\E[|\bt_{\bal}^T\phi_\bal(\x)-\bt_{\bal'}^T\phi_{\bal'}(\x)|]\\
&\leq \Gamma[\E[|(\bt_{\bal}-\bt_{\bal'})^T\phi_\bal(\x)|]+\E[|\bt_{\bal'}^T(\phi_{\bal}(\x)-\phi_{\bal'}(\x))|]]\\
&\leq \Gamma[\E[\tn{\bt_{\bal}-\bt_{\bal'}}\tn{\phi_\bal(\x)}]+\E[\tn{\bt_{\bal'}}\tn{\phi_{\bal}(\x)-\phi_{\bal'}(\x)}]]\\
&\leq \Gamma(RL\sqrt{B}\tn{\bal-\bal'}+\sqrt{\h B}\tn{\bal-\bal'} \tn{\y}/\sqrt{\laz})\\
&\leq \Gamma \sqrt{B}\tn{\y}(5R^3B^{3/2}\sqrt{\nt^3\h}\laz^{-2}+\sqrt{\frac{\h}{\laz}})\eps\\
&\leq \lipt\eps,
\end{align*}
where we used the fact that $\la_0\leq \sigma_{\min}^2(\bPhi_\bal)\leq \nt R^2B$. Thus we obtain that for all $\bal\in\Bal$%\nt R^2B^2
\[
\Lc_{\Dc}(\hat{f})\leq \frac{2\Gamma \sqrt{B\y^T\Kb_\bal^{-1}\y}+C\sqrt{\h\log(3R/\eps)+\tau}}{\sqrt{\nt}}+\lipt\eps.
\]
Here setting $\eps^{-1}=\lipf/3R$, we obtain the desired bound \eqref{excess bound}.
\end{proof}

\section{Results on Algorithmic Lipschitzness}\label{sec algo lip}

\subsection{Proof of Lemma \ref{ridge thm}}
\begin{proof} First applying Lemma \ref{lip fmap lemma}, we find that Assumption \ref{algo lip} holds with \eqref{ridge lip}. Plugging in the boundedness of labels (i.e.~$\tn{\yT}\leq \sqrt{\nt}$), $\Gamma=1$, $\nt\geq h$, $R\geq 1$, we end up with the refined bound
\[
L\leq 6R^3B\nt\laz^{-2}(B\nt\sqrt{\h}+1).
\]
Finally, using the fact that $\bal$ is $h+1$ dimensional, we obtain $\defz=(h+1)\log(\liptwo)$.
\end{proof}

\subsection{Proof of Lemma \ref{convex smooth}}
The following lemma is a rephrasing of Lemma \ref{convex smooth}.
\begin{lemma} [Lipschitz Solutions under Strong-Convexity / Smoothness] Let $\Bal$ be a convex set. Let $\Fc(\bal,\bt):\Bal\times \R^p \rightarrow \R$ be a $\mu$ strongly-convex function of $\bt$ and $L$-smooth function of $\bal$ over the feasible domain $\R^p\times \Bal$. Let
\[
\bt_\bal=\arg\min_{\bt}\Fc(\bal,\bt).
\]
Then, $\bt_\bal$ is $\sqrt{L/\mu}$-Lipschitz function of $\bal$.
\end{lemma}
\begin{proof} Pick a pair $\bal,\bal'\in\Bal$. From strong-convexity of $\bt$ and optimality of $\bt_{\bal'}$, we have that
\[
\Fc(\bal',\bt_\bal)-\Fc(\bal',\bt_{\bal'})\geq \frac{\mu}{2}\tn{\bt_\bal-\bt_{\bal'}}^2.
\]
On the other hand, smoothness of $\bal$ implies
\begin{align}
&|\Fc(\bal',\bt_\bal)-\Fc(\bal,\bt_\bal)|\leq \frac{L}{2}\tn{\bal-\bal'}^2.
\end{align}
Putting these together,
\[
\Fc(\bal,\bt_\bal)+\frac{L}{2}\tn{\bal-\bal'}^2\geq \Fc(\bal',\bt_\bal)\geq \Fc(\bal',\bt_{\bal'})+\frac{\mu}{2}\tn{\bt_\bal-\bt_{\bal'}}^2.
\]
Since the inequality is symmetric with respect to $\bal,\bal'$, we also have
\begin{align}
\Fc(\bal',\bt_{\bal'})+\frac{L}{2}\tn{\bal-\bal'}^2\geq\Fc(\bal,\bt_\bal)+\frac{\mu}{2}\tn{\bt_\bal-\bt_{\bal'}}^2.
\end{align}
Summing up both sides yield the desired result
\[
L\tn{\bal-\bal'}^2\geq \mu\tn{\bt_\bal-\bt_{\bal'}}^2.
\]
\end{proof}
\subsection{Stability of Linear Regression}
We begin by stating the following useful lemma.
\begin{lemma} \label{inverse diff}Let $\A$ and $\B$ be two positive semidefinite matrices with minimum eigenvalue of $\A$ bounded below by $\gamma>0$. Set $\Pb=\B-\A$ and suppose $\|\Pb\|\leq \delta$. Then
\begin{align}
\|\A^{-1}-\B^{-1}\|\leq \frac{\delta}{\gamma(\gamma-\delta)}.\label{ab gap}
\end{align}
\end{lemma}
\begin{proof} Let $\A$ and $\B$ be two positive semidefinite matrices with minimum eigenvalue of $\A$ bounded below by $\gamma>0$. Set $\Pb=\B-\A$ and suppose $\|\Pb\|\leq \delta$. Let $\A$ have eigen decomposition $\Ub\bSi\Ub^T$. Let $\bar{\Pb}=\Ub^T\Pb\Ub$ and $\tilde{\Pb}=\bSi^{-1/2}\Ub^T\Pb\Ub\bSi^{-1/2}$. Observe that $\|\tilde{\Pb}\|\leq \gamma^{-1}\delta$. We have that
\begin{align*}
\|\A^{-1}-\B^{-1}\|&=\|\bSi^{-1}-\Ub^T\B^{-1}\Ub\|\\
&=\|\bSi^{-1}-\Ub^T(\A+\Pb)^{-1}\Ub\|\\
&=\|\bSi^{-1}-\Ub^T(\Ub\bSi\Ub^T+\Ub\bar{\Pb}\Ub^T)^{-1}\Ub\|\\
&=\|\bSi^{-1}-(\bSi+\bar{\Pb})^{-1}\|\\
&\leq \|\bSi^{-1}-(\bSi+\bar{\Pb})^{-1}\|\\
&\leq \|\bSi^{-1/2}(\Iden-(\Iden+\bSi^{-1/2}\bar{\Pb}\bSi^{-1/2})^{-1})\bSi^{-1/2}\|\\
&\leq \sigma^{-1}_{\min}(\bSi)\|\Iden-(\Iden+\tilde{\Pb})^{-1}\|\\
&\leq \frac{1}{\gamma}\frac{\gamma^{-1}}{1-\delta\gamma^{-1}}=\frac{\delta}{\gamma(\gamma-\delta)}.
\end{align*}
For the last statement, we observe the fact that $(\Iden+\tilde{\Pb})^{-1}=\sum_{i\geq 0}(-1)^i\tilde{\Pb}^i$ which yields $\|\Iden-(\Iden+\tilde{\Pb})^{-1}\|\leq \frac{\gamma^{-1}\delta}{1-\gamma^{-1}\delta}$. 
\end{proof}
The lemma below shows the stability of ridge(less) regression to feature or regularizer changes.
\begin{lemma} [Feature Robustness of Linear Models]\label{lem robust}Fix $\X,\Xb\in\R^{n\times p}$ with $\|\X\|,\|\Xb\|\leq B$. Fix $\la,\lab\geq0$ and assume $\laz=\la+\sigma_{\min}^2(\X)>0$. Suppose $2B\|\X-\Xb\|+|\la-\lab|< \laz/2$. Consider the ridge regression (or min Euclidean solution)
\[
\bt=\Acr(\y,\X,\la)=\begin{cases}(\X^T\X+\la\Iden)^{-1}\X^T\y\quad\text{if}\quad\la>0~\text{or}~n\geq p\\\X^T(\X\X^T)^{-1}\y\quad\quad~~~\text{if}\quad\la=0~\text{and}~n<p\end{cases}.
\]
Also define $\btb=\Acr(\y,\Xb,\lab)$ which solves the problem with features $\Xb$.
In either cases (whether $\la>0$ or not), we have that
\begin{align}
\tn{\bt-\btb}\leq \lip\|\X-\Xb\|+\frac{2B\tn{\y}}{\laz^2}|\la-\lab|.\label{robust bound}
\end{align}
\end{lemma}
\begin{proof} Suppose $\la>0$ or $n>p$. Recall that $\X^T\X+\la\Iden\succeq \laz\Iden$. Observe that $\X^T\X-\Xb^T\Xb=\X^T(\X-\Xb)+(\X-\Xb)^T\Xb$ which implies $\|\X^T\X+\la\Iden-\Xb^T\Xb+\lab\Iden\|\leq \bB$ where $\bB=2\eps B+|\la-\lab|$. Recall that $\bB\leq \laz/2$. Using this and the result \eqref{ab gap} above, we can bound
\begin{align}
\|(\Xb^T\Xb+\lab\Iden)^{-1}-(\X^T\X+\la\Iden)^{-1}\|&\leq \frac{2B\eps+|\la-\lab|}{\laz(\laz -(2B\eps+|\la-\lab|))}\\
&\leq \frac{\bB}{\laz(\laz-2B\eps)}\\
&\leq \frac{2\bB}{\laz^2}
\end{align}
We then find
\begin{align*}
\tn{\btb-\bt}&\leq \|(\Xb^T\Xb+\la\Iden)^{-1}\Xb^T-(\X^T\X+\la\Iden)^{-1}\X^T\|\tn{\y}\\
&\leq (\|(\Xb^T\Xb+\la\Iden)^{-1}\Pb^T\|+\|(\Xb^T\Xb+\la\Iden)^{-1}-(\X^T\X+\la\Iden)^{-1}\|\|\X\|)\tn{\y}\\
&\leq (\eps \laz^{-1}+\frac{2B\bB}{\laz^2})\tn{\y}\\
&\leq \lip\eps+\frac{2B\tn{\y}}{\laz^2}|\la-\lab|.
\end{align*}
Similarly when $\la=0$ and $n<p$, using $\X\X^T\succeq \laz\Iden$
\begin{align*}
\|(\Xb\Xb^T)^{-1}-(\Xb\Xb^T)^{-1}\|&\leq \frac{2B\eps}{\sigma_{\min}^2(\X)(\sigma_{\min}^2(\X)-\|\X\X^T-\Xb\Xb^T\|)}\\
&\leq \frac{4B\eps}{\laz^2}
\end{align*}
Thus, in an essentially identical fashion, we find
\begin{align*}
\tn{\btb-\bt}&\leq \|\Xb^T(\Xb\Xb^T)^{-1}-\X^T(\X\X^T)^{-1}\|\tn{\y}\\
&\leq (\|(\Xb-\X)^T(\Xb\Xb^T)^{-1}\|+\|\X^T\|\|(\Xb\Xb^T)^{-1}-(\X\X^T)^{-1}\|)\tn{\y}\\
&\leq \lip\eps.
\end{align*}
This finishes the proof of $\lip$ Lipschitzness.

Finally, let $\X=\Ub\La\V^T$ be the singular value decomposition with $i$th singular value $\sigma_i$. Suppose $\la=0$, $\sigma_i^2\geq \laz$ and $\lab\neq 0$. We consider the gap
\begin{align*}
\|(\X^T\X+\lab\Iden)^{-1}\X^T-\X^T(\X\X^T)^{-1}\|&=\|\V(\La^2+\bar{\la}\Iden)^{-1}\La\Ub^T-\V\La^{-1}\Ub^T\|\\
&\leq \sup_{1\leq i\leq n}|\frac{\la_i}{\la_i^2+\bar{\la}}-\frac{\la_i}{\la_i^2}|\leq \frac{\bar{\la}}{\laz^2}.
\end{align*}
Thus, using this bound and the triangle inequality, for the scenario $\la=0$ and $\lab\neq 0$, we again find the desired bound.
\end{proof}

\subsection{Lipschitzness of Feature Maps}
\begin{lemma} [Lipschitzness of Feature Map Solutions]\label{lip fmap lemma} Let $\sup_{\x\in\Xc,1\leq i\leq \h}\tn{\phi_i(\x)}^2\leq B$. Let $\la\geq 0$ be the strength of regularization. Suppose $n\leq p$ and $\la+\inf_{\bal\in\Bal}\sigma^2_{\min}(\bPhi_\bal)\geq \laz>0$. Set $R=\sup_{\bal\in\Bal}\tone{\bal}$. Suppose $\Bal$ is convex and the Algorithm $\Ac$ solves the ridge regression given $\bPhi_\bal$ i.e.
\[
\bt_\bal=\Acr(\y,\bPhi_\bal,\la).
\]
Then, $\bt_\bal$ is $\lipp$-Lipschitz function of $\bal$ (in $\ell_2$ norm) and Assumption \ref{algo lip} holds with $L=\lips$.

Additionally, consider $\bt_\bal=\Acr(\y,\bPhi_\bal,\bal_{h+1})$ where $\bal_{h+1}\in[\la,\la_{\max}]$ where $\la+\inf_{\bal\in\Bal}\sigma^2_{\min}(\bPhi_\bal)\geq \laz$. Then Assumption \ref{algo lip} holds with 
\begin{align}
L=\lips+2R^2B\sqrt{\nt}\laz^{-2}\tn{\y}.\label{ridge lip}
\end{align}
\end{lemma}
\begin{proof} Observe that $\tn{\phi_\bal(\x)}\leq R\sqrt{B}$ which also implies $\|\bPhi\|\leq \bar{B}=R\sqrt{B\nt}$. Fix $\bal,\bal'\in\Bal$ satisfying $\tn{\bal-\bal'}=\eps< \frac{\laz}{4RB\nt}$ where $\eps>0$ is to be determined. This implies that 
\begin{align}
&\tn{\phi_\bal(\x)-\phi_{\bal'}(\x)}=\tn{\sum_{i=1}^\h (\bal_i-\bal'_i)\phi_i(\x)}\leq \tone{\bal-\bal'}\sqrt{B}\\
&\implies \|\bPhi_\bal-\bPhi_{\bal'}\|\leq \sqrt{\nt}\tone{\bal-\bal'}\sqrt{B}\leq \sqrt{\nt \h}\tn{\bal-\bal'}\sqrt{B}\leq \eps\sqrt{\nt \h B}.
\end{align}
Using the fact that problem is $\la_0$-strongly convex, applying Lemma \ref{lem robust} with $\laz$, and observing that the initial choice of $\eps$ implies $\|\bPhi_\bal-\bPhi_{\bal'}\|\leq \eps\sqrt{\nt \h B}<\frac{\laz}{4R\sqrt{B \nt}}$, we find that
\[
\frac{\tn{\bt_\bal-\bt_{\bal'}}}{\|\bPhi_\bal-\bPhi_{\bal'}\|}\leq {5\bar{B}^2\tn{\y}}/\laz^2=5R^2B{\nt}\tn{\y}/\laz^2.
\]
This also implies the Lipschitzness $\frac{\tn{\bt_\bal-\bt_{\bal'}}}{\tn{\bal-\bal'}}\leq \bar{L}:=\lipp$. Observe that this argument showed the desired Lipschitzness around $\bal$ in a ball of radius $\eps>0$. However, this same local Lipschitzness holds for any choice of $\bal\in\Bal$. Since $\Bal$ is convex, local Lipschitzness implies global (as we can draw a straightline between any two points and repeatedly apply local Lipschitzness).

Let $\bt_\bal$ be $\bar{L}$ Lipschitz function of $\bal$ (which is provided above). To conclude with Assumption \ref{algo lip}, we need Lipschitzness over the input space. Observe that $\tn{\bt_\bal}\leq \laz^{-1/2}\tn{\y}$. Consequently, using the fact $\clr{R^2B(\nt-1)\geq \la}$ (which implies $\clr{R^2B\nt\geq \laz}$), we find
\begin{align*}
|f_\bal(\x)-f_{\bal'}(\x)|&\leq |\bt_\bal^T\phi_\bal(\x)-\bt_{\bal'}^T\phi_{\bal'}(\x)|\\
&\leq R\sqrt{B}\bar{L}\tn{\bal-\bal'}+\tn{\bt_\bal}\tn{\phi_\bal(\x)-\phi_{\bal'}(\x)}\\
&\leq (R\sqrt{B}\bar{L}+\tn{\y}\laz^{-1/2}\sqrt{\h B})\tn{\bal-\bal'}\\
&\leq \sqrt{B}(R\bar{L}+\tn{\y}\laz^{-1/2}\sqrt{\h })\tn{\bal-\bal'}\\
&\leq \lips\tn{\bal-\bal'},
\end{align*}
concluding the first result.

To show the second point, using the exact same argument and applying \eqref{robust bound}, we find
\begin{align}
\frac{\tn{\bt_\bal-\bt_{\bal'}}}{\tn{\bal-\bal'}}\leq \lipp +2R\sqrt{B\nt}\laz^{-2}\tn{\y}.\nn
\end{align}
Repeating the input space argument, we find the advertised bound.
\end{proof}

\section{Proofs for Neural Feature Maps (Theorem \ref{gen thm generic})}\label{deep fmap}

The following theorem is a restatement of Theorem \ref{gen thm generic}. The main difference is that conditions \eqref{conditions} on $\defz,\wi$ are more precise versions compared to Theorem \ref{gen thm generic}.
\begin{theorem} [Restatement of Theorem \ref{gen thm generic}]Suppose Assumptions \ref{lip ker} and \ref{nn conc} hold and we solve feature map regression (Def.~\ref{fmap def}) with neural feature maps \eqref{neural fmap} upper bounded by $\sqrt{B}$ in $\ell_2$-norm. Define the set
\[
\Bal_0=\{\bal\in\Bal\bgl \Kb_\bal\geq \laz\}.
\]
Let loss $\ell$ be $\Gamma$-Lipschitz and bounded by a constant. Suppose the following conditions on $\wi,\la$ hold and define $\defz$ as
\begin{align}\label{conditions}
&\defz=\deff\log(4\bC L(\la^{-2}B\Gamma+\sqrt{B}) \nt^3) \quad\text{,}\\
& \wi\gtrsim \eps^{-4}\la_0^{-4}\Gamma^4 B^2\nu(\deff\log(2L\bC\wi/(\nu\deff))+t)  \quad\text{,}\nn\\
& \la\leq \frac{\eps\laz^2}{4\sqrt{B\nt}}.\nn
\end{align}
Then, with probability at least $1-5e^{-t}$, for some constant $C>0$
\begin{align}\label{bound for cond}
\Lc(\ft_\bah)\leq \min_{\bal\in\Bal_0}2\Gamma\sqrt{\frac{B\y^T\Kb_\bal^{-1}\y}{\nt}}+C\sqrt{\frac{\defz+\tau}{\nt}}+\eps+\del.
\end{align}
Finally, this result also applies to the $0-1$ loss $\Lcz$ by setting $\Gamma=1$. To see this, choose $\ell$ to be the Hinge loss and note that it dominates the $0-1$ loss.
\end{theorem}
\begin{proof} For a new input example $\x$, define the vector $\kappa_\bal(\x)=k_\bal(\x_i,\x)$ where $k_\bal$ is the empirical kernel function associated with neural feature map at initialization. By assumption $\kappa_\bal$ is $L$-Lipschitz in $\ell_2$ norm. 

First we settle a couple of bounds.
\begin{enumerate}
\item Set $\eps_0\leq \laz/2$ and note that using Assumption \ref{nn conc} we have
\begin{align}
\wi\geq 4\eps_0^{-2}\nu(\deff\log(2L\bC\wi/(\nu\deff))+t).\label{k low bound}
\end{align}
With probability at least $1-e^{-t}$, for all $\bal$, $\|\Kb_\bal-\Kbh_\bal\|\leq \eps_0$.
\item Let $f_\bal^\la$ be the solution with ridge regularization $0<\la\leq \la_0/2$ and $f_\bal^0$ be the ridgeless solution. For any input $\x\in\Xc$, using Lemma \ref{inverse diff}, we have that
\begin{align}
|f_\bal^\la(\x)-f_\bal^0(\x)|&=|\y\Kb_\bal^{-1}\kappa_\bal(\x)-\y(\Kb_{\bal}+\la\Iden)^{-1}\kappa_{\bal'}(\x)|\\
&\leq\tn{\y}\|\Kb_\bal^{-1}-(\Kb_{\bal}+\la\Iden)^{-1}\|\tn{\kappa_{\bal}(\x)}\\
&\leq \frac{2\sqrt{Bn}\la}{\laz^2}.\label{ridge diff}
\end{align}
\item Next, we apply Theorem \ref{thm val}. First note that (for sufficiently small $\bal-\bal'$)
\[
\|(\Kbh_\bal+\la\Iden)^{-1}-(\Kbh_{\bal'}+\la\Iden)^{-1}\|\leq \frac{L}{\la^2}\tn{\bal-\bal'}.
\]
The Lipschitz constant of $\Ac$ can be found via 
\begin{align*}
|\y(\Kbh_\bal+\la\Iden)^{-1}\kappa_\bal(\x)&-\y(\Kbh_{\bal'}+\la\Iden)^{-1}\kappa_{\bal'}(\x)|\leq\tn{\y}\tn{(\Kbh_\bal+\la\Iden)^{-1}\kappa_\bal(\x)-(\Kbh_{\bal'}+\la\Iden)^{-1}\kappa_{\bal'}(\x)}\\
&\leq \tn{\y}\|(\Kbh_\bal+\la\Iden)^{-1}-(\Kbh_{\bal'}+\la\Iden)^{-1}\|\tn{\kappa_\bal(\x)}+\|\Kbh_{\bal'}^{-1}\|\tn{\kappa_\bal(\x)-\kappa_{\bal'}(\x)}\\
&\leq (\frac{2L\tn{\y}}{\la^2}\tn{\kappa_\bal(\x)}+\frac{L\tn{\y}}{\la})\tn{\bal-\bal'}\\
&\leq \frac{2LB\tn{\y}\sqrt{\nt}}{\la^2}\eps\\
&\leq \frac{2LB\nt}{\la^2}\eps\\
\end{align*}
Thus Theorem \ref{thm val} yields with probability $1-2e^{-t}$
\begin{align}
\Lc(\ft_\bah)\leq \inf_{\bal\in \Bal}\Lc(\ft_\bal)+ C\sqrt{\frac{\deff\log(4\la^{-2}\bC LB\Gamma \nt\nv)+\tau}{\nv}}+\del.\label{excess risk val3}
\end{align}
\item Finally, applying Theorem \ref{unif excess lem} on the ridgeless kernel estimators (trained with $\Kbh_\bal$) $f'_\bal$ over the set $\Bal_0$, we obtain that for each $\bal\in\Bal_0$
\[
\Lc(f'_\bal)\leq  \frac{2\Gamma \sqrt{B\y^T\Kbh_\bal^{-1}\y}+C\sqrt{t}}{\sqrt{\nt}}.
\]
Observe that for $\bal\in\Bal_0$
\begin{align}
|\sqrt{\y^T\Kbh_\bal^{-1}\y}-\sqrt{\y^T\Kbh_{\bal'}^{-1}\y}|&\leq \frac{|\y^T\Kbh_\bal^{-1}\y-\y^T\Kbh_{\bal'}^{-1}\y|}{\sqrt{\y^T\Kbh_\bal^{-1}\y}}\leq \sqrt{B\nt} \frac{|\y^T\Kbh_\bal^{-1}\y-\y^T\Kbh_{\bal'}^{-1}\y|}{\tn{\y}}\\
&\leq \sqrt{B\nt}\tn{\y}{\|\Kbh_{\bal}^{-1}-\Kbh_{\bal'}^{-1}\|}\\
&\leq \frac{2\sqrt{B\nt }L\tn{\y}}{\laz^2}\sqrt{\tn{\bal-\bal'}}
\end{align}
Same argument also gives 
\begin{align}
|\sqrt{\y^T\Kbh_\bal^{-1}\y}-\sqrt{\y^T\Kb_{\bal}^{-1}\y}|\leq \sqrt{\y^T(\Kbh_\bal^{-1}-\Kb_\bal^{-1})\y}\leq \frac{\tn{\y}\sqrt{2\eps_0}}{\la_0}.\label{inv bound}
\end{align}

Applying Lemma \ref{unif excess lem} with $\kappa=\frac{2\sqrt{B}\nt\tn{\y}}{\laz^2}$ and using \eqref{inv bound}, we find with probability $1-2e^{-t}$ that 
\begin{align}
\min_{\bal\in\Bal}\Lc(f'_\bal)&\leq\min_{\bal\in\Bal_0}\frac{2\Gamma \sqrt{B\y^T\Kbh_\bal^{-1}\y}+C\sqrt{{\deff\log(2\bC\sqrt{B}\nt^3L)+\tau}}}{\sqrt{\nt}}\\
&\leq \min_{\bal\in\Bal_0}\frac{2\Gamma \sqrt{B\y^T\Kbh_\bal^{-1}\y}+C\sqrt{{\deff\log(2\bC\sqrt{B}\nt^3L)+\tau}}}{\sqrt{\nt}}+3\Gamma \frac{\sqrt{B\eps_0}}{\la_0}.
\end{align}
\end{enumerate}
To stitch the results together, observe that with probability $1-5e^{-t}$ all events hold. Then, using \eqref{ridge diff} and right above, we obtain
\begin{align}
\min_{\bal\in\Bal} \Lc(f_\bal)&\leq\min_{\bal\in\Bal} \Lc(f'_\bal) +\frac{2\sqrt{B\nt}\la}{\laz^2}\nn\\
&\leq\min_{\bal\in\Bal_0} \frac{2\Gamma \sqrt{B\y^T\Kb_\bal^{-1}\y}+C\sqrt{{\deff\log(2\bC\sqrt{B}\nt^3L)+\tau}}}{\sqrt{\nt}}+3\Gamma \frac{\sqrt{B\eps_0}}{\la_0}+\frac{2\sqrt{B\nt}\la}{\laz^2}.\label{d15}
\end{align}
Now fix $\eps>0$ and set $\la\leq \frac{\eps\laz^2}{4\sqrt{B\nt}}$ and $\eps_0=\frac{\eps^2 \la_0^2}{36\Gamma^2 B}$. The sum $3\Gamma \frac{\sqrt{B\eps_0}}{\la_0}+\frac{2\sqrt{B\nt}\la}{\laz^2}$ on the right hand-side of \eqref{d15} is upper bounded by $\eps$ as soon as the conditions \eqref{conditions} hold (recall \eqref{k low bound} for $\wi$ bound). Under these conditions, combining \eqref{d15} with the validation bound \eqref{excess risk val3}, yields the following upper bound
\[
\min_{\bal\in\Bal_0} \frac{2\Gamma \sqrt{B\y^T\Kb_\bal^{-1}\y}+C\sqrt{{\deff\log(2\bC\sqrt{B}\nt^3L)+\tau}}}{\sqrt{\nt}}+ C\sqrt{\frac{\deff\log(4\la^{-2}\bC LB\Gamma \nt\nv)+\tau}{\nv}}+\del+\eps.
\]
This is equivalent to \eqref{bound for cond} after plugging in the definition of $\defz$ and recalling $\nt\geq\nv$.
\end{proof}

\subsection{Uniform Convergence to Population Neural Kernel}
\begin{lemma} [Uniform Convergence to NTK]\label{lem sup simple2} Suppose Assumptions \ref{lip ker} and \ref{nn conc} hold. With probability at least $1-e^{-t}$ over initialization, for all $\bal$, $\|\Kb_\bal-\Kbh_\bal\|\leq 2\sqrt{\frac{\nu\deff\log(2L\bC\wi/(\nu\deff))+t}{\wi}}$.
\end{lemma}
{\begin{proof} Fix $\bal,\bal'$ such that $\tn{\bal-\bal'}= \eps/2L$ and $\Cc_\eps$ be the cover set of size $\log|\Cc_\eps|\leq \deff \log(2L\bC/\eps)$. We find that $\|\Kb_\bal-\Kbh_\bal\|\leq \eps/2$. Now, using the concentration bound \eqref{K conc} and setting $\tau=\deff\log(2L\bC/\eps)+t$, we obtain that with probability $1-e^{-t}$, all $\bal\in\Cc_\eps$ obeys $\|\Kbh_\bal-\Kb_\bal\|\leq \sqrt{\frac{\nu\deff\log(2L\bC/\eps)+\nu t}{\wi}}$. For $\bal\in\Bal$, picking a neighbor $\bal'\in\Cc_\eps$, via triangle inequality, we find
\begin{align*}
&\|\Kb_\bal-\Kbh_\bal\|\\
&\leq \|\Kb_\bal-\Kb_{\bal'}\|+\|\Kb_{\bal'}-\Kbh_{\bal'}\|+\|\Kbh_{\bal'}-\Kbh_\bal\|\\
&\leq \sqrt{\frac{\nu\deff\log(2L\bC/\eps)+\nu t}{\wi}}+\eps.
\end{align*}
Setting $\eps=\sqrt{\nu\deff/\wi}$, we conclude with the result.
\end{proof}}
\section{Proof of Lemma \ref{lemma deep act} and Structure of the Jacobian of a Deep Neural Net}\label{deep j sec}

\subsection{Expression for Jacobian}
In this section we wish to bound the Lipschitzness of the Jacobian matrix with respect to the architecture parameters $\bal$. To this aim we begin by considering the structure of the Jacobian. Specifically, assume we have $n$ data points $\x_i\in\R^d$ for $i=1,2,\ldots,n$. We shall use 
\begin{align*}
\mtx{X}=\begin{bmatrix}\x_1^T \\\x_2^T \\\vdots \\\x_n^T\end{bmatrix}\in\R^{n\times d} 
\end{align*}
for the data matrix.

Set $\vb:=\W^{(D+1)}$. Assume a neural network mapping $f:\R^d\rightarrow \R$ with $D$ hidden layers of the form
\begin{align*}
f(\vb,\W^{(1)},\W^{(2)},\ldots,\W^{(D)},\x)=\vb^T\sbl{D}\left(\W^{(D)}\sbl{D-1}\left(\W^{(D-1)}\dots\sbl{1}\left( \W^{(1)}\x\right)\right)\right)
\end{align*}
with $\W^{(r)}\in\R^{d_r\times d_{r-1}}$ where $d_0=d$ and $d_D=k$. We also define the hidden unit vector 
\begin{align*}
\hh^{(1)}(\x)=\sbl{1}\left(\W^{(1)}\x\right)
\end{align*}
for the first layer and inductively for the remaining layers via
\begin{align*}
\hh^{(r)}(\x)=\sbl{r}\left(\W^{(r)}\hh^{(r-1)}(\x)\right)
\end{align*}
for $r=2,3,\ldots,D$. 

The Jacobian with respect to the weights of the last layer is equal to
\begin{align*}
\Jc^T\left(\vb\right)=\sbl{D}\left(\W^{(D)}\sbl{D-1}\left(\W^{(D-1)}\ldots\sbl{1}\left( \W^{(1)}\mtx{X}^T\right)\right)\right)
\end{align*}
which can also be rewritten as
\begin{align*}
\Jc\left(\vb\right)=\begin{bmatrix}
\sbl{D}\left(\hh^{(D)}(\x_1)\right)^T\\
\sbl{D}\left(\hh^{(D)}(\x_2)\right)^T\\
\vdots\\
\sbl{D}\left(\hh^{(D)}(\x_n)\right)^T\\
\end{bmatrix}
\end{align*}
With respect to $\W^{(D)}$ the columns take the form
\begin{align*}
\Jc^T\left(\W^{(D)}\right)=\text{diag}\left(\vb\right)\sbl{D}'\left(\W^{(D)}\mtx{X}\right) *\hh^{(D)}\left(\mtx{X}^T\right)
\end{align*}
To calculate the derivative with respect to $\W^{(r)}$ first note that
\begin{align*}
f\left(\vb,\W^{(1)},\W^{(2)},\ldots,\W^{(D)},\x\right)=\vb^T\sbl{D}\left(\W^{(D)}\sbl{D-1}\left(\W^{(D-1)}\ldots\sbl{1}\left( \W^{(r)}\hh^{(r-1)}(\x)\right)\right)\right).
\end{align*}
Furthermore, note that
\begin{align*}
f\left(\vb,\W^{(1)},\W^{(2)},\ldots,\W^{(D)},\x\right)=\vb^T\sbl{D}\left(\W^{(D)}\sbl{D-1}\left(\W^{(D-1)}\ldots\sbl{r}\left( g\left(\W^{(r)}\right)\right)\right)\right).
\end{align*}
where $g\left(\W^{(r)}\right)=\W^{(r)}h^{(r-1)}(\x)$. Using the chain rule we have
\begin{align*}
\mathcal{D}_f\left(\W^{(r)}\right)=&\left(\vb^T\prod_{s=r+1}^D\text{diag}\left(\sbl{s}'\left(\hh^{(s)}(\x)\right)\right)\W^{(s)}\right)\text{diag}\left(\sbl{r}'\left(\hh^{(r)}(\x)\right)\right)\mathcal{D}_g\left(\W^{(r)}\right)\\
=&\left(\vb^T\prod_{s=r+1}^D\text{diag}\left(\sbl{s}'\left(\hh^{(s)}(\x)\right)\right)\W^{(s)}\right)\text{diag}\left(\sbl{r}'\left(\hh^{(r)}(\x)\right)\right)\left(\mtx{I}_{d^{(r)}}\otimes \left(h^{(r-1)}(\x)\right)^T\right)
\end{align*}
Therefore,
\begin{align*}
\Jc\left(\W^{(r)}\right)=
\begin{bmatrix}
\left(\vb^T\prod_{s=r+1}^D\text{diag}\left(\sbl{s}'\left(\hh^{(s)}(\x_1)\right)\right)\W^{(s)}\right)\text{diag}\left(\sbl{r}'\left(\hh^{(r)}(\x_1)\right)\right)\left(\mtx{I}_{d^{(r)}}\otimes \left(\hh^{(r-1)}(\x_1)\right)^T\right)\\
\left(\vb^T\prod_{s=r+1}^D\text{diag}\left(\sbl{s}'\left(\hh^{(s)}(\x_2)\right)\right)\W^{(s)}\right)\text{diag}\left(\sbl{r}'\left(\hh^{(r)}(\x_2)\right)\right)\left(\mtx{I}_{d^{(r)}}\otimes \left(\hh^{(r-1)}(\x_2)\right)^T\right)\\
\vdots\\
\left(\vb^T\prod_{s=r+1}^D\text{diag}\left(\sbl{s}'\left(\hh^{(s)}(\x_n)\right)\right)\W^{(s)}\right)\text{diag}\left(\sbl{r}'\left(\hh^{(r)}(\x_n)\right)\right)\left(\mtx{I}_{d^{(r)}}\otimes \left(\hh^{(r-1)}(\x_n)\right)^T\right)
\end{bmatrix}
\end{align*}
For simplicity we shall use the short-hand
\begin{align*}
\Jc^{(\ell)}:=\Jc\left(\W^{(\ell)}\right),
\end{align*}
with $\Jc^{(D+1)}=\Jc(\vb)$. We shall also use $\Jc_i^{(\ell)}$ to denote the $i$-th row of $\Jc^{(\ell)}$. Finally, define the overall Jacobian to be
\[
\Jc=\bPhi_\bal=[\Jc^{(D+1)}~\Jc^{(D)}~\dots~\Jc^{(1)}]\in\R^{n\times p}
\]
Additionally, define the gram matrix as $\Kbh=\Jc\Jc^T=\bPhi_\bal\bPhi_\bal^T$. For the discussion below set $S_\ell=\|\W^{(\ell)}\|$, define the quantities
\begin{align}
&M=\prod_{\ell=1}^{D+1}S_\ell,\quad M_+^i=\prod_{\ell=1}^{i}S_\ell,\quad M_-^i=\prod_{\ell=i}^{D+1} S_\ell.
\end{align}
Also suppose $\tn{\x}\leq N$ for some $N\geq \sqrt{d}$ for all feasible inputs $\x\in\Xc$. Observe that, if the activations are zero-mean using $|\phi'_\bal|\leq B$,
\[
\tn{\hh^{(\ell)}}\leq B^\ell M_+^\ell N.
\]
Define $k_0=d$ and let $k_\ell$ be the width of the $\ell$th layer. Define 
\[
\bar{S}_i=S_i+\clr{\max(1,\sqrt{k_i/k_{i-1}})}.
\] In general, since $|\phi_\bal(0)|\leq B$, we have that
\begin{align}
\tn{\hh^{(\ell)}}\leq B^\ell \bar{M}_+^\ell N.\label{hell bound}
\end{align}
where 
\[
\bar{M}=\prod_{\ell=1}^{D+1}\bar{S}_\ell,\quad \bar{M}_+^\ell =\prod_{i=1}^{\ell}\bar{S}_i,\quad \bar{M}_-^\ell =\prod_{i=\ell}^{D+1}\bar{S}_i.
\]
\subsection{Upper Bounding Jacobian}
Before Lipschitzness, let us upper bound the Jacobian spectral norm. Clearly given two activation parameters $\bal,\bab$, noting $\Jc$ is concatenation of $\Jc^{(\ell)}$'s, Jacobian's obey
\begin{align}
&\|\Jc_\bal\|\leq \sqrt{\sum_{\ell=1}^{D+1}\|\Jc^{(\ell)}_\bal\|^2}\leq \sqrt{2D}\max_{1\leq \ell\leq D+1}\|\Jc^{(\ell)}_\bal\|\\
&\|\Jc_\bal-\Jc_\bab\|\leq \sqrt{D+1}\max_{1\leq \ell\leq D+1}\|\Jc^{(\ell)}_\bal-\Jc^{(\ell)}_\bab\|\\
&\quad\quad\quad\quad\quad\leq \sqrt{2Dn}\max_{1\leq \ell,i\leq D+1}\|\Jc^{(\ell)}_{i,\bal}-\Jc^{(\ell)}_{i,\bab}\|.\label{j diff bound}
\end{align} 
Additionally denoting $\Jc:=\Jc_\bal$ observe that
\begin{align}
\|\Jc^{(\ell)}\|&\leq \sqrt{n}B^{D-\ell+1}M_-^{\ell+1} \times \tn{\hh^{(\ell-1)}}\leq \sqrt{n}B^{D-\ell+1}M_-^{\ell+1} \times B^{\ell-1} \bar{M}_+^{\ell-1} N\\
&\leq \sqrt{n}\mult.
\end{align}
Thus, we find that
\begin{align}
\|\Jc\|\leq \sqrt{2Dn}\mult.\label{j bound m1}
\end{align}

\subsection{Lipschitzness of the Gram Matrix}

Now that we have an expression for the Jacobian matrices with respect to different weights we wish to bound the Lipshitzness with respect to $\bal$. Let $\bal,\bab$ be two activation choices. Observe via \eqref{j bound m1} that
\begin{align}
\|\Jc_\bal\Jc_\bal^T-\Jc_\bab\Jc_\bab^T\|&\leq 2\sqrt{2Dn}\mult\|\Jc_\bal-\Jc_\bab\|.\label{kernel diff}
\end{align}
The right side will be bounded via \eqref{j diff bound}. Thus, to proceed, we will consider the derivative of $\Jc^{(\ell)}_i$ (fixing $\bal$). Using the chain rule we have
\begin{align*}
\frac{\partial \Jc_i^{(\ell)}}{\partial \bal}=\sum_{s=\ell-1}^D \frac{\partial \Jc_i^{(\ell)}}{\partial \hh^{(s)}(\x_i)}\frac{\partial \hh^{(s)}(\x_i)}{\partial \bal} 
\end{align*}
Thus, using the triangular inequality
\begin{align}
\opnorm{\frac{\partial \Jc_i^{(\ell)}}{\partial \bal}}\le \sum_{s=\ell-1}^D \opnorm{\frac{\partial \Jc_i^{(\ell)}}{\partial \hh^{(s)}(\x_i)}}\opnorm{\frac{\partial \hh^{(s)}(\x_i)}{\partial \bal} }\le D\max_{\ell-1\leq s\leq D} \opnorm{\frac{\partial \Jc_i^{(\ell)}}{\partial \hh^{(s)}(\x_i)}}\opnorm{\frac{\partial \hh^{(s)}(\x_i)}{\partial \bal} }.\label{j bound 0}
\end{align}
Now note that based on the structure of the Jacobian discussed above and using the fact that the first and second order derivatives of the activations are bounded by $B\clr{\geq 1}$ we have
\begin{align}
&\opnorm{\frac{\partial \Jc_i^{(\ell)}}{\partial \hh^{(\ell-1)}(\x_i)}}\le B^{D-\ell+1}\prod_{u=\ell+1}^{D+1}\opnorm{\W^{(u)}}\leq \mult,\quad\quad\quad\text{for}\quad\quad s=\ell-1\\
&\opnorm{\frac{\partial \Jc_i^{(\ell)}}{\partial \hh^{(s)}(\x_i)}}\le B^{D-\ell+1}\left(\prod_{u=\ell+1}^{D+1}\opnorm{\W^{(u)}}\right)\twonorm{\hh^{(\ell-1)}(\x_i)}\le \mult\quad\quad\quad\text{for}\quad\quad s\ge\ell,\\
&\opnorm{\frac{\partial \Jc_i^{(\ell)}}{\partial \hh^{(s)}(\x_i)}}=0\quad\quad\quad\text{for}\quad\quad s< \ell-1\label{j bound 1}.
\end{align}
To proceed, we need to bound the remaining term $\|\frac{\partial \hh^{(s)}(\x_i)}{\partial \bal}\|$. We can write
\begin{align}
\|\frac{\partial \hh^{(s)}(\x_i)}{\partial \bal}\|&\leq \sqrt{\sum_{\ell=1}^s\|\frac{\partial \hh^{(s)}(\x_i)}{\partial \bl{\ell}}\|^2}\\
&\leq \sqrt{D}\max_{1\leq\ell\leq s}\|\frac{\partial \hh^{(s)}(\x_i)}{\partial \hh^{(\ell)}(\x_i)}\|\|\frac{\partial \hh^{(\ell)}(\x_i)}{\partial \bl{\ell}}\|\\
&\leq \sqrt{D}\mult\sqrt{h}.\label{j bound 2}
\end{align}
To see this, define $\hb^{(\ell)}=\W^{(\ell)}\hh^{(\ell-1)}$ and note $\hh^{(\ell)}=\sbl{\ell}(\hb^{(\ell)})$. Now observe that $\frac{\partial \hh^{(\ell)}(\x_i)}{\partial \bl{\ell}}$ has the following clean form, with columns that are bounded via \eqref{hell bound} 
\[
\frac{\partial\hh^{(\ell)}(\x)}{{\partial \bl{\ell}}}=\left[\phi_1(\hb^{(\ell)})~\dots~\phi_h(\hb^{(\ell)})\right]\in\R^{k_\ell\times h}.
\]
Combining \eqref{j bound 1} with \eqref{j bound 2} and plugging in \eqref{j bound 0}, we obtain
\begin{align}
\opnorm{\frac{\partial \Jc_i^{(\ell)}}{\partial \bal}} \leq D\mult\times \sqrt{D}\mult\sqrt{h}\leq D^{3/2}(\mult)^2\sqrt{h}.\label{j bound 3}
\end{align}

\subsection{Finalizing the Proof of Lemma \ref{lemma deep act}} 
We now put the bounds above together. Overall, combining \eqref{kernel diff}, \eqref{j diff bound}, \eqref{j bound 3}, we find that
\[
\frac{\|\Kbh_\bal-\Kbh_{\bab}\|}{\tn{\bal-\bab}}\leq 2\sqrt{2Dn}\mult\times \sqrt{2Dn}\times D^{3/2}(\mult)^2\sqrt{h}\leq  4n\sqrt{h}(D\mult)^3.
\]
This is summarized in the following lemma.
\begin{lemma} [Gram matrix Lipschitzness] \label{lem deter lip bound}Suppose $i$th layer has $k_i$ neurons with $k_0=d$ and $\W^{(i)}\in\R^{k_i\times k_{i-1}}$. Suppose $\phi_i$ obeys $|\phi_i(0)|,|\phi_i'(x)|,|\phi_i''(x)|\leq B$. Suppose input features obey $\tn{\x}\leq N$ for some $N\geq \sqrt{d}$. Also let $S_i$ be an upper bound on $\|\W_i\|$ and define $\bar{M}=\prod_{i=1}^{D+1}(S_i+\clr{\max(1,\sqrt{k_i/k_{i-1}})})$. Suppose $\bal=[\bl{1}~\dots~\bl{D}]$ where $\bl{i}$ governs $i$th layer activation and $\tone{\bl{i}}\leq 1$. We have that
\begin{align}
\|\Kbh_\bal-\Kbh_{\bab}\|\leq 4n\sqrt{h}(D\mult)^3\tn{\bal-\bab}.\label{rem terms}
\end{align}
\end{lemma}
The following lemma is a restatement of Lemma \ref{lemma deep act} (i.e.~its more precise version) and essentially follows from the deterministic bound of Lemma \ref{lem deter lip bound}.
\begin{lemma} \label{lem rand lip bound} Suppose $i$th layer has $k_i$ neurons with $k_0=d$ and $\W^{(i)}\in\R^{k_i\times k_{i-1}}$. Suppose $\phi_i$ obeys $|\phi_i(0)|,|\phi_i'(x)|,|\phi_i''(x)|\leq B$. Suppose input features are normalized to ensure $\tn{\x}\lesssim \sqrt{d}$. Fix constants $\rho\geq 1$, $\cc>0$. Suppose the aspect ratios obey $k_i/k_{i-1}+1\leq \rho$ for $2\leq i\leq D$. Suppose layer $i$ is initialized as $\Nn(0,c_i)$ for $1\leq i\leq D+1$. Additionally suppose
\[
c_i\leq\begin{cases}\cc\quad\text{if}\quad i=1\\\cc/k_{i-1}\quad\text{if}\quad i\geq 2\end{cases}.
\] 
Set $k_{\min}=\min_{1\leq i\leq D}k_i$. Then, there exists a constant $C>0$ such that, with probability $1-De^{-10k_{\min}}$, for all $\bal,\bab\in\Bal$
\[
\frac{\|\Kbh_\bal-\Kbh_{\bab}\|}{\tn{\bal-\bab}}\leq (CB\sqrt{\cc\rho})^{3D} n\sqrt{h}D^3(k_1/d+1)^{3/2}d^3.
\]
Similarly $\Kb_\bal=\E[\Kbh_\bal]$ is also $(CB\sqrt{\cc\rho})^{3D} n\sqrt{h}D^3(k_1/d+1)^{3/2}d^3$-Lipschitz function of $\bal$.
\end{lemma}
\begin{proof} We just need to plug in the proper quantities to Lemma \ref{lem deter lip bound}. Let $C>0$ be an absolute constant to be determined. Let $\rho_i$ be the $i$th layer aspect ratio $k_i/k_{i-1}+1$. Let $\cc_i=c_ik_{i-1}$. Observe that, using Gaussian tail bound and Lemma \ref{moment gauss}, for all $D+1\geq i\geq 1$, 
\begin{align}
&\Pro(\sqrt{\frac{k_{i-1}}{\cc_i}}\|\W^{(i)}\|\geq {\sqrt{k_i}+\sqrt{k_i-1}+t})\leq e^{-t^2/2}\label{tail b}\\
&\E[\|\W^{(i)}\|^3]\leq 16\cc_i^{3/2} (\frac{\sqrt{k_i}+\sqrt{k_i-1}+5}{\sqrt{k_{i-1}}})^3\leq (C^2\cc_i\rho_i)^{3/2}.
\end{align}
Define $\Gamma_i=C\sqrt{\cc_i\rho_i}$. This also implies that, with probability at least $1-(D+1)e^{-10k_{\min}}$ (union bound over all layers),
\begin{align}
\|\W^{(i)}\|\leq \Gamma_i,\quad \E[\|\W^{(i)}\|^3]^{1/3}\leq \Gamma_i.\nn
\end{align}
Since $\bar{S}_i\leq \|\W^{(i)}\|+\sqrt{\rho_i}$, this also implies (after adjusting the constant $C$)
\begin{align}
\bar{S}_i\leq \Gamma_i,\quad \E[\bar{S}_i^3]^{1/3}\leq \Gamma_i.\nn
\end{align}
Finally, we simply need to calculate $\bar{M}$. Noticing $\cc_i\rho_i\leq \cc\rho$ for $i\geq 2$ and $\cc_1\rho_1\leq \cc d(k_1+d)/d=\cc (k_1+d)$, we find
\begin{align}
&\bar{M}=\prod_{\ell=1}^{D+1}\bar{S}_i\leq \prod_{\ell=1}^{D+1}\Gamma_i\leq  (C^2\sqrt{\cc\rho})^{D}\sqrt{k_1+d}\\
&\E[\bar{M}^3]\leq (C\sqrt{\cc\rho})^{3D}(k_1+d)^{3/2}.
\end{align}
We conclude with the result after multiplying the remaining terms $n\sqrt{h}D^3B^{3D}N^3$ from \eqref{rem terms} with $N=\sqrt{d}$.
\end{proof}

\section{Properties of the Spectral Estimator}\label{spectral thm}

The following establishes an asymptotic guarantee for spectral estimator when learning an overparameterized rank-1 matrix. Unlike Theorem \ref{thm low-rank}, this result allows for label noise.
\begin{theorem}[Guarantees for the spectral estimator]\label{spec thm 2} Let $(\X_i)_{i=1}^n\subset\R^{\h\times p}$ be i.i.d.~matrices with i.i.d.~$\Nn(0,1)$ entries. Let $y_i=\bal^T\X_i\bt+\sigma z_i$ for a unit norm $\bal\in\R^\h,\bt\in\R^p$ and suppose the noise is $(z_i)_{i=1}^n\distas\Nn(0,1)$. Form the cross-moment matrix
\[
\hat{\M}=\frac{1}{n}\sum_{i=1}^n y_i\X_i.
\]
Let $\bah$ be the top left singular vector of $\hat{\M}$. Let $\bp=p/n$ and $\bh=h/n$. Let $1\geq \rho\geq 0$ be the absolute correlation between $\bah,\bal$ i.e. $\rho=|\bal^T\bah|$. In the large dimensional limit $p,n,h\rightarrow \infty$ (while keeping $\bp,\bh$ constant in the limit), with probability $1$, we have
\begin{align}
&\frac{\rho}{1-\rho^2}\geq \frac{(1+\sigma^2)^{-1}/\sqrt{\bh}-(2\sqrt{\bp}+\sqrt{\bh})}{2(\sqrt{\bp}+1)}.\label{spec bound main}
\end{align}
Specifically, assuming $(1+\sigma^2)\sqrt{\bp\bh}\leq 1/6$, we find $\rho^2\geq 1-64(1+\sigma^2)^2\bp\bh$.
\end{theorem}
\begin{proof} When an inequality (e.g.~$\leq$) holds in the large dimensional limit, with probability 1, we use the P-overset notation (e.g.~$\pleq$). During the proof, $\text{vec}(\cdot)$ denotes the vectorization of a matrix obtained by putting all columns on top of each other. $\text{mtx}(\cdot)$ denotes the inverse operation that constructs a matrix from a vector.

Let $\x_i=\text{vec}(\X_i)$. Let $g_i=\bt^T\x_i$ and $h_i\sim\Nn(0,1)$ independent of others. Decompose $\x_i=\x'_i+(g_i-h_i)\bt$ where $\x'_i=(\Iden-\bt\bt^T)\x_i+h_i\bt\distas\Nn(0,1)$ is independent of $g_i$.
\begin{align}
\text{vec}(\hat{\M})&=\frac{1}{n}\sum_{i=1}^n y_i\x_i=\frac{1}{n}\sum_{i=1}^n \x_i(\x_i^T\bt+\sigma z_i)\\
&=\frac{1}{n}[\sum_{i=1}^n \x'_i(\x_i^T\bt+\sigma z_i)+g_i^2\bt-g_ih_i\bt+\sigma (g_i-h_i)z_i\bt]\\
&= \bt+\rest_1+\frac{\bgam}{\sqrt{n}}\quad\text{where}\quad\te{\tn{\rest_1}}\lesssim \frac{1+\sigma}{\sqrt{n}}.
\end{align}
where $\bgam\sim\Nn(0,\zig)$ where $\zig=1+\sigma^2$ and $\rest_1$ is allowed to have an adversarial direction. Concretely, $\bgam$ (approximately) equal to the $\frac{1}{\sqrt{n}}\sum_{i=1}^n \x'_i(\x_i^T\bt+\sigma z_i)$ term.

Following this, we can rewrite as
\[
\text{vec}(\M)=\frac{1}{n}\sum_{i=1}^n y_i\x_i=\text{vec}(\Mb)+\rest_1\pb,
\]
where $\Mb,\Gb\in\R^{h\times p}$, $\Gb\distas\Nn(0,\zig)=\text{mtx}(\bgam)$ and
\[
\Mb=\bal\bt^T+\frac{\Gb}{\sqrt{n}}.
\]
Let $\g=\Gb\bt\distas\Nn(0,\zig)$. Consider the covariance matrix
\[
\Mb\Mb^T=\bal\bal^T+\frac{1}{n}\Gb\Gb^T+\frac{1}{\sqrt{n}}(\bal\g^T+\g\bal^T).
\]
Observe that
\[
\tn{\bal^T\Mb}^2=1+\frac{\tn{\Gb^T\bal}^2}{n}+\frac{2\g^T\bal}{\sqrt{n}}.
\]
Set $\bp=p/n$ and $\bh=h/n$. Set $\gamma=\sqrt{\bp\bh}$. This implies that
\[
\tn{\bal^T\Mb}^2=1+\frac{\zig p}{n}+\rest_2=1+\zig\bp+\rest_2\quad\text{where}\quad \te{\rest_2}\lesssim \frac{\sqrt{\zig}\sqrt{n+p}}{n}=\frac{\sqrt{\zig}\sqrt{1+\bp}}{\sqrt{n}}.
\]
Thus in the high-dimensional limit ($n,p$ sufficiently large), contributions of $\rest_1,\rest_2$ disappears and
\begin{align}
\tn{\bal^T\Mb}^2\rP 1+\zig\bp.\label{bal bound}
\end{align}
Now, let $\eb=\text{top\_eigvec}(\Gb\Gb^T)$ and $\gb=\g/\tn{\g}$. Observe that $\eb$ is generated uniformly randomly over the range space of $\Gb$. Thus, the inner products $\gb^T\eb,\gb^T\bal,\eb^T\bal$ all have subgaussian norm at most $1/\sqrt{n}$ and become orthogonal in the high-dimensional limit. Also note that since $\eb$ is an eigenvector, $\Gb\Gb^T\eb$ is again asymptotically orthogonal to $\bal,\gb$ (as it is perfectly parallel to $\eb$ which is orthogonal to $\bal,\gb$). Top eigenvalue obeys the Bai-Yin law, 
\begin{align}
\frac{\eb^T\Gb\Gb^T\eb}{n}\rP \zig\frac{(\sqrt{p}+\sqrt{\h})^2}{n}= \zig(\sqrt{\bp}+\sqrt{\bh})^2
\end{align}
Since $\bal$ is fixed (and independent of $\Gb$), we have that
\[
\bal^T\frac{\Gb\Gb^T}{n}=\zig\bp\bal+\zig\sqrt{\bp\bh}\hb+\rest_3\quad\text{where}\quad \te{\tn{\rest_3}}\lesssim \zig\frac{\sqrt{\bp}}{\sqrt{n}},
\]
where $\hb$ term is distributed as $\distas\Nn(0,1/n)$. Let $\ab$ be the top eigenvector of $\Mb\Mb^T$ and $\ab=a\bal+b\vb$ where $\bal,\vb$ are two orthogonal unit vectors and $a^2+b^2=1$. Note that $\hb$ is uniformly generated in the range space of $\Gb^T$. Thus $\tn{\Gb\hb}^2/n\rP \zig(\bp+\bh)$. It is also orthogonal to $\eb,\ab,\gb$ in the limit.  Consequently, (in the limit) we have 
\begin{align}
(a\bal+b\vb)^T\frac{\Gb\Gb^T}{n}(a\bal+b\vb)&\pleq \zig[\bp a^2+2ab\sqrt{\bp\bh} |\hb^T\vb|+b^2(\sqrt{\bp}+\sqrt{\bh})^2]\\
&\pleq \zig[\bp+2|ab|\gamma+2b^2\gamma+b^2\bh]
\end{align}
We additionally have the bound
\[
|\frac{2\ab^T\g\bal^T \ab}{\sqrt{n}}|\pleq 2\sqrt{\zig} |ab|\sqrt{\bh}.
\]
Combining, in large dimensional limit, we find that
\begin{align}
\tn{\ab^T\Mb}^2\pleq a^2+\zig\bp+2\zig|ab|(\gamma+\sqrt{\bh})+\nu b^2(2\gamma+\bh).\label{fin bound}
\end{align}
Since $\ab$ is the top eigenvector, using bounds \eqref{bal bound} and \eqref{fin bound}, we obtain via $\tn{\ab^T\Mb}^2\geq \tn{\bal^T\Mb}^2$ that
\begin{align}
&a^2+\zig\bp+2\zig|ab|(\gamma+\sqrt{\bh})+\nu b^2(2\gamma+\bh)\geq 1+\zig\bp\\
\implies&\zig\bp+2\zig|ab|(\gamma+\sqrt{\bh})+\nu b^2(2\gamma+\bh)\geq b^2\\
\implies&2\zig|ab|(\gamma+\sqrt{\bh})\geq (1-\nu(2\gamma+\bh))b^2\\
\implies&2\zig|a|(\gamma+\sqrt{\bh})\geq (1-\nu(2\gamma+\bh))|b|\\
\implies &\frac{|a|}{|b|}\geq \frac{1-\nu(2\gamma+\bh)}{2\zig(\gamma+\sqrt{\bh})}.
\end{align}
After simplification, the line above is identical to \eqref{spec bound main}. To proceed, using the fact that $\bp\geq \gamma\geq \bh$ and assuming $\nu\gamma\leq 1/6$ we find
\[
\frac{|a|}{|b|}\geq \frac{1}{8\nu\gamma}.
\]
This also yields $|a|^2\geq 1/(1+64\nu^2\gamma^2)\geq 1-64\nu^2\gamma^2$.
\end{proof}
\section{Finalizing the Proof of Theorem \ref{thm low-rank}}\label{thm low-rank sec}

The proof is completed in two stages. First, we prove \eqref{high corr}. The statement is directly implied by Theorem \ref{spec thm 2} by setting the noise level to zero. This gives us $\bah$ which is $\rho$ (absolute) correlated with $\bas$ where $\rho$ obeys \eqref{high corr}. For the second statement, we assume that validation training is complete, \eqref{high corr} and focus on the empirical risk minimization over training data to bound the risk of overparameterized learning with $\bts$.  Below, without losing generality, we assume the correlation between $\bah$ and $\bas$ are nonnegative (as the situation is symmetric).

Observe that when $\bah$ is fixed (i.e.~conditioned on the outcome of the spectral estimator), we define the feature matrix for the ERM \eqref{high dim linear}. Specifically, define the matrix $\bPhi_\bah\in\R^{n\times p}$ where $i$th row of $\bPhi_\bal$ is given by $\xh_i=\X_i^T\bah$. Also define the ideal feature matrix $\bPhi_\bas$ where the $i$th row is given by $\x_i=\X_i^T\bas$. Observe that $\xh_i$ are essentially noisy features that contain a mixture of the right features (i.e.~$\x_i$)  as well as the wrong features that are induced by the component of $\bah$ orthogonal to $\bas$. 

Specifically, decompose $\bah=\rho\bas+\sqrt{1-\rho^2}\bar{\bal}$ where $\bar{\bal}$ is also unit norm. Then set $\z_i=\X_i^T\bar{\bal}$ and observe that
\[
\xh_i=\rho\x_i+\sqrt{1-\rho^2}\x_i.
\]

The outcome of Theorem \ref{spec thm 2} upper bounds the magnitude of these wrong features (by lower bounding $\rho$). To proceed, we shall establish the exact asymptotic risk when fitting these noisy features. Given this discussion, the proof of the risk bound \eqref{high dim linear} is essentially established via the following lemma. The first statement applies for any value of $\rho$ and the second statement chooses $\rho$ induced by the spectral estimator to conclude with the proof of \eqref{high dim linear}.
\begin{lemma}[Asymptotic risk of regression with suboptimal feature map] Fix $\bts\in\R^p$ and $(\x_i,\z_i)_{i=1}^n\distas\Nn(0,\Iden_p)$. Set $y_i=\x_i^T\bts$ and define the noisy features $\xh_i=\rho\x_i+\sqrt{1-\rho^2} \z_i$. Solve the problem (which is same as \eqref{erm prob})
\[
\bth=\arg\min_{\bt} \Lc(\bt)\quad\text{where}\quad \Lc(\bt)=\tn{\y-\bPhi_\bah\bt}^2.
\]
Consider the double asymptotic regime with $p,n\rightarrow\infty$ and $p/n\rightarrow\bp>1$. We have that
\[
\lim_{n\rightarrow\infty}\Lc(\bth)=\E[(y-\x^T\bth)^2]=\frac{\bp^2-2\bp\rho+2\rho-\rho^2}{\bp(\bp-1)}.
\]
Specifically, assume $\bp\bh\leq c\leq 1/6$ for sufficiently small constant $c>0$. Recalling $\rho^2\geq 1-64\bh\bp$ as given by Theorem \ref{spec thm 2} (where we set $\sigma=0$), we find
\begin{align}
\lim_{n\rightarrow\infty}\Lc(\bth)\leq 1-\frac{1}{\bp}+\frac{200 \bh}{1-1/\bp}.\label{high dim linear2}
\end{align}
\end{lemma}
\begin{proof} We remark that related results/analysis exist in the literature (in the context of overparameterized high-dimensional learning and the properties of the min-norm interpolating solutions) \cite{hastie2019surprises}. Our strategy uses the results from \cite{chang2020provable}.

Define the vector $\ab=\sqrt{1-\rho^2} \x-\rho\z$ and note that $\xh,\ab$ are independent. Additionally, note that
\[
y=\x^T\bts=\rho\xh^T\bts+\sqrt{1-\rho^2} \ab^T\bts.
\]
Set $w=\ab^T\bts\sim\Nn(0,1)$ which corresponds to the noise level of the problem. The original data in terms of the noisy features can be written as follows
\begin{align}
y_i&=\rho\xh_i^T\bts+\sqrt{1-\rho^2} w_i.
\end{align}
Define the asymptotic risk $\text{risk}(\rho,\bp)=\lim_{n\rightarrow\infty}\E[(y-\x^T\bth)^2]=\E[\tn{\bth-\bts}^2]$. Let $\hh\sim \Nn(0,\Iden_p/p)$. Let us introduce the random vector 
\[
\btb\sim \frac{\rho}{\bp}\bts+\sqrt{\frac{1-\rho^2}{\bp-1}+\frac{(\bp-1)\rho^2}{\bp^2}}\hh=\frac{\rho}{\bp}\bts+\sqrt{\frac{\bp^2-\rho^2\bp^2+\bp^2\rho^2-2\bp\rho^2+\rho^2}{\bp^2(\bp-1)}}\hh=\frac{\rho}{\bp}\bts+\sqrt{\frac{\bp^2-2\bp\rho^2+\rho^2}{\bp^2(\bp-1)}}\hh.
\] Specializing the results of \cite{chang2020provable} to identity covariance shows that, in the double asymptotic overdetermined regime ($\bp=p/n>1$),
\begin{align}
\text{risk}(\rho,\bp)&=\lim_{n\rightarrow\infty}\E[\tn{\btb-\bts}^2]=(1-\frac{\rho}{\bp})^2+\frac{\bp^2-2\bp\rho^2+\rho^2}{\bp^2(\bp-1)}\\
&=\frac{\bp^3-2\bp^2\rho+\rho^2\bp-\bp^2+2\bp\rho-\rho^2}{\bp^2(\bp-1)}+\frac{\bp^2-2\bp\rho^2+\rho^2}{\bp^2(\bp-1)}\\
&=\frac{\bp^2-2\bp\rho+\rho^2-\bp+2\rho}{\bp(\bp-1)}+\frac{\bp-2\rho^2}{\bp(\bp-1)}\\
&=\frac{\bp^2-2\bp\rho+2\rho-\rho^2}{\bp(\bp-1)}.
\end{align}
This proves the first statement. Observe that in the special case of $\rho=1$, the risk reduces to $\text{risk}(\bas):=\text{risk}(\rho,\bp)=\frac{\bp^2-2\bp+1}{\bp(\bp-1)}=\frac{\bp-1}{\bp}=1-1/\bp$.

To bound the risk on $\bah$, we study the derivative at $\rho=1$. Note that
\[
\frac{\partial \text{risk}(\rho,\bp)}{\partial \rho}=\frac{2-2\rho-2\bp}{\bp(\bp-1)}\implies \frac{\partial \text{risk}(\rho,\bp)}{\partial \rho}\Big|_{\rho=1}=\frac{-2\bp}{\bp(\bp-1)}=-\frac{2}{\bp-1}.
\]
This implies that if $1-\eps\leq \rho\leq 1$ for sufficiently small $\eps>0$, we have that
\[
\text{risk}(\rho,\bp)\geq \text{risk}(1,\bp)+\frac{3(1-\rho)}{\bp-1}
\]
Since $\bh\bp\leq c$ by choosing $c$ sufficiently small, we can ensure $\rho\geq \rho^2\geq 1-64 \bp\bh\geq 1-\eps$. Plugging in $\rho$ lower bound above yields
\[
\text{risk}(\rho,\bp)\geq \text{risk}(1,\bp)+\frac{3(1-64 \bp\bh)}{\bp-1}\leq \text{risk}(1,\bp)+\frac{200 \bh}{1-1/\bp}
\]
concluding the overall proof of \eqref{high dim linear2} which is a restatement of \eqref{high dim linear}.
\end{proof}

\section{Proofs for Shallow Neural Networks}\label{shallow proof}
We consider the NAS algorithm of Section \ref{sec:NAS} where the solution to the lower-level problem is obtained via gradient-based. First define the Jacobian of the network and NTK kernel at the random initialization. Given training dataset $\Tc$, define the Jacobian of the network 
\[
\Jb_\bal(\W)=[\frac{\fF{\bal}(\x_1)}{\pa \W}~\frac{\fF{\bal}(\x_2)}{\pa \W}~\dots~~\frac{\fF{\bal}(\x_{\nt})}{\pa \W}]^T\in \R^{\nt\times p}.
\]
The Neural Tangent Kernel with activation $\sigma_\bal$ has the following kernel matrix
\[
\Kb_\bal=\E_{\W_0\distas\Nn(0,1)}[\Jc_\bal(\W_0)\Jc^T_\bal(\W_0)]
\] 

We first introduce some short-hand notation. Set $\bt=\text{vec}(\W-\W_0)$. When $\bal$ is clear from context, given weights $\W$ define the network via $\fnn^\bt(\x)=\vb^T\sigma_\bal(\W\x)$ and linearized network as
\[
\flin^\bt(\x)=\vb^T \left[\sigma_\bal'(\W_0\x)\odot (\W-\W_0)\x\right].
\]
Based on this, introduce the initial prediction vector 
\[
\pb:=\pb_\bal=[\fnn(\x_1,\W_0)~\fnn(\x_2,\W_0)~\dots~~\fnn(\x_n,\W_0)].
\]
We then define the linearized problem
\begin{align}
\Lch^{\text{lin}}_\Tc(\W)=\frac{1}{2}\tn{\yT-\pb-\Jb_\bal(\W_0)\bt}^2.\label{linearized prob}
\end{align}
For the theorem below, we denote $\bt_t=\vc(\W_t-\W_0)$. We also denote $\btt_t=\vc(\Wt_t-\W_0)$ where $\Wt_t$ is the linearized iterations which are obtained by training on the linearized problem $\Lch^{\text{lin}}_\Tc$. 
\begin{theorem}[Shallow NAS Master Theorem]\label{one layer nas} Suppose input features and labels are normalized to $\tn{\x}\leq 1,|y|\leq 1$. Fix $\vb$ with half $\sqrt{c_0/k}$ and half $-\sqrt{c_0/k}$ entries for sufficiently small $c_0>0$. Initialize $\W_0\distas\Nn(0,1)$. Let $B>0$ upper bound $|\sigma'_\bal|,|\sigma''_\bal|$. Suppose the loss $\ell$ is bounded by a constant and $1$-Lipschitz and NTK lower bound Assumption \ref{ntk assump} holds and set the normalized lower bound $\blaz=\laz/\cz$. Suppose the network width obeys
\[
k\gtrsim \clr{k_0:=k_0(\eps,\blaz,\nt)}
\]
Additionally suppose $\nv\gtrsim \ordet{\h}$ where $\ordet{\cdot}$ hides the log terms. Suppose the following holds with probability $\clr{1-p_0}$ (over the initialization $\W_0$) uniformly over all $\bal\in\Bal$ and for all $\clr{T\geq T_0:=T_0(\eps,\blaz,\nt)}$
\begin{enumerate}
\item $\E_{\x\sim\Dc}[|\vb^T\sigma_\bal(\W_0\x)|], \frac{1}{\nv}\sum_{i=1}^{\nv}|\vb^T\sigma_\bal(\W_0\xt_i)|\leq \eps_0$.
\item $T$'th iterate $\bt_T$ obeys $\tf{\W_T-\Wt_\infty}=\tn{\bt_T-\btt_\infty}\leq \eps_1$
\item Rows are bounded via $\trow{\W_T-\W_0}\leq \sqrt{C_0/k}$.
\item At initialization, the network prediction is at most $\eps_2$ i.e. $\tn{\pb_\bal}\leq \eps_2$.
\item Initial Jacobians obey $\frac{\Jb_\bal\Jb_\bal^T}{\cz}\succeq \blaz\Iden_{\nt}/2$.
\item Initial Jacobians obey $\|(\Jb_\bal\Jb_\bal^T)^{-1}-\Kb_\bal^{-1}\|\leq \eps_3$. \clr{(Via Lemma \ref{inverse diff}, this is implied by $\|\Jb_\bal\Jb_\bal^T-\Kb_\bal\|\leq \cz^2\blaz^2\eps_3/2$.)}
\end{enumerate}
Fix $M=\lipnn$. Then, with probability $\clr{1-4e^{-t}-p_0}$, $\delta$-approximate NAS output obeys
\begin{align}
\Lc(\ft_\bah)\leq  \min_{\bal\in\Bal}2B\sqrt{\frac{c_0\y^T\Kb_\bal^{-1}\y}{\nt}}+C\sqrt{\frac{\h\log(M)+t}{\nv}}\clr{+\eps+\delta},\label{nas bound messy}
\end{align}
where $\clr{\eps=3(\eps_0+\sqrt{c_0}BC_0/{\sqrt{k}}+\sqrt{c_0}B\eps_1+2B\eps_2/\sqrt{\blaz})+B\sqrt{\cz\eps_3}}$. Additionally, since hinge loss dominates the 0-1 loss (standard classification error), the bound above also applied for the 0-1 loss $\Lcz$.
\end{theorem}
\begin{proof} For the proof, we would like to employ Theorem \ref{e2e bound}. To this aim, we introduce the so-called ideal feature map regression problem. Unlike \eqref{linearized prob}, ideal problem uses the exact labels $\y$ and solves
\begin{align}
\Lch^{\text{ideal}}_\Tc(\W)=\frac{1}{2}\tn{\yT-\Jb_\bal(\W_0)\bt}^2.\label{linearized prob ideal}
\end{align}
We define the ideal model to be the pseudo-inverse 
\begin{align}
\btid=\Jb_\bal^\dagger\yT.\label{bt ideal}
\end{align}
Note that the ideal problem is equivalent to the feature map regression task described in Definition \ref{fmap def} where feature maps are $\frac{\pa \fnn(\x)}{\pa \W_0}$. Thus, we can also study the generalization risk of $\btid$ on the new examples given by $\Lc(\flin^{\btid})$ where
\begin{align}
\Lc(\flin^{\bt})=\E_\Dc[\ell(y,\bt^T\frac{\pa \fnn(\x)}{\pa \W_0})]\label{linearized risk}.
\end{align}

To proceed with the proof, set $\eps'=\eps_0+BC_0\sqrt{c_0/k}+\sqrt{c_0}B\eps_1+2\sqrt{c_0}B\eps_2/\sqrt{\la_0}$. Let $\bah$ be a $\del$-approximate solution of the NAS problem. We first apply a triangle inequality on Lemmas \ref{lem nn1}, \ref{lem nn2}. Specifically, with probability $1-p_0$ over $\W_0$, for all $\bal\in\Bal$, Lemmas \ref{lem nn1}, \ref{lem nn2} hold. Fix $\Lcg\in\{\Lc,\Lch_\Vc\}$ (i.e.~either validation or population loss). Thus recalling $\ft_\bal= \fF{\bal}^{\bt_T}$, we write
\begin{align}
&|\Lcg(\fnn^{\bt_T})-\Lcg(\flin^{\btt_\infty})|\leq\eps_0+\frac{\sqrt{c_0}BC_0}{\sqrt{k}}+\sqrt{c_0}B\eps_1\nn\\
&|\Lcg(\flin^{\btt_\infty})-\Lcg(\flin^{\btid})|\leq 2\sqrt{c_0}B\eps_2/\sqrt{\la_0}\nn\\
\implies& |\Lcg(\ft_\bal)-\Lcg(\flin^{\btid})|\leq \eps'.\label{lg eq}
\end{align}
Thus any $\del$-approximate solution $\bah$ of the NAS problem ensures that
\[
\Lch_\Vc(f^{\btid_\bah})\leq \Lch_\Vc(\ft_{\bah})+\eps'\leq \min_{\bal\in\Bal}\Lch_\Vc(\ft_{\bal})+\eps'+\del\leq \inf_{\bal\in\Bal}\Lch_\Vc(f^{\btid_\bal})+2\eps'+\del.
\]
Thus, $f^{\btid_\bah}$ is a $(2\eps'+\del)$-approximate solution of the linearized feature map regression. To proceed, Lemma \ref{lem nn3} establishes the generalization guarantee for such a $f^{\btid_\bah}$ via
\[
\Lc(f^{\btid_\bah})\leq  \min_{\bal\in\Bal}2\sqrt{c_0}B\sqrt{\frac{\y^T(\Jb_\bal\Jb_\bal^T)^{-1}\y}{\nt}}+C\sqrt{\frac{\h\log(M)+\tau}{\nv}}+2\eps'+\del.\label{ideal nn bound2}
\]
Finally, we go back to neural net's generalization via setting $\Lcg=\Lc$ in \eqref{lg eq} which gives $|\Lc(\ft_\bah)-\Lc(\flin^{\btid_{\bah}})|\leq \eps'$ where we note that $\ft_\bah=\fF{\bah}^{\bt_T}$. To conclude also plug in \eqref{k diff} to move to $\Kb_\bal$. These as a whole imply Theorem \ref{one layer nas}'s statement \eqref{nas bound messy} after (1) applying the change of variable $3\eps'+\sqrt{c_0}B\sqrt{\eps_3}\leftrightarrow \eps$ and then applying the change of variable $\blaz=\cz\laz$.
\end{proof}
\begin{lemma} \label{lem nn1}Consider the setup of Theorem \ref{one layer nas}, specifically the itemized assumptions involving the initialization $\W_0$ which holds with probability $\clr{1-p_0}$. Let $\bt_T,\btt_T$ be the iterations induced by any fixed activation $\bal\in\Bal$. For $\Lcg\in\{\Lc,\Lch_\Vc\}$ (i.e. for population or validation risk), we have that
\[
|\Lcg(\fnn^{\bt_T})-\Lcg(\flin^{\btt_\infty})|\leq \eps_0+\frac{\sqrt{c_0}BC_0}{\sqrt{k}}+\sqrt{c_0}B\eps_1.
\]
\end{lemma}
\begin{proof} Applying Lemma \ref{lemma simple nn}, we find that
\[
|\Lcg(\fnn^{\bt_T})-\Lcg(\flin^{\bt_T})|\leq \eps_0+\frac{\sqrt{c_0}C_0B}{\sqrt{k}}.
\]
Next, observe that for any input with $\tn{\x}\leq 1$, the neural feature maps are bounded by
\[
\tn{\frac{\pa \fnn(\x)}{\pa \W_0}}\leq \sqrt{c_0}B.
\]
This means that
\[
|\Lcg(\flin^{\bt_T})-\Lcg(\flin^{\btt_\infty})|\leq \sqrt{c_0}B\tn{\btt_\infty-\bt_T}\leq \sqrt{c_0}B\eps_1.
\]
To conclude use a triangle inequality to combine the bounds above.
\end{proof}
\begin{lemma}[Bounding the perturbation of the linearized model]\label{lem nn2} Consider the setup of Theorem \ref{one layer nas}, specifically the itemized assumptions involving the initialization $\W_0$ which holds with probability $\clr{1-p_0}$. Recall $\btid$ from \eqref{bt ideal} and that $\pb$ is the prediction vector on $\Tc$ at $\W_0$ bounded as $\tn{\pb}\leq \eps_2$.
For $\Lcg\in\{\Lc,\Lch_\Vc\}$, we have that
\[
\Lcg(\flin^{\btt_\infty})\leq \Lcg(\flin^{\btid})+2\sqrt{c_0}B\eps_2/\sqrt{\la_0}.
\]
Additionally, the perturbation due to empirical vs population Jacobian is bounded via
\begin{align}
\sqrt{\y^T(\Jb\Jb^T)^{-1}\y}\leq \sqrt{\eps_3}\tn{\y}+\sqrt{\y^T\Kb^{-1}\y}.\label{k diff}
\end{align}
\end{lemma}
\begin{proof} Recall that feature map norm is bounded by $\sqrt{c_0}B$ and thus 
\[
\Lcg(\flin^{\btt_\infty})\leq \Lcg(\flin^{\btid})+\sqrt{c_0}B\tn{\btid-\btt_\infty}.
\]
We upper bound the right hand side via
\[
\tn{\btid-\btt_\infty}\leq \tn{\Jb_\bal^\dagger\y-\Jb_\bal^\dagger(\y-\pb)}\leq 2\tn{\pb}/\sqrt{\la_0}\leq 2\eps_2/\sqrt{\la_0}.
\]
For the next result let $\Pb=\Kb^{-1}-(\Jb\Jb^T)^{-1}$. Using $\|\Pb\|\leq \eps_3$, we have that 
\begin{align}
\sqrt{\yT^T(\Jb\Jb^T)^{-1}\yT}\leq \sqrt{\yT^T(\Kb^{-1}-\Pb)\yT}\leq \sqrt{\yT^T\Kb^{-1}\yT}+\sqrt{\yT^T\Pb\yT}\leq  \sqrt{\yT^T\Kb^{-1}\yT}+\tn{\y}\sqrt{\|\Pb\|}.\nn
\end{align}
\end{proof}
The next result shows a uniform upper bound on the ideal solutions $\btid_\bal$ which solve the feature map regression with Jacobian matrix $\Jb_\bal$.
\begin{lemma} \label{lem nn3}Fix $M=\lipnn$. Let $\bah$ be a $\del$-approximate solution of \eqref{opt alpha} with linearized Jacobian feature map with labels $\y$. Set $\Kbh_\bal=\Jb_\bal\Jb_\bal^T$ and suppose $\Kbh_\bal\succeq \laz\Iden_{\nt}/2$ for all $\bal\in\Bal$ with $\sup_{\bal\in\Bal}\tone{\bal}\leq 1$. Recall the definition \eqref{linearized risk}. With probability at least $1-2\e^{-\tau}$, we have that
\begin{align}
\Lc^{\text{ideal}}(\ft_\bah)\leq  \min_{\bal\in\Bal}2\sqrt{c_0}B\sqrt{\frac{\y^T\Kbh_\bal^{-1}\y}{\nt}}+C\sqrt{\frac{\h\log(M)+\tau}{\nv}}+\del.\label{ideal nn bound}
\end{align}
where $\ft_\bah$ is the $\delta$-approximate solution of \eqref{opt alpha} with the feature map regression problem \eqref{linearized prob ideal}.
\end{lemma}
\begin{proof} We need to plug in the right quantities into Theorem \ref{e2e bound}. First note that we assumed $\tone{\bal}=R=1$. First observe that neural feature maps $\frac{\pa \fnn(\x)}{\pa \W}$ are bounded by $\sqrt{c_0}B$ in Euclidean norm thus we substitute $B\leftrightarrow c_0B^2$. Secondly, the Jacobian feature matrix $\Jb_\bal$ obeys \eqref{low bound} with $\laz/2$. Thus we also set $\laz\leftrightarrow\laz/2$, $\Gamma=1$ and $R=1$. Finally, apply the change of variable to normalized $\laz$ via $\blaz=\laz/\cz$. Thus we exactly find \eqref{ideal nn bound} for $M=\lipnn$.
\end{proof}

\begin{lemma}\label{lemma simple nn} Let $\W_0\in\R^{k\times d}$. Suppose $\sigma$ is a function with second derivative bounded by $B>0$ in absolute value. Let $c_0,C_0>0$ be scalars. Suppose $\W\in\R^{k\times d}$ is such that $\sup_{1\leq i\leq k} \tn{\w_i-\w_{0,i}}\leq \sqrt{C_0/k}$ and $\tin{\vb}\leq \sqrt{c_0/k}$. Define neural net  $f_{\text{nn}}(\x)=\vb^T\sigma(\W\x)$ and its linearization
\[
f_{\text{lin}}(\x)=\vb^T (\sigma'(\W_0\x)\odot (\W-\W_0)\x).
\]
Suppose input space $\Xc$ is subset of unit Euclidean ball and $\E_{\x\sim\Dc}[|\vb^T\sigma_\bal(\W_0\x)|], \frac{1}{\nv}\sum_{i=1}^{\nv}|\vb^T\sigma_\bal(\W_0\xt_i)|\leq \eps_0$. Let $\ell$ be a $\Gamma$-Lipschitz loss. Then for $\Lcg\in\{\Lc,\Lch_\Vc\}$
\[
|\Lcg(f_{\text{nn}})-\Lcg(f_{\text{lin}})|\leq \Gamma(\eps_0+\frac{\sqrt{c_0}C_0B}{\sqrt{k}}).
\]
\end{lemma}
\begin{proof} Let $\bar{f}_{\text{nn}}(\x)=f_{\text{nn}}(\x)-\vb^T\sigma(\W_0\x)$. Via Taylor series expansion, for any $\tn{\x}\leq 1$
\begin{align*}
|\bar{f}_{\text{nn}}(\x)-f_{\text{lin}}(\x)|&=\sum_{i=1}^k|\vb_i \sigma''(\w_{0,i}^T\x)((\w_i-\w_{0,i})^T\x)^2|\\
&=\sum_{i=1}^k\tin{\vb} B\tn{\w_i-\w_{0,i}}^2\tn{\x}^2\\
&\leq B\tn{\x}^2\sum_{i=1}^k \frac{\sqrt{c_0}}{\sqrt{k}} (\sqrt{\frac{C_0}{k}})^2\\
&\leq \frac{\sqrt{c_0}C_0\tn{\x}^2B}{\sqrt{k}}\leq \frac{\sqrt{c_0}C_0B}{\sqrt{k}}.
\end{align*}
Since loss function is $\Gamma$ Lipschitz, we obtain
\[
|\Lcg(\bar{f}_{\text{nn}})-\Lcg(f_{\text{lin}})|\leq \Gamma(\eps_0+\frac{\sqrt{c_0}C_0B}{\sqrt{k}}).
\]
We conclude via triangle inequality after using the condition of small $\Lcg$ prediction at $\W_0$ (which bounds $|f_{\text{nn}}-\bar{f}_{\text{nn}}|$).
\end{proof}

\section{Gradient Descent Analysis for Shallow Networks}
This section only focuses on the training dataset $\Tc$. Thus, to keep notation more concise, throughout we suppose $\Tc$ is a dataset with $n$ samples (i.e. we set $\nt\gets n$). Following Section \ref{sec:NAS}, starting at a random initialization $\W_0\distas\Nn(0,1)$, we optimize the training loss
\[
\Lch_{\Tc}(\W)=\frac{1}{2}\sum_{i=1}^{n}(y_i-\fF{\bal}(\x_i,\W))^2=\frac{1}{2}\tn{\y-f_\bal(\W)}^2,
\]
via gradient updates $\W_{\tau+1}=\W_\tau-\eta \nabla \Lch_\Tc(\W_\tau)$  for $T$ iterations. Here $\y$ is the concatenated label vector and $f_\bal(\W)$ is the prediction vector with entries $\fF{\bal}(\x_i,\W)$. We will drop the subscript $\bal$ as the $\bal$-dependence is clear from context. Consider the mixture of activation functions given by
\begin{align*}
	\sigma_{\bal}(z)=\sum_{r=1}^\h \bal_r \sigma_r(z)
\end{align*}
Throughout this section we assume $\bal\in\Bal$. Assume $\Bal$ is subset of the unit $\ell_1$ ball i.e.~all $\bal\in\Bal$ obeys $\tone{\bal}\leq 1$. Next we define the neural tangent kernel.
\begin{definition}[Neural tangent kernel and minimum eigenvalue]\label{nneig} Let $\vct{w}\in\R^d$ be a random vector with a $\mathcal{N}(\vct{0},\mtx{I}_d)$ distribution. Also consider a set of $n$ input data points $\vct{x}_1,\vct{x}_2,\ldots,\vct{x}_n\in\R^d$ aggregated into the rows of a data matrix $\X\in\R^{n\times d}$. Associated to a network $\vct{x}\mapsto\vct{v}^T\sigma_\bal\left(\W\x\right)$ and the input data matrix $\X$ we define the neural tangent kernel matrix as
	\[
	\Kb_\bal=\E_{\W_0\distas\Nn(0,1)}[\Jc_\bal(\W_0)\Jc^T_\bal(\W_0)]
	\] 
	We also define the eigenvalue $\lambda_\bal(\X)$ based on $\Kb_\bal(\X)$ as
	\begin{align*}
		\lambda_\bal(\X):=\lambda_{\min}\left(\Kb_\bal(\X)\right).
	\end{align*}
\end{definition}

\begin{assumption}\label{mineigass} We assume
	\begin{align*}
		\underset{\bal\in\Bal}{\min}\text{ }\lambda_\bal(\X)\ge \laz(\X)
	\end{align*}
	Additionally define the invariant initialization-scale lower bound $$\bar{\la}_0(\X) =  \frac{\la_0(\X)}{\twonorm{\vct{v}}^2}$$ and state the bounds in terms of this quantity.
\end{assumption}
\begin{theorem}\label{maincor}
	Consider a data set of input/label pairs $\vct{x}_i\in\R^d$ and $y_i\in\R$ for $i=1,2,\ldots,n$ aggregated as rows/entries of a data matrix $\mtx{X}\in\R^{n\times d}$ and a label vector $\vct{y}\in\R^n$. Without loss of generality we assume the dataset is normalized so that $\twonorm{\vct{x}_i}=1$. Also consider a one-hidden layer neural network with $k$ hidden units and one output of the form $\vct{x}\mapsto \vct{v}^T\sigma_\bal\left(\mtx{W}\vct{x}\right)$ with $\mtx{W}\in\R^{k\times d}$ and $\vct{v}\in\R^k$ the input-to-hidden and hidden-to-output weights. We assume the activations $\sigma_1, \sigma_2, \ldots, \sigma_\h$ with $h\le n$ has bounded derivatives i.e.~$\abs{\sigma_j'(z)}\le B$ and $\abs{\sigma_j''(z)}\le B$ for all $z$. Also let \clr{$\lazb(\X)$} denote the minimum eigenvalue of the neural net covariance per Assumption \ref{mineigass}. Furthermore, we fix $\vct{v}$ by setting half of the entries of $\vct{v}\in\R^k$ to $\frac{\czmli}{\sqrt{k}}$ and the other half to $-\frac{\czmli}{\sqrt{k}}$ with $\czmli$ obeying $ \czmli\le \frac{1}{4\sqrt{\log n}}$ and train only over $\mtx{W}$.  Starting from an initial weight matrix $\mtx{W}_0$ selected at random with i.i.d.~$\mathcal{N}(0,1)$ entries, we run Gradient Descent (GD) updates of the form $\mtx{W}_{\tau+1}=\mtx{W}_\tau-\eta\nabla \mathcal{L}(\mtx{W}_\tau)$ with step size $\eta\le \frac{1}{2\czsqr B^2\opnorm{\mtx{X}}^2}$. Then, as long as, for some $\gamma\le 1$ and and $C>0$ a fixed numerical constant, we have
	\begin{align}
		\label{kcond}
		k\ge C\frac{1}{\gamma^4\lazb^8(\mtx{X})} (\log n)  B^{16}\opnorm{\mtx{X}}^{16}h+C\frac{B^8n\|\X\|^8}{c_0\gamma^2\blaz^5},
	\end{align}
	then there is an event of probability at least $1-\frac{4}{n^3}-4e^{-10h}$ such that on this event, for all activation choices $\bal\in\Bal$, all GD iterates obey
	\begin{align}
		\tn{f(\mtx{W}_\tau)-\y}^2\leq& 4n\left(1-\eta\frac{\czsqr\lazb(\mtx{X})}{8}\right)^\tau,\label{ineqs conv}\\
		\fronorm{\mtx{W}_\tau-\mtx{W}_0}\leq& \frac{16\sqrt{n}}{\sqrt{\czsqr\lazb(\mtx{X})}},\nonumber\\
		\|\mtx{W}_\tau-\mtx{W}_0\|_{2,\infty}\le& \frac{  32B\|\X\|}{\sqrt{k}\czmli\lazb(\mtx{X})} \sqrt{n},\nonumber\\
		\fronorm{\mtx{W}_\tau-\widetilde{\W}_\infty}\le& \frac{5}{2}\frac{\gamma}{\czmli B\opnorm{\X}}\sqrt{n}+4\left(1-\frac{1}{4}\eta c_0\lazb(\X)\right)^t\frac{\sqrt{n}}{\sqrt{c_0\lazb(\X)}}		\label{ineqs}
	\end{align}
	Furthermore, on the same event, we also have:
	
	\noi$\bullet$ (a) for any two distributions $\Dc_1$ and $\Dc_2$ over the unit Euclidean ball of $\R^d$ 
	\begin{align}
		\label{ineqc3}
		\E_{\x\sim\Dc_{1}}[\vb^T\sigma_\bal\left(\W_0\x\right)]\le \czmli\left(1+3\sqrt{\log n}\right) B\quad\text{and}\quad \E_{\x\sim\Dc_{2}}[\vb^T\sigma_\bal\left(\W_0\x\right)]\le \czmli \left(1+3\sqrt{\log n}\right) B,
	\end{align}
	\noi$\bullet$ (b) and prediction at initialization obeys
	\begin{align}
		\label{ineqc1}
		\twonorm{\sigma_\bal\left(\X\W_0^T\right)\vct{v}}\le \czmli \sqrt{n}\left(1+3\sqrt{\log n}\right) B,
	\end{align}
	\noi$\bullet$ (c) and the following bound on the Jacobian matrix
	\begin{align}
		\label{ineqc2}
		\opnorm{\mathcal{J}_\bal(\W_0)\mathcal{J}_\bal^T(\W_0)-\Kb_\bal(\mtx{X})}\le \eps_0^2
	\end{align}
	holds for $\eps_0=\frac{\gamma }{512}\frac{\czmli\lazb^2(\mtx{X})}{ B^3\opnorm{\mtx{X}}^3}$.
\end{theorem}

\subsection{Proof of Theorem \ref{maincor}}
In order to prove this result we first need to state some auxiliary lemmas that characterize various properties of the Jacobian matrix. The first two concern the uniform concentration of the Jacobian matrix and uniform bound on the minimum eigenvalue at initialization and will be proven later on in this section. 
\begin{lemma}[Jacobian Concentration]\label{Jacconcen} Consider a one-hidden layer neural network model of the form $\vct{x}\mapsto \vct{v}^T\sigma_\bal\left(\W\x\right)$ where the activations $\sigma_1, \sigma_2, \ldots, \sigma_\h$ have bounded second derivatives obeying $\abs{\sigma_j''(z)}\le B$. Also assume we have $n$ data points $\vct{x}_1, \vct{x}_2,\ldots,\vct{x}_n\in\R^d$ with unit euclidean norm ($\twonorm{\vct{x}_i}=1$). Then, as long as
	\begin{align*}
		\frac{1}{\|\vct{v}\|_{\ell_4}^4}\ge \frac{C}{\eps_0^4}(\log n) B^4\opnorm{\mtx{X}}^4h,
	\end{align*}
	the Jacobian matrix at a random point $\mtx{W}_0\in\R^{k\times d}$ with i.i.d.~$\mathcal{N}(0,1)$ entries obeys 
	\begin{align*}
		\opnorm{\mathcal{J}_\bal(\W_0)\mathcal{J}_\bal^T(\W_0)-\Kb_\bal(\mtx{X})}\le \eps_0^2
	\end{align*}
	holds simultaneously for all $\bal\in\Bal$ with probability at least $1-4e^{-10h}$.
\end{lemma}
\begin{lemma}[Minimum eigenvalue of the Jacobian at initialization]\label{minspectJ} Consider a one-hidden layer neural network model of the form $\vct{x}\mapsto \vct{v}^T\sigma_\bal\left(\W\x\right)$ where the activations $\sigma_1, \sigma_2, \ldots, \sigma_\h$ have bounded derivatives obeying $\abs{\sigma_j'(z)}\le B$. Also assume we have $n$ data points $\vct{x}_1, \vct{x}_2,\ldots,\vct{x}_n\in\R^d$ with unit euclidean norm ($\twonorm{\vct{x}_i}=1$). Then, as long as
	\begin{align*}
		\frac{\twonorm{\vct{v}}}{\infnorm{\vct{v}}}\ge \sqrt{30h\log(nk)}\frac{\opnorm{\X}}{\sqrt{\lazb(\X)}}B,
	\end{align*}
	the Jacobian matrix at a random point $\mtx{W}_0\in\R^{k\times d}$ with i.i.d.~$\mathcal{N}(0,1)$ entries obeys 
	\begin{align*}
		\underset{\bal\in\Bal}{\min}\text{ }\sigma_{\min}\left(\mathcal{J}_\bal(\W_0)\right)\ge \frac{1}{2}\sqrt{\czsqr\lazb(\X)},
	\end{align*}
	with probability at least $1-\frac{1}{n^3}$.
\end{lemma}
The next three lemmas are immediate consequences of similar results in \cite{oymak2020towards} (specifically Lemmas 5.7, 5.8, 6.12 respectively) and we therefore state them without proof.

\begin{lemma}[Spectral norm of the Jacobian]\label{spectJ} Consider a one-hidden layer neural network model of the form $\vct{x}\mapsto \vct{v}^T\sigma_\bal\left(\W\x\right)$ where the activations $\sigma_1, \sigma_2, \ldots, \sigma_\h$ have bounded derivatives obeying $\abs{\sigma_j'(z)}\le B$. Also assume we have $n$ data points $\vct{x}_1, \vct{x}_2,\ldots,\vct{x}_n\in\R^d$. Then the Jacobian matrix with respect to the input-to-hidden weights obeys 
	\begin{align*}
		\underset{\bal\in\Bal}{\max}\text{ }\opnorm{\mathcal{J}_\bal(\mtx{W})}\le \sqrt{k}B\infnorm{\vct{v}}\opnorm{\X}.
	\end{align*}
\end{lemma}

\begin{lemma}[Jacobian Lipschitzness]\label{JLlem} Consider a one-hidden layer neural network model of the form $\vct{x}\mapsto \vct{v}^T\sigma_\bal\left(\W\x\right)$ where the activations $\sigma_1, \sigma_2, \ldots, \sigma_\h$ have bounded second order derivatives obeying $\abs{\sigma_j''(z)}\le M$. Also assume we have $n$ data points $\vct{x}_1, \vct{x}_2,\ldots,\vct{x}_n\in\R^d$ with unit euclidean norm ($\twonorm{\vct{x}_i}=1$). Then the Jacobian mapping with respect to the input-to-hidden weights obeys
	\begin{align*}
		\underset{\bal\in\Bal}{\max}\text{ }\opnorm{\mathcal{J}_\bal(\widetilde{\mtx{W}})-\mathcal{J}_\bal(\mtx{W})}\le M\infnorm{\vct{v}}\opnorm{\mtx{X}}\fronorm{\widetilde{\mtx{W}}-\mtx{W}}\quad\text{for all}\quad \widetilde{\W},\W\in\R^{k\times d}.
	\end{align*}
\end{lemma}

\begin{lemma}[Upper bound on initial prediction]\label{upresz}  Consider a one-hidden layer neural network model of the form $\vct{x}\mapsto \vct{v}^T\sigma_\bal\left(\W\x\right)$ where the activations $\sigma_1, \sigma_2, \ldots, \sigma_\h$ have bounded derivatives obeying $\abs{\sigma_j'(z)}\le B$ and $h\le n$. Also assume we have $n$ data points $\vct{x}_1, \vct{x}_2,\ldots,\vct{x}_n\in\R^d$ with unit euclidean norm ($\twonorm{\vct{x}_i}=1$) aggregated as rows of a matrix $\X\in\R^{n\times d}$ and the corresponding labels given by $\vct{y}\in\R^n$. Furthermore, assume we set half of the entries of $\vct{v}\in\R^k$ to $\frac{\czmli }{\sqrt{k}}$ and the other half to $-\frac{\czmli }{\sqrt{k}}$. Then for $\mtx{W}\in\R^{k\times d}$ with i.i.d.~$\mathcal{N}(0,1)$ entries 
	\begin{align*}
		\twonorm{\sigma_\bal\left(\X\W^T\right)\vct{v}}\le \czmli \sqrt{n}\left(1+3\sqrt{\log n}\right) B,
	\end{align*}
	holds with probability at least $1-\frac{1}{n^3}$. Additionally, let $\Dc$ be any distribution supported over the unit Euclidean ball. With probability at least $1-\frac{1}{n^3}$, we have that
	\[
	\E_{\x\sim\Dc}|\vb^T\sigma_\bal(\W\x)|\leq \czmli \left(1+3\sqrt{\log n}\right) B.
	\]
\end{lemma}
\begin{proof}
	By the triangular inequality we have
	\begin{align*}
		\twonorm{\sigma_\bal\left(\X\W^T\right)\vct{v}}\le& \onenorm{\bal}\max_j \twonorm{\sigma_j\left(\X\W^T\right)\vct{v}}\\
		=&\max_j \twonorm{\sigma_j\left(\X\W^T\right)\vct{v}}
	\end{align*}
	The result holds by applying the union bound to Lemma 6.12 of \cite{oymak2020towards} with $\delta=3\sqrt{\log n}$.
	
	For the second result, observe that $\vb^T\sigma_\bal(\W\x)$ is a $\czmli B$ Lipschitz function of $\W$ via
	\begin{align}
		|\vb^T\sigma_j(\W\x)-\vb^T\sigma_j(\W'\x)|&\leq \tn{\vb}\tn{\sigma_j(\W\x)-\sigma_j(\W'\x)}\\
		&\leq \czmli B\tn{(\W-\W')\x}\leq \czmli B.
	\end{align}
	This means that the expectation function $f(\W)=\E_{\x\sim\Dc}|\vb^T\sigma_\bal(\W\x)|$ is $\czmli B$ Lipschitz as well. Finally, let us find its expectation over $\W\distas\Nn(0,1)$ via
	\begin{align}
		\E[f(\W)]=\E[|\vb^T\sigma_\bal(\W\x)|]=\sup_{1\leq j\leq h} \E_{\g\sim\Nn(0,\Iden)}[|\vb^T\sigma_j(\g)|]\leq B\E_{\g\sim\Nn(0,\Iden)}[|\vb^T\g|]\leq \czmli B.
	\end{align}
	Here, the final line follows from Gaussian contraction inequality. Overall, the Lipschitz tail bound yields $\Pro(f(\W)\geq (1+3\sqrt{\log n})\czmli B)\leq n^{-3}$. 
\end{proof}
With these auxiliary lemmas in place we now state a more general version of the main theorem proven later on in this section.

\begin{theorem}[Meta theorem]\label{thmshallowsmooth} Consider a data set of input/label pairs $\vct{x}_i\in\R^d$ and $y_i\in\R$ for $i=1,2,\ldots,n$ aggregated as rows/entries of a data matrix $\mtx{X}\in\R^{n\times d}$ and a label vector $\vct{y}\in\R^n$. Without loss of generality we assume the dataset is normalized so that $\twonorm{\vct{x}_i}=1$. Also consider a one-hidden layer neural network with $k$ hidden units and one output of the form $\vct{x}\mapsto \vct{v}^T\sigma_\bal\left(\mtx{W}\vct{x}\right)$ with $\mtx{W}\in\R^{k\times d}$ and $\vct{v}\in\R^k$ the input-to-hidden and hidden-to-output weights. We assume the activations $, \sigma_2, \ldots, \sigma_\h$ has bounded derivatives i.e.~$\abs{\sigma_j'(z)}\le B$ and $\abs{\sigma_j''(z)}\le M$ for all $z$. Also let \clr{$\lazb(\X)$ denote the normalized} minimum eigenvalue of the neural net covariance per Assumption \ref{mineigass}. Furthermore, we fix $\vct{v}$ and train only over $\mtx{W}$. Starting from an initial weight matrix $\mtx{W}_0$ selected at random with i.i.d.~$\mathcal{N}(0,1)$ entries we run Gradient Descent (GD) updates of the form $\mtx{W}_{\tau+1}=\mtx{W}_\tau-\eta\nabla \mathcal{L}(\mtx{W}_\tau)$ with step size $\eta\le\frac{1}{2kB^2\infnorm{\vct{v}}^2\opnorm{\mtx{X}}^2}$. Then, as long as
	\begin{align}
		\label{overparam1}
		\frac{\twonorm{\vct{v}}^2}{\infnorm{\vct{v}}}\ge& 64M\frac{\czsqr\opnorm{\mtx{X}}}{\laz(\X)}\tn{f(\mtx{W}_0)-\y}\nonumber\\
		\frac{\twonorm{\vct{v}}}{\infnorm{\vct{v}}}\ge& \sqrt{30h\log(nk)}\frac{\opnorm{\X}}{\sqrt{\lazb(\X)}}B\nonumber\\
		\frac{\twonorm{\vct{v}}^5}{k^{1.5}\infnorm{\vct{v}}^4}\ge&\frac{9216}{\gamma}MB^3\tn{f(\mtx{W}_0)-\y}\frac{\opnorm{\mtx{X}}^4}{\lazb^{2.5}(\mtx{X})}
	\end{align}
	and $c>0$ a fixed numerical constant, then with probability at least $1-\frac{1}{n^3}-4e^{-10h}$, for all activation choices $\bal\in\Bal$, all GD iterates obey
	\begin{align}
		&\tn{f(\mtx{W}_\tau)-\y}^2\leq \left(1-\eta\frac{\twonorm{\vct{v}}^2\lazb(\mtx{X})}{8}\right)^\tau\tn{f(\mtx{W}_0)-\y}^2,\label{ineq1}\\
		&\fronorm{\mtx{W}_\tau-\mtx{W}_0}\leq \frac{8}{\twonorm{\vct{v}}\sqrt{\lazb(\mtx{X})}}\tn{f(\mtx{W}_0)-\y}\label{ineq2}\\
		&\|\mtx{W}_\tau-\mtx{W}_0\|_{2,\infty}\le \frac{16B\tin{\vb}\|\X\|}{\twonorm{\vct{v}}^2\lazb(\mtx{X})} \tn{f(\mtx{W}_0)-\y}\label{ineq3}
	\end{align}
	Furthermore, assume 
	\begin{align}
		\label{jacobconcen}
		\max_{\bal\in\Bal}\text{ }\opnorm{\mathcal{J}_\bal(\mtx{W}_0)\mathcal{J}_\bal^T(\mtx{W}_0)- K_\bal(\mtx{X})}\le \eps_0^2
	\end{align}
	holds with $\eps_0\le \frac{\gamma}{512}\frac{\twonorm{\vct{v}}^4\lazb^2(\mtx{X})}{k^{1.5}\infnorm{\vct{v}}^3B^3\opnorm{\mtx{X}}^3}$ for some $\gamma\le 1$. Then
	\begin{align}
		\label{ineq4}
		\fronorm{\mtx{W}_t-\widetilde{\W}_\infty}\le \frac{5}{4}\frac{\gamma}{\sqrt{k}B\infnorm{\vct{v}}\opnorm{\X}}\twonorm{\vct{r}_0}+2\left(1-\frac{1}{4}\eta\twonorm{\vct{v}}^2\lazb(\X)\right)^t\frac{\twonorm{\vct{r}_0}}{\twonorm{\vct{v}}\sqrt{\lazb(\X)}}
	\end{align}
	holds with probability at least $1-\frac{1}{n^3}-4e^{-10h}$ on the same event. 
\end{theorem}

\subsubsection*{Finalizing the Proof of Theorem \ref{maincor}}
\begin{proof}
We now demonstrate how Theorem \ref{maincor} follows from the meta theorem above. To this aim first note that $\twonorm{\vct{v}}=\czmli $ and $\infnorm{\vct{v}}=\frac{\czmli }{\sqrt{k}}$. We also note that the choice of \eqref{kcond} (specifically the second summand involving $c_0$) implies \eqref{overparam1} so that the above meta theorem applies (with probability at least $1-\frac{1}{n^3}-4e^{-10h}$).

We now proceed by proving the various identities.

\noindent\underline{\textbf{Proof of \eqref{ineqc1}:}}\\
This follows immediately from Lemma \ref{upresz}.

\noindent\underline{\textbf{Proof of \eqref{ineqc2}:}}\\
Note that by Lemma \ref{Jacconcen} equation \eqref{ineqc1} holds with $\eps_0=\frac{\gamma}{512}\frac{\czmli\lazb^2(\mtx{X})}{ B^3\opnorm{\mtx{X}}^3}$ with probability at least $1-4e^{-10h}$ as long as we have
\begin{align*}
	\frac{1}{\|\vct{v}\|_{\ell_4}^4}\ge \frac{C}{\eps_0^4}(\log n) B^4\opnorm{\mtx{X}}^4h\quad\Leftrightarrow\quad& k\ge C\frac{\czsqr^2}{\eps_0^4}(\log n) B^4\opnorm{\mtx{X}}^4h\\
	\quad\Leftrightarrow\quad&k\ge C\frac{1}{\gamma^4\lazb^8(\mtx{X})}(\log n)B^{16}\opnorm{\mtx{X}}^{16}h 
\end{align*}
This is true as the latter is the same as \eqref{kcond}.

\noindent\underline{\textbf{Proofs of \eqref{ineqs conv} and \eqref{ineqs}:}}\\
For this statement, the critical ingredient is the fact that $\eps_0\le \frac{\gamma }{512}\frac{\czmli\lazb^2(\mtx{X})}{B^3\opnorm{\mtx{X}}^3}$ which follows from the proof of \eqref{ineqc2} (right above). With this in mind, the critical condition \eqref{jacobconcen} holds and \eqref{ineq4} is applicable. Thus, the proof of inequalities in \eqref{ineqs conv} and \eqref{ineqs} follow from their counter parts in Theorem \ref{thmshallowsmooth} (more specifically equations \eqref{ineq1}, \eqref{ineq2}, \eqref{ineq3}, and \eqref{ineq4}) by substituting the choice of $\vb$ and then noting that by Lemma \ref{upresz} and the upper bound on $\czmli$ 
\begin{align*}
	\twonorm{f(\mtx{W}_0)-\y}=\twonorm{\sigma_\bal\left(\X\W^T\right)\vct{v}-\y}\le\twonorm{\y}+\czmli \sqrt{n}\left(1+3\sqrt{\log n}\right) B\le 2\sqrt{n}
\end{align*}
holds with probability at least $1-\frac{1}{n^3}$. 

\noindent\underline{\textbf{Proof of \eqref{ineqc3}:}}\\
This result is also a direct application of the second statement of Lemma \ref{upresz} (i.e.~bounding the expected prediction over a distribution). Since we have two distributions ($\Dc_1,\Dc_2$), the probability of success is $1-2/n^3$.

\vspace{2pt}
\noi\textbf{\underline{Final step:}} Union bounding above results in an additional $\frac{3}{n^3}$ probability of error in the final statement to obtain an overall probability of success of $1-\frac{4}{n^3}-4e^{-10h}$.
\end{proof}

\subsection{Proof of Meta Theorem (Theorem \ref{thmshallowsmooth})}
\underline{\textbf{Proof of \eqref{ineq1} and \eqref{ineq2}:}}\\
The proof of equations \eqref{ineq1} and \eqref{ineq2} follow from Corollary 6.11 \cite{oymak2020towards} by replacing the following quantities in Corollary 6.11 \cite{oymak2020towards} using the auxiliary Lemmas \ref{minspectJ}, \ref{spectJ}, and \ref{JLlem}.
\begin{align*}
	\alpha:=\frac{1}{4}\twonorm{\vct{v}}\sqrt{\lazb(\X)}, \quad\beta:=\sqrt{k}B\infnorm{\vct{v}}\opnorm{\X},\quad L=M\infnorm{\vct{v}}\opnorm{\mtx{X}}
\end{align*}
\underline{\textbf{Proof of \eqref{ineq3}:}}\\
To prove this inequality note that
\begin{align*}
	\trow{\text{mat}\left(\mathcal{J}_\bal^T(\mtx{W})\vct{r}\right)}&= \trow{\text{diag}(\vct{v})\sigma_\bal'\left(\mtx{W}\mtx{X}^T\right)\text{diag}(\vct{r})\mtx{X}}\\
	&\le \tin{\vb}\max_{1\leq \ell\leq k}\tn{\sigma_\bal'\left(\w_\ell^T\mtx{X}^T\right)\text{diag}(\vct{r})\mtx{X}}\\
	&\leq \tin{\vb}\|\X\|\max_{1\leq \ell\leq k}\tn{\sigma_\bal'\left(\w_\ell^T\mtx{X}^T\right)\text{diag}(\vct{r})}\\
	&\leq B\tin{\vb}\|\X\|\tn{\vct{r}}
\end{align*}
Furthermore, by the triangular inequality we have
\begin{align*}
	\|\mtx{W}_\tau-\mtx{W}_0\|_{2,\infty}\le& \sum_{t=0}^{\tau-1} \|\mtx{W}_{t+1}-\mtx{W}_t\|_{2,\infty}\\
	=& \eta\sum_{t=0}^{\tau-1} \trow{\text{mat}\left(\mathcal{J}_\bal^T(\mtx{W}_t)\vct{r}_t\right)}\\
	\le& \eta B\tin{\vb}\|\X\|\sum_{t=0}^{\tau-1} \tn{\vct{r}_t}\\
	\le& \eta B\tin{\vb}\|\X\|\left(\sum_{t=0}^{\tau-1}\left(1-\eta\frac{\twonorm{\vct{v}}^2\laz(\mtx{X})}{8}\right)^{\frac{t}{2}} \right) \tn{f(\mtx{W}_0)-\y}\\
	\le& \frac{\eta B\tin{\vb}\|\X\|}{1-\sqrt{1-\eta\frac{\twonorm{\vct{v}}^2\czsqr\lazb(\mtx{X})}{8}}} \tn{f(\mtx{W}_0)-\y}\\
	\le& \frac{16B\tin{\vb}\|\X\|}{\twonorm{\vct{v}}^2\lazb(\mtx{X})} \tn{f(\mtx{W}_0)-\y}
\end{align*}
where in the last inequality we used the fact that for $0\le x\le 1$ we have $\frac{1}{1-\sqrt{1-x}}\le \frac{2}{x}$.

\noindent\underline{\textbf{Proof of \eqref{ineq4}:}}\\
To prove this inequality we utilize Theorem 4 of \cite{heckel2020compressive} with $\alpha:=\sqrt{2}\alpha$, $\beta:=\beta$, $\eps_0:=2\eps_0=\gamma\frac{\alpha^4}{\beta^3}$, $\eps:=\gamma\frac{\alpha^4}{\beta^3}$. Assumption 1 of this theorem is satisfied by the definition of $\alpha$ and $\beta$. Also using \eqref{jacobconcen} Assumption 2 of \cite{heckel2020compressive} holds with $2\eps_0\le \frac{\alpha^4}{\beta^3}$. Also in this case the value of $R$ in this theorem becomes equal to 
\begin{align*}
	R:=2\twonorm{\mtx{J}^\dagger\vct{r}_0}+\frac{2.5\gamma}{\beta}\twonorm{\vct{r}_0}\le \frac{4.5}{\alpha}\twonorm{\vct{r}_0}
\end{align*}
Furthermore, note that as long as
\begin{align*}
	9216MB^3\frac{\twonorm{\vct{r}_0}}{\gamma}\frac{\opnorm{\mtx{X}}^4}{\lazb^{2.5}(\mtx{X})}\le\frac{\twonorm{\vct{v}}^5}{k^{1.5}\infnorm{\vct{v}}^4},
\end{align*}
we have
\begin{align*}
	9\twonorm{\vct{r}_0}L\le \frac{\gamma\alpha^5}{\beta^3}.
\end{align*}
which implies that 
\begin{align*}
	RL\le\frac{4.5}{\alpha}\twonorm{\vct{r}_0}L\le \frac{\gamma\alpha^4}{2\beta^3}=\frac{\eps}{2}
\end{align*}
so that Assumption 3 of this theorem is also satisfied. Thus, equation (37) of \cite{heckel2020compressive} implies
\begin{align*}
	\fronorm{\mtx{W}_t-\widetilde{\W}_t}\le \frac{5}{4}\frac{\gamma}{\beta}\twonorm{\vct{r}_0}
\end{align*}
which together with the triangular inequality implies
\begin{align}
\label{mistake}
	\fronorm{\mtx{W}_t-\widetilde{\W}_\infty}\le\fronorm{\mtx{\W}_t-\widetilde{\W}_t}+\fronorm{\widetilde{\W}_t-\widetilde{\W}_\infty} \le \frac{5}{4}\frac{\gamma}{\beta}\twonorm{\vct{r}_0}+\fronorm{\widetilde{\W}_t-\widetilde{\W}_\infty}
\end{align}
Define $\widetilde{r}_\tau=\mtx{J}\text{vect}(\widetilde{\W}_\tau)-\vct{y}$. All that remains to complete the proof is to bound the last term. To this aim consider the singular value decomposition of $\mtx{J}=\mtx{U}_{\mtx{J}}\mtx{\Sigma}_{\mtx{J}}\mtx{V}_{\mtx{J}}^T$ and note that
\begin{align*}
\text{vect}(\widetilde{\W}_t)-\text{vect}(\widetilde{\W}_{\infty})=&\eta\mtx{J}^T\sum_{\tau=t}^\infty\widetilde{\vct{r}}_\tau\\
=&\eta\mtx{J}^T\sum_{\tau=0}^\infty\widetilde{\vct{r}}_{\tau+t}\\
=&\eta\mtx{J}^T\left(\mtx{I}-\eta\mtx{J}\mtx{J}^T\right)^t\sum_{\tau=0}^\infty\widetilde{\vct{r}}_{\tau}\\
=&\eta\mtx{V}_{\mtx{J}}\mtx{\Sigma}_{\mtx{J}}\left(\mtx{I}-\eta \mtx{\Sigma}_{\mtx{J}}^2\right)^t\mtx{U}_{\mtx{J}}^T\sum_{\tau=0}^\infty\widetilde{\vct{r}}_{\tau}\\
=&\eta\mtx{V}_{\mtx{J}}\left(\mtx{I}-\eta \mtx{\Sigma}_{\mtx{J}}^2\right)^t\mtx{\Sigma}_{\mtx{J}}\mtx{U}_{\mtx{J}}^T\sum_{\tau=0}^\infty\widetilde{\vct{r}}_{\tau}\\
=&\eta\mtx{V}_{\mtx{J}}\left(\mtx{I}-\eta \mtx{\Sigma}_{\mtx{J}}^2\right)^t\mtx{V}_{\mtx{J}}^T\mtx{V}_{\mtx{J}}\mtx{\Sigma}_{\mtx{J}}\mtx{U}_{\mtx{J}}^T\sum_{\tau=0}^\infty\widetilde{\vct{r}}_{\tau}\\
=&\eta\mtx{V}_{\mtx{J}}\left(\mtx{I}-\eta \mtx{\Sigma}_{\mtx{J}}^2\right)^t\mtx{V}_{\mtx{J}}^T\mtx{J}^T\sum_{\tau=0}^\infty\widetilde{\vct{r}}_{\tau}\\
=&\mtx{V}_{\mtx{J}}\left(\mtx{I}-\eta \mtx{\Sigma}_{\mtx{J}}^2\right)^t\mtx{V}_{\mtx{J}}^T\left(\text{vect}(\widetilde{\W}_0)-\text{vect}(\widetilde{\W}_{\infty})\right).
\end{align*}
Now using the fact that $\text{vect}(\widetilde{\W}_0)-\text{vect}(\widetilde{\W}_{\infty})$ belongs to span$(\mtx{J}^T)$ we conclude that
\begin{align*}
\fronorm{\mtx{W}_t-\widetilde{\W}_\infty}\le& \left(1-4\eta\alpha^2\right)^t\fronorm{\mtx{W}_0-\widetilde{\W}_\infty}\\
\le& \left(1-4\eta\alpha^2\right)^t\frac{\twonorm{\vct{r}_0}}{2\alpha}\\
=&2\left(1-\frac{1}{4}\eta\twonorm{\vct{v}}^2\lazb(\X)\right)^t\frac{\twonorm{\vct{r}_0}}{\twonorm{\vct{v}}\sqrt{\lazb(\X)}}
\end{align*}
Plugging the latter into \eqref{mistake} completes this proof.
\subsection{Proofs for Jacobian concentration (Proof of Lemma \ref{Jacconcen})}
\begin{proof}
	To lower bound the minimum eigenvalue of $\mathcal{J}_{\bal}(\W_0)$ universally for all $\bal$, we focus on lower bounding the minimum eigenvalue of $\mathcal{J}_{\bal}(\W_0)\mathcal{J}_\bal(\W_0)^T$ for a fixed $\bal$. To this aim we use the identity
	\begin{align*}
		\mathcal{J}_\bal(\mtx{W})\mathcal{J}_\bal^T(\mtx{W})=&\left(\sigma_\bal'\left(\mtx{X}\mtx{W}^T\right)\text{diag}\left(\vct{v}\right)\text{diag}\left(\vct{v}\right)\sigma_\bal'\left(\mtx{W}\mtx{X}^T\right)\right)\odot\left(\mtx{X}\mtx{X}^T\right)\\
		=&\left(\sum_{\ell=1}^k\vct{v}_\ell^2\sigma_\bal'\left(\X\vct{w}_\ell\right)\sigma_\bal'\left(\X\vct{w}_\ell\right)^T\right)\odot \left(\X\X^T\right),
	\end{align*}
	mentioned earlier to conclude that
	\begin{align}
		\E\big[\mathcal{J}_\bal(\W_0)\mathcal{J}_\bal(\W_0)^T\big]=&\twonorm{\vct{v}}^2\left(\E_{\vct{w}\sim\mathcal{N}(0,\mtx{I}_d)}\big[\sigma_\bal'\left(\X\vct{w}\right)\sigma_\bal'\left(\X\vct{w}\right)^T\big]\right)\odot\left(\X\X^T\right),\nn\\
		:=\Kb_\bal\left(\X\right).
	\end{align}
	Now define 
	\begin{align*}
		\mtx{S}_\ell(\bal)=\big[\vct{v}_\ell^2\left(\sigma_\bal'\left(\X\vct{w}_\ell\right)\sigma_\bal'\left(\X\vct{w}_\ell\right)^T\right)\odot \left(\X\X^T\right)-\Kb_\bal\left(\X\right)\big]
	\end{align*}
	and note that $\E[\mtx{S}_\ell]=0$. Thus,
	\begin{align*}
		\mathcal{J}_\bal(\W_0)\mathcal{J}_\bal^T(\W_0)-\Kb_\bal(\mtx{X})
		=&\sum_{\ell=1}^k\big[\vct{v}_\ell^2\left(\sigma_\bal'\left(\X\vct{w}_\ell\right)\sigma_\bal'\left(\X\vct{w}_\ell\right)^T\right)\odot \left(\X\X^T\right)-\Kb_\bal\left(\X\right)\big]\\
		=&\sum_{\ell=1}^k\mtx{S}_\ell(\bal)
	\end{align*}
	To show that the spectral norm is small we will use the matrix Hoeffding inequality. Next note that
	\begin{align*}
		\mtx{S}_\ell(\bal) \preceq& v_\ell^2\left(\sigma_\bal'\left(\X\vct{w}_\ell\right)\sigma_\bal'\left(\X\vct{w}_\ell\right)^T\right)\odot \left(\X\X^T\right)\\
		=&v_\ell^2\text{diag}\left(\sigma_\bal'\left(\X\vct{w}_\ell\right)\right)\mtx{X}\mtx{X}^T\text{diag}\left(\sigma_\bal'\left(\X\vct{w}_\ell\right)\right)\\
		\preceq&v_\ell^2B^2\mtx{X}\mtx{X}^T
	\end{align*}
	Similarly, $\mtx{S}_\ell(\bal)\succeq -v_\ell^2B^2\mtx{X}\mtx{X}^T$. Thus, by matrix Hoeffding inequality we have
	\begin{align*}
		\mathbb{P}\Bigg\{\opnorm{\sum_{\ell=1}^k\mtx{S}_\ell(\bal)}\ge t\Bigg\}\le 2ne^{-\frac{t^2}{8\sigma^2}}\quad\text{where}\quad \sigma^2:=\opnorm{\sum_{\ell=1}^k v_\ell^4B^4\mtx{X}\mtx{X}^T\mtx{X}\mtx{X}^T}
	\end{align*}
	Thus, for a fixed $\bal$ we have
	\begin{align}
		\label{mytmpc}
		\mathbb{P}\Bigg\{\opnorm{\sum_{\ell=1}^k\mtx{S}_\ell(\bal)}\ge t\Bigg\}\le 2ne^{-\frac{t^2}{8\sigma^2}}\le 2ne^{-\frac{t^2}{B^4\|\vct{v}\|_{\ell_4}^4\opnorm{\mtx{X}}^4}}.
	\end{align}
	Next note that for fixed $\bal$ and $\widetilde{\bal}$ we have
	\begin{align*}
		&\mtx{S}_\ell(\bal)-\mtx{S}_\ell(\widetilde{\bal})\\
		&=\vct{v}_\ell^2\Bigg[\left(\sigma_{\bal-\widetilde{\bal}}'\left(\X\vct{w}_\ell\right)\sigma_{\bal-\widetilde{\bal}}'\left(\X\vct{w}_\ell\right)^T+\sigma_{\bal-\widetilde{\bal}}'\left(\X\vct{w}_\ell\right)\sigma_{\widetilde{\bal}}'\left(\X\vct{w}_\ell\right)^T+\sigma_{\widetilde{\bal}}'\left(\X\vct{w}_\ell\right)\sigma_{\bal-\widetilde{\bal}}'\left(\X\vct{w}_\ell\right)^T\right)\odot \left(\X\X^T\right)\Bigg]\\
		&\quad-\vct{v}_\ell^2\E\Bigg[\left(\sigma_{\bal-\widetilde{\bal}}'\left(\X\vct{w}_\ell\right)\sigma_{\bal-\widetilde{\bal}}'\left(\X\vct{w}_\ell\right)^T+\sigma_{\bal-\widetilde{\bal}}'\left(\X\vct{w}_\ell\right)\sigma_{\widetilde{\bal}}'\left(\X\vct{w}_\ell\right)^T+\sigma_{\widetilde{\bal}}'\left(\X\vct{w}_\ell\right)\sigma_{\bal-\widetilde{\bal}}'\left(\X\vct{w}_\ell\right)^T\right)\odot \left(\X\X^T\right)\Bigg]\\
		&\preceq 8\vct{v}_\ell^2B^2\onenorm{\bal-\widetilde{\bal}}\opnorm{\mtx{X}}^2\mtx{I}_n
	\end{align*}
	Thus, for fixed $\bal$ and $\widetilde{\bal}$ we have
	\begin{align*}
		\mathbb{P}\Bigg\{\opnorm{\sum_{\ell=1}^k\mtx{S}_\ell(\bal)-\mtx{S}_\ell(\widetilde{\bal})}\ge t\Bigg\}\le 2ne^{-\frac{t^2}{64B^4\onenorm{\bal-\widetilde{\bal}}^2\|\vct{v}\|_{\ell_4}^4\opnorm{\mtx{X}}^4}}.
	\end{align*}
	Thus in turn implies that for fixed $\bal$ and $\widetilde{\bal}$ we have
	\begin{align*}
		\Bigg\|\opnorm{\sum_{\ell=1}^k\mtx{S}_\ell(\bal)}-\opnorm{\sum_{\ell=1}^k\mtx{S}_\ell(\widetilde{\bal})}\Bigg\|_{\psi_2}
		\le&\Bigg\| \opnorm{\sum_{\ell=1}^k\mtx{S}_\ell(\bal)-\mtx{S}_\ell(\widetilde{\bal})}\Bigg\|_{\psi_2}\\
		\le&c \sqrt{\log n}B^2\onenorm{\bal-\widetilde{\bal}}\|\vct{v}\|_{\ell_4}^2\opnorm{\mtx{X}}^2
	\end{align*}
	Thus using Talagrand's comparison inequality (Corollary 8.6.2 of \cite{vershynin2018high} and Exercise ) for a fixed $\bal_0\in\Bal$ and all $\bal\in\Bal$ we have
	\begin{align*}
		\sup_{\bal\in\Bal}\text{ }\abs{\opnorm{\sum_{\ell=1}^k\mtx{S}_\ell(\bal)}-\opnorm{\sum_{\ell=1}^k\mtx{S}_\ell(\bal_0)}}\le c \sqrt{\log n}B^2\|\vct{v}\|_{\ell_4}^2\opnorm{\mtx{X}}^2\left(\sqrt{h}+2u\right)
	\end{align*}
	holds with probability at least $1-2e^{-u^2}$. Plugging in $u=\sqrt{10h}$ we arrive at
	\begin{align*}
		\sup_{\bal\in\Bal}\text{ }\abs{\opnorm{\sum_{\ell=1}^k\mtx{S}_\ell(\bal)}-\opnorm{\sum_{\ell=1}^k\mtx{S}_\ell(\bal_0)}}\le c \sqrt{\log n}B^2\|\vct{v}\|_{\ell_4}^2\opnorm{\mtx{X}}^2\sqrt{h}
	\end{align*}
	holds with probability at least $1-2e^{-10h}$. Thus, using the triangular inequality combined with \eqref{mytmpc} we have
	\begin{align*}
		\sup_{\bal\in\Bal}\text{ }\opnorm{\sum_{\ell=1}^k\mtx{S}_\ell(\bal)}\le& c \sqrt{\log n}B^2\|\vct{v}\|_{\ell_4}^2\opnorm{\mtx{X}}^2\sqrt{h}+\opnorm{\sum_{\ell=1}^k\mtx{S}_\ell(\bal_0)}\\
		\le& 2c \sqrt{\log n}B^2\|\vct{v}\|_{\ell_4}^2\opnorm{\mtx{X}}^2\sqrt{h}
	\end{align*}
	holds with probability at least $1-4e^{-10h}$.
\end{proof}

\subsection{Proof for minimum eigenvalue of Jacobian at initialization (Proof of Lemma \ref{minspectJ})}
To lower bound the minimum eigenvalue of $\mathcal{J}_{\bal}(\W_0)$ universally for all $\bal$, we focus on lower bounding the minimum eigenvalue of $\mathcal{J}_{\bal}(\W_0)\mathcal{J}_\bal(\W_0)^T$ for a fixed $\bal$. To this aim we use the identity
\begin{align*}
	\mathcal{J}_\bal(\mtx{W})\mathcal{J}_\bal^T(\mtx{W})=&\left(\sigma_\bal'\left(\mtx{X}\mtx{W}^T\right)\text{diag}\left(\vct{v}\right)\text{diag}\left(\vct{v}\right)\sigma_\bal'\left(\mtx{W}\mtx{X}^T\right)\right)\odot\left(\mtx{X}\mtx{X}^T\right)\\
	=&\left(\sum_{\ell=1}^k\vct{v}_\ell^2\sigma_\bal'\left(\X\vct{w}_\ell\right)\sigma_\bal'\left(\X\vct{w}_\ell\right)^T\right)\odot \left(\X\X^T\right),
\end{align*}
mentioned earlier to conclude that
\begin{align}
	\E\big[\mathcal{J}_\bal(\W_0)\mathcal{J}_\bal(\W_0)^T\big]=&\twonorm{\vct{v}}^2\left(\E_{\vct{w}\sim\mathcal{N}(0,\mtx{I}_d)}\big[\sigma_\bal'\left(\X\vct{w}\right)\sigma_\bal'\left(\X\vct{w}\right)^T\big]\right)\odot\left(\X\X^T\right),\nn\\
	:=\Kb_\bal\left(\X\right).
\end{align}
Thus
\begin{align}
	\label{mineigexp}
	\lambda_{\min}\left(\E\big[\mathcal{J}_\bal(\W_0)\mathcal{J}_\bal(\W_0)^T\big]\right)\ge\lambda_\bal(\X).
\end{align}
To relate the minimum eigenvalue of the expectation to that of $\mathcal{J}_\bal(\W_0)\mathcal{J}_\bal(\W_0)^T$ we utilize the matrix Chernoff identity stated below.
\begin{theorem}[Matrix Chernoff] Consider a finite sequence $\mtx{A}_\ell\in\R^{n\times n}$ of independent, random, Hermitian matrices with common dimension $n$. Assume that $\mtx{0}\preceq \mtx{A}_\ell\preceq R\mtx{I}$ for $\ell=1,2,\ldots,k$. Then
	\begin{align*}
		\mathbb{P}\Bigg\{\lambda_{\min}\left(\sum_{\ell=1}^k \mtx{A}_\ell\right)\le (1-\delta)\lambda_{\min}\left(\sum_{\ell=1}^k\E[\mtx{A}_\ell]\right)\Bigg\}\le n\left(\frac{e^{-\delta}}{(1-\delta)^{(1-\delta)}}\right)^{\frac{\lambda_{\min}\left(\sum_{\ell=1}^k\E[\mtx{A}_\ell]\right)}{R}}
	\end{align*}
	for $\delta\in[0,1)$.
\end{theorem}
We shall apply this theorem with $\mtx{A}_\ell:=\mathcal{J}_\bal(\vct{w}_\ell)\mathcal{J}_\bal^T(\vct{w}_\ell)=\vct{v}_\ell^2\text{diag}(\sigma_\bal'(\mtx{X}\vct{w}_\ell))\mtx{X}\mtx{X}^T\text{diag}(\sigma_\bal'(\mtx{X}\vct{w}_\ell))$. To this aim note that
\begin{align*}
	\vct{v}_\ell^2\text{diag}(\sigma_\bal'(\mtx{X}\vct{w}_\ell))\mtx{X}\mtx{X}^T\text{diag}(\sigma_\bal'(\mtx{X}\vct{w}_\ell))\preceq B^2\infnorm{\vct{v}}^2\opnorm{\mtx{X}}^2\mtx{I},
\end{align*}
so that we can use Chernoff Matrix with $R=B^2\infnorm{\vct{v}}^2\opnorm{\mtx{X}}^2$ to conclude that
\begin{align*}
	\mathbb{P}\Bigg\{\lambda_{\min}\left(\mathcal{J}_\bal(\W_0)\mathcal{J}_\bal^T(\W_0)\right)\le (1-\delta)\lambda_{\min}\left(\E\big[\mathcal{J}_\bal(\W_0)\mathcal{J}_\bal^T(\W_0)\big]\right)\Bigg\}\\
	\quad\quad\quad\quad\le n\left(\frac{e^{-\delta}}{(1-\delta)^{(1-\delta)}}\right)^{\frac{\lambda_{\min}\left(\E\big[\mathcal{J}_\bal(\W_0)\mathcal{J}_\bal^T(\W_0)\big]\right)}{B^2\infnorm{\vct{v}}^2\opnorm{\mtx{X}}^2}}.
\end{align*}
Thus using \eqref{mineigexp} in the above with $\delta=\frac{1}{2}$ we have
\begin{align*}
	\mathbb{P}\Bigg\{\lambda_{\min}\left(\mathcal{J}_\bal(\W_0)\mathcal{J}_\bal^T(\W_0)\right)\le \frac{1}{2}\tnv^2\lazb\left(\X\right)\Bigg\}\le&\mathbb{P}\Bigg\{\lambda_{\min}\left(\mathcal{J}_\bal(\W_0)\mathcal{J}_\bal^T(\W_0)\right)\le \frac{1}{2}\tnv^2\bar{\lambda}_\bal\left(\X\right)\Bigg\}\\
	\le& n\cdot e^{-\frac{1}{10}\frac{\tnv^2\bar{\lambda}_\bal(\X)}{B^2\infnorm{\vct{v}}^2\opnorm{\mtx{X}}^2}} \\
	\le& n\cdot e^{-\frac{1}{10}\frac{\tnv^2\lazb(\X)}{B^2\infnorm{\vct{v}}^2\opnorm{\mtx{X}}^2}}
\end{align*}
This proves the result for a fixed $\bal$. Now let $\mathcal{N}_{\epsilon}$ be an $\epsilon$, $\ell_1$ ball cover of $\Bal$ with $\epsilon:=\frac{\sqrt{2}-1}{2B\sqrt{kn}}\frac{\tnv}{\infnorm{\vct{v}}}\sqrt{\lazb(\mtx{X})}$. Then using the union bound we conclude that for all $\bal\in\mathcal{N}_{\eps}$
\begin{align*}
	\sigma_{\min}\left(\mathcal{J}_\bal(\W_0)\right)\ge \frac{\tnv}{\sqrt{2}}\sqrt{\lazb(\X)}
\end{align*}
holds with probability at least
\begin{align*}
	1-\abs{\mathcal{N}_\eps}n\cdot e^{-\frac{1}{10}\frac{\tnv^2\lazb(\X)}{B^2\infnorm{\vct{v}}^2\opnorm{\mtx{X}}^2}}\ge& 1-n\cdot\left(\frac{3}{\eps}\right)^\h  e^{-\frac{1}{10}\frac{\tnv^2\lazb(\X)}{B^2\infnorm{\vct{v}}^2\opnorm{\mtx{X}}^2}}\\
	=&1- \frac{1}{n}e^{2\log n+h\log\left(\frac{6B\sqrt{kn}\infnorm{\vct{v}}}{(\sqrt{2}-1)\tnv\sqrt{\lazb(\mtx{X})}}\right)-\frac{1}{10}\frac{\tnv^2\lazb(\X)}{B^2\infnorm{\vct{v}}^2\opnorm{\mtx{X}}^2}}\\
	\ge&1-\frac{1}{n^3}
\end{align*}
where the last line holds as long as
\begin{align*}
	\left(\frac{\tnv\sqrt{\lazb(\mtx{X})}}{B\infnorm{\vct{v}}}\right)^2\ge 10 \opnorm{\mtx{X}}^2\left(4\log n+h\log\left(\frac{6B\sqrt{kn}\infnorm{\vct{v}}}{(\sqrt{2}-1)\tnv\sqrt{\lazb(\mtx{X})}}\right)\right)
\end{align*}
which in turn holds as long as
\begin{align*}
	\frac{\tnv\sqrt{\lazb(\mtx{X})}}{B\infnorm{\vct{v}}}\ge 30\opnorm{\mtx{X}}\sqrt{h\log(nk)} \ge 30\opnorm{\mtx{X}}\cdot\max\left(\sqrt{\log(n)}, \sqrt{h\log(nk)}\right).
\end{align*}
Next, we focus on the deviation with respect to the $\bal$ parameter
\begin{align*}
	\mathcal{J}_{\widetilde{\bal}}\left(\mtx{W}_0\right)-\mathcal{J}_\bal\left(\mtx{W}_0\right)=\left(\text{diag}(\vct{v})\left(\sigma_{\widetilde{\bal}}'\left(\mtx{X}\mtx{W}_0^T\right)-\sigma_{\bal}'\left(\mtx{X}\mtx{W}_0^T\right)\right)\right)*\mtx{X}.
\end{align*}
Now using the fact that $(\mtx{A}*\mtx{B})(\mtx{A}*\mtx{B})^T=\left(\mtx{A}\mtx{A}^T\right)\odot \left(\mtx{B}\mtx{B}^T\right)$ we conclude that
\begin{align}
	\label{JJT}
	&\left(\mathcal{J}_{\widetilde{\bal}}\left(\mtx{W}_0\right)-\mathcal{J}_\bal\left(\mtx{W}_0\right)\right)\left(\mathcal{J}_{\widetilde{\bal}}\left(\mtx{W}_0\right)-\mathcal{J}_\bal\left(\mtx{W}_0\right)\right)^T\nonumber\\
	&\quad\quad\quad\quad=\left(\left(\sigma_{\widetilde{\bal}}'\left(\mtx{X}\mtx{W}_0^T\right)-\sigma_{\bal}'\left(\mtx{X}\mtx{W}_0^T\right)\right)\text{diag}(\vct{v})\text{diag}(\vct{v})\left(\sigma_{\widetilde{\bal}}'\left(\mtx{X}\mtx{W}_0^T\right)-\sigma_{\bal}'\left(\mtx{X}\mtx{W}_0^T\right)\right)^T\right)\nn\\
	&\hspace{60pt}\odot\left(\mtx{X}\mtx{X}^T\right).
\end{align}
To continue further we use the fact that for to PSD matrices $\mtx{A}$ and $\mtx{B}$ we have $\lambda_{\max}\left(\mtx{A}\odot \mtx{B}\right)\le\left(\max_i \mtx{A}_{ii}\right) \lambda_{\max}(\mtx{B})$ combined with \eqref{JJT} to conclude that 
\begin{align*}
	\opnorm{\mathcal{J}_{\widetilde{\bal}}\left(\mtx{W}_0\right)-\mathcal{J}_\bal\left(\mtx{W}_0\right)}^2\le&\opnorm{\mtx{X}}^2\left(\max_i \twonorm{\text{diag}(\vct{v})\left(\sigma_{\widetilde{\bal}}'\left(\mtx{W}_0\vct{x}_i\right)-\sigma_{\bal}'\left(\mtx{W}_0\vct{x}_i\right)\right)}^2\right)\\
	\le&\infnorm{\vct{v}}^2\opnorm{\mtx{X}}^2\left(\max_i \twonorm{\sigma_{\widetilde{\bal}}'\left(\mtx{W}_0\vct{x}_i\right)-\sigma_{\bal}'\left(\mtx{W}_0\vct{x}_i\right)}^2\right)\\
	=&\infnorm{\vct{v}}^2\opnorm{\mtx{X}}^2\left(\max_i \twonorm{\sum_{j=1}^\h\left(\widetilde{\bal}_j-\bal_j\right)\sigma_j'\left(\mtx{W}_0\vct{x}_i\right)}^2\right)\\
	\le&\infnorm{\vct{v}}^2\opnorm{\mtx{X}}^2\left(\max_i \max_j \twonorm{\sigma_j'\left(\mtx{W}_0\vct{x}_i\right)}^2\right)\onenorm{\widetilde{\bal}-\bal}^2\\
	\le&kB^2\infnorm{\vct{v}}^2\opnorm{\mtx{X}}^2\onenorm{\widetilde{\bal}-\bal}^2\\
	\le&kB^2\infnorm{\vct{v}}^2\fronorm{\mtx{X}}^2\onenorm{\widetilde{\bal}-\bal}^2\\
	=&knB^2\infnorm{\vct{v}}^2\onenorm{\widetilde{\bal}-\bal}^2.
\end{align*}
Thus,
\begin{align*}
	\opnorm{\mathcal{J}_{\widetilde{\bal}}\left(\mtx{W}_0\right)-\mathcal{J}_\bal\left(\mtx{W}_0\right)}\le \sqrt{kn}B\infnorm{\vct{v}}\onenorm{\widetilde{\bal}-\bal}
\end{align*}
By the definition of the $\mathcal{N}_\eps$ cover for any $\bal\in\Bal$ there exists a $\widetilde{\bal}\in\mathcal{N}_\eps$ obeying $\onenorm{\widetilde{\bal}-\bal}\le \eps$. Thus, using the above deviation inequality for any $\bal\in\Bal$ we have
\begin{align*}
	\sigma_{\min}\left(\mathcal{J}_\bal(\W_0)\right)\ge& \sigma_{\min}\left(\mathcal{J}_{\widetilde{\bal}}(\W_0)\right)-\sqrt{kn}B\infnorm{\vct{v}}\onenorm{\widetilde{\bal}-\bal}\\
	\ge&  \frac{1}{\sqrt{2}}\tnv\sqrt{\lazb(\X)}-\sqrt{kn}B\infnorm{\vct{v}}\eps\\
	=&  \frac{1}{\sqrt{2}}\tnv\sqrt{\lazb(\X)}-\sqrt{kn}B\infnorm{\vct{v}}\frac{\sqrt{2}-1}{2B\sqrt{kn}}\frac{\tnv}{\infnorm{\vct{v}}}\sqrt{\lazb(\mtx{X})}\\
	=&  \frac{1}{2}\tnv\sqrt{\lazb(\X)}
\end{align*}
completing the proof.
\section{Proof of Theorem \ref{one layer nas supp}}\label{sec mingchen}

The result follows by plugging the proper quantities in Theorem \ref{one layer nas}. Due to the output-layer scaling $\cz$, $\laz$ grows proportional to the initialization $\czsqr$. Thus to state a bound invariant of the initialization, we define the invariant lower bound $\bar{\la}_0 =  \laz/\czsqr$ and state the bounds in terms of this quantity. We remark that this is consistent with the literature on neural tangent kernel analysis. Specifically, we show that, in Theorem~\ref{one layer nas}, one can choose
\begin{itemize}
\item $k_0\propto \frac{B^{16}h\nt^8\log \nt}{\eps^4\lazb^8}$
\item $T_0\propto \frac{B^2\nt}{\lazb}\log{(\frac{B\sqrt{\nt}}{\eps\sqrt{\lazb}})}$. 
\item $p_0=4\nt^{-3}+4e^{-10h}$.
\item $c_0\propto \frac{\eps^2\lazb}{B^4\nt(1+3\sqrt{\log{\nt}})^2}$
\end{itemize}
to conclude with the proof of Theorem \ref{one layer nas supp}. The verification of this choice will be accomplished via Theorem~\ref{maincor}. The following is a restatement (more precise version) of Theorem \ref{one layer nas supp}.
\begin{theorem}[Neural activation search]\label{one layer nas supp2} Suppose input features are normalized as $\tn{\x}= 1$ and labels take values in $\{-1,1\}$. Pick $\Bal$ to be a subset of the unit $\ell_1$ ball. Suppose Assumption \ref{ntk assump} holds for $\bt_0\leftrightarrow\W_0$ and the candidate activations have first two derivatives ($|\sigma'_i|,|\sigma''_i|$) upper bounded by $B>0$. Furthermore, fix $\vb$ with half $\sqrt{\cz/k}$ and half $-\sqrt{\cz/k}$ entries for $\cz\propto \frac{\eps^2\lazb}{B^4\nt(1+3\sqrt{\log{\nt}})^2}$ (see supplementary). Also define the initialization-invariant lower bound $\blaz=\laz/\cz$. Finally, assume the network width obeys
\vs\[
k\gtrsim \eps^{-4}\lazb^{-8}B^{16}h\nt^8\log(\nt),
\]
for a tolerance level $\eps>0$ and the size of the validation data obeys $\nv\gtrsim \ordet{\h}$. Following the bilevel optimization scheme for the shallow activation search with learning rate $\eta= \frac{1}{2\cz B^2\|\X\|^2}$ choice and number of iterations obeying $T\gtrsim\frac{B^2\nt}{\lazb}\log{(\frac{B\sqrt{\nt}}{\eps\sqrt{\lazb}})}$, the misclassification bound (0-1 loss)
\vs\begin{align}
\Lcz(\ft_\bah)\leq  \min_{\bal\in\Bal}2B\sqrt{\frac{\cz\yT^T\Kb_\bal^{-1}\yT}{\nt}}+C\sqrt{\frac{\ordet{\h}+t}{\nv}}+\eps+\delta,\label{valid claim}
\end{align}
holds with probability at least $1-4(e^{-t}+\nt^{-3}+e^{-10h})$ (over the randomness in $\W_0,\Tc,\Vc$). Here, $\yT=[y_1~y_2~\dots~y_{\nt}]$. Finally, on the same event, for all $\bal\in\Bal$, training classification error obeys $\Lczh_{\Tc}(\ft_\bal)\leq \eps$.
\end{theorem}

\subsection{Finalizing the proof by verifying the $k_0,T_0,c_0,p_0$ choices}

{The main strategy for the proof is combining Theorem~\ref{maincor} with Theorem~\ref{one layer nas}. We ensure that all five summands in the error $3(\eps_0+\sqrt{c_0}BC_0/{\sqrt{k}}+\sqrt{c_0}B\eps_1+2B\eps_2/\sqrt{\blaz})+B\sqrt{\cz\eps_3}$ in Theorem~\ref{one layer nas} is less or equal to $1/5$ of the tolerance error $\eps$. To achieve this goal, we set  
\[
\gamma=\frac{\eps\opnorm{\X}}{75\sqrt{\nt}}\quad\text{and}\quad\czsqr = (\frac{\eps\sqrt{\lazb}}{30B^2\sqrt{\nt}(1+3\sqrt{\log{\nt}})})^2.
\]
in this section and prove the result (that these choices of $k_0,T_0,c_0,p_0$ are valid).}

\noindent \textbf{Step 0: Verifying $k\geq k_0$ satisfies conditions.} First, recall from Theorem~\ref{maincor} that we need to verify
\begin{align}
k\ge \frac{C(\log \nt) B^{16}\opnorm{\X}^{16}h}{\gamma^4\lazb^8} +C\frac{\nt B^8\|\X\|^8}{c_0\gamma^2\blaz^5}\label{eq verify k}
\end{align}
Observe that the second summand is bounded as follows (plugging in our $c_0$ choice)
\begin{align*}
\frac{nB^8\|\X\|^8}{c_0\gamma^2\blaz^5}&\propto \frac{\nt B^8\|\X\|^8}{\gamma^2\blaz^5}\times \frac{B^4\nt(1+3\sqrt{\log{\nt}})^2}{\eps^2\lazb}\\
&\propto \frac{\nt^2B^{12}\|\X\|^8 \log{\nt}}{\eps^2\gamma^2\blaz^6}.
\end{align*}
To proceed, by plugging in the value of $\gamma$ (and using $B^2\nt\geq \blaz$) we find that $k\geq k_0$ implies \eqref{eq verify k} via the following list of implications
\begin{align*}
&k\gtrsim \frac{ B^{16}\nt^2\opnorm{\X}^{12}h\log \nt}{\eps^4\lazb^8} +\frac{\nt^3B^{12}\|\X\|^6 \log{\nt}}{\eps^4\blaz^6} \\
&k\gtrsim \frac{ B^{16}\nt^8h\log \nt}{\eps^4\lazb^8}+\frac{\nt^6B^{12} \log{\nt}}{\eps^4\blaz^6} \impliedby \\
&k \propto \frac{ B^{16}\nt^8h\log \nt}{\eps^4\lazb^8}\impliedby\\
&k\geq k_0.
\end{align*}

\noindent \textbf{Step 1: Verifying the choice of $p_0$ and obtaining the values of $\eps_0,\eps_1,C_0,\eps_2,\eps_3$ for the itemized list of Theorem~\ref{one layer nas}.} We will substitute the bounds \eqref{ineqs}, \eqref{ineqc3}, \eqref{ineqc1}, \eqref{ineqc2} from Theorem~\ref{maincor} in the itemized assumptions of Theorem~\ref{one layer nas}. Thus, setting $p_0=4\nt^{-3}+4e^{-10h}$, the following conditions hold with probability ${1-\tfrac{4}{\nt^3}-4e^{-10h}}$ (which is the success probability of Theorem~\ref{maincor})
\begin{enumerate}
	\item $\E_{\x\sim\Dc}[|\vb^T\sigma_\bal(\W_0\x)|], \frac{1}{\nv}\sum_{i=1}^{\nv}|\vb^T\sigma_\bal(\W_0\xt_i)|\leq \eps_0$, where {$\eps_0=\czmli  B(1+3\sqrt{\log{\nt}})$}

	\item $T$'th iterate $\bt_T$ obeys 
	\[
	\tf{\W_T-\Wt_\infty}=\tn{\bt_T-\btt_\infty}\leq \eps_1=\frac{5}{2}\frac{\gamma}{\czmli B\opnorm{\X}}\sqrt{\nt}+4\left(1-\frac{1}{4}\eta \cz\lab_0(\mtx{X})\right)^t\frac{\sqrt{\nt}}{\sqrt{\cz\lazb(\X)}}.
	\] 
	\item Rows are bounded via $\trow{\W_T-\W_0}\leq \sqrt{C_0/k}$ and {$C_0=\frac{32^2   B^2 \opnorm{\X}^2}{\czsqr\lazb^2}$}
	\item At initialization, the network prediction is at most $\eps_2$ i.e. $\tn{\pb_\bal}\leq\eps_2=\czmli  \sqrt{\nt}B (1+3\sqrt{\log{\nt}})$
	\item Initial Jacobians obey $\frac{\Jb_\bal\Jb_\bal^T}{\czsqr}\succeq \lazb\Iden_{\nt}/2$.
	\item Initial Jacobians obey $\|(\Jb_\bal\Jb_\bal^T)^{-1}-\Kb_\bal^{-1}\|\leq \eps_3=\tfrac{2\gamma^2\lazb^2}{512^2 \czsqr B^6 \opnorm{\X}^6 }$.
\end{enumerate}

\noindent \textbf{Step 1.1: Bounding $\eps_1$ and verifying the choice of $T_0$.} In the second itemized condition of Step 1, we apply $\log$ on the second summand,
\begin{align*}
	\log{\left(4\left(1-\frac{1}{4}\eta \cz\lab_0(\mtx{X})\right)^t\frac{\sqrt{\nt}}{\sqrt{\cz\lazb(\X)}}\right)}&= t\log{\left(1-\frac{1}{4}\eta \cz\lab_0(\mtx{X})\right)}+\log{4\frac{\sqrt{\nt}}{\sqrt{\cz\lazb(\X)}}}\\
	&= t\log{\left(1-\frac{1}{4}\frac{1}{2\cz B^2\|\X\|^2} \cz\lab_0(\mtx{X})\right)}+\log(4\frac{\sqrt{\nt}}{\sqrt{\cz\lazb(\X)}})\\
	&\leq t\log{\left(1-\frac{\lab_0}{8 B^2\nt} \right)}+\log(4\frac{\sqrt{\nt}}{\sqrt{\cz\lazb}})\\
	&\leq t\left(-\frac{\lab_0}{8 B^2\nt} \right)+\log(4\frac{\sqrt{\nt}}{\sqrt{\cz\lazb}}).
\end{align*}
Here, we hope to ensure that $\eps_1 \le 5\frac{\gamma}{\czmli B\opnorm{\X}}\sqrt{\nt}$, Thus, we can bound t as following.
\begin{align*}
	t\left(-\frac{\lab_0}{8 B^2\nt} \right)+\log{4\frac{\sqrt{\nt}}{\sqrt{\cz\lazb}}} & \leq \log(\frac{5}{2}\frac{\gamma}{\czmli B\opnorm{\X}}\sqrt{\nt})\\
	\left(-\frac{\lab_0}{8 B^2\nt} \right)t &\leq \log(\frac{5}{8}\frac{\gamma\sqrt{\lazb}}{B\opnorm{\X}})\\
	t &\geq \frac{8 B^2\nt}{\lazb}\log(\frac{120 B\sqrt{\nt}}{\eps\sqrt{\lazb}}).
\end{align*}
Thus, as long as\footnote{Here, we choose the coefficient of $T_0$ to be 16 rather than $8$ which will help in \eqref{t0eq}.} $$T> T_0=\frac{16B^2\nt}{\lazb}\log(\frac{120 B\sqrt{\nt}}{\eps\sqrt{\lazb}})
\propto\frac{B^2\nt}{\lazb}\log(\frac{B\sqrt{\-nt}}{\eps\sqrt{\lazb}}).$$ We have $$\eps_1 \le 5\frac{\gamma}{\czmli B\opnorm{\X}}\sqrt{\nt}.$$

\noindent \textbf{Step 1.2: Applying $\eps_0,\eps_1,C_0,\eps_2,\eps_3$ to the error in Theorem~\ref{one layer nas}.} Considering Theorem~\ref{one layer nas}, we have the following overall error
\begin{align}
\text{error}=3(\eps_0+{\czmli BC_0}/{\sqrt{k}}+\czmli B\eps_1+2 B\eps_2/\sqrt{\lazb})+\czmli B\sqrt{\eps_3}. \label{eq verify eps}
\end{align}
We will now show the values of $k_0,c_0,T_0$ ensure that $\text{error}<\eps$. Specifically, plugging in the value of $\eps_0,\eps_1,C_0,\eps_2,\eps_3$ to \eqref{eq verify eps}, we have
\begin{align}
	\text{error} \le& 3\czmli  B(1+3\sqrt{\log{\nt}})+3\frac{32^2 B^3 \opnorm{\X}^2}{\czmli\lazb^2\sqrt{k}}+15\frac{\gamma}{\opnorm{\X}}\sqrt{\nt}\nn\\
	&+6\frac{\czmli B^2\sqrt{\nt} (1+3\sqrt{\log{\nt}})}{\sqrt{\lazb}}+\frac{\sqrt{2}\gamma\lazb}{512 B^2 \opnorm{\X}^3}. \label{eq verify eps 2}
\end{align}

\noindent \textbf{Step 2: Verifying $c_0$ and $\gamma$ satisfies conditions.} First, plugging in $c_0$ and $k\geq k_0$ into the second summand of \eqref{eq verify eps 2}, we find
\[
\frac{B^3 \opnorm{\X}^2}{\czmli\lazb^2\sqrt{k}}\leq\frac{B^3 \opnorm{\X}^2}{\lazb^2} \frac{B^2\sqrt{\nt}(1+3\sqrt{\log{\nt}})}{\eps\sqrt{\lazb}}\frac{\eps^2\lazb^4}{B^{8}\nt^4\sqrt{h\log \nt}}\leq c\eps\frac{\lazb^{3/2}}{B^3\nt^{3/2}\sqrt{h}}\leq c\eps.
\]
where $c>0$ can be made arbitrarily small by enlarging the constant multiplier of the $k_0$ lower bound on the width $k$.

Setting $\gamma=\frac{\eps\opnorm{\X}}{75\sqrt{\nt}}$ and $\czsqr = (\frac{\eps\sqrt{\lazb}}{30B^2\sqrt{\nt}(1+3\sqrt{\log{\nt}})})^2$ in \eqref{eq verify eps 2} and using the $c\eps$ bound above (set $c<(15\cdot32^2)^{-1}$), we find that each term in \eqref{eq verify eps 2} is less to or equal than $\frac{\eps}{5}$. 
\begin{align*}
	\text{error} &\le \frac{\eps \sqrt{\lazb}}{10B\sqrt{\nt}}+3\cdot32^2c\eps+\frac{\eps}{5}+\frac{\eps}{5}+\tfrac{\sqrt{2}\eps\lazb}{38400 B^2 \opnorm{\X}^2\sqrt{\nt}}.\\
	 \text{error} &\le 5\frac{\eps}{5} =\eps.
\end{align*}

Combine this with Theorem~\ref{one layer nas}, fix $M=\lipnn$, with probability ${1-4e^{-t}-\tfrac{4}{\nt^3}-4e^{-10h}}$, $\delta$-approximate NAS output obeys
\begin{align}
	\Lc(\ft_\bah)&\leq  \min_{\bal\in\Bal}2\czmli B\sqrt{\frac{\y^T\Kb_\bal^{-1}\y}{\nt}}+C\sqrt{\frac{\h\log(M)+t}{\nv}}+\text{error}+\delta,\nn
\end{align}
with $\text{error}\leq \eps$. This concludes the proof of the main claim \eqref{valid claim}. 

Finally, to conclude with the claim on the training risk satisfying $\Lczh_{\Tc}(\ft_\bal)\leq \eps$. Here, we will employ the first statement \eqref{ineqs conv} of Theorem \ref{maincor}. For all architectures $\bal$, using the fact that the least-squares dominate the 0-1 loss, for $T\geq T_0$, using $B^2n_\Tc\geq \laz$, we can write
\begin{align}
\Lczh(\ft_{\bal})&\leq \frac{1}{n}\tn{f(\mtx{W}_T)-\y}^2\leq 4\left(1-\eta\frac{\czsqr\lazb(\mtx{X})}{8}\right)^T\nn\\
&\leq 4\exp(-\eta T_0\frac{\czsqr\lazb(\mtx{X})}{8})\nn\\
&\leq 4\exp(-\frac{1}{2c_0B^2\|\X\|^2}\frac{c_0\blaz}{8}\frac{16 B^2\nt}{\lazb}\log(\frac{120 B\sqrt{\nt}}{\eps\sqrt{\lazb}}))\label{t0eq}\\
&\leq 4\exp(-\frac{\nt}{\|\X\|^2}\log(\frac{120 B\sqrt{\nt}}{\eps\sqrt{\lazb}}))\nn\\
&\leq 4\times {\frac{\eps\sqrt{\lazb}}{120B\sqrt{\nt}}}\leq \eps.\nn
\end{align}

\end{document}